\documentclass[final]{article}

\PassOptionsToPackage{numbers,square,sort&compress}{natbib}
\usepackage[main, final]{neurips_2026}

\usepackage[utf8]{inputenc} %
\usepackage[T1]{fontenc}    %
\usepackage{hyperref}       %
\usepackage{url}            %
\usepackage{booktabs}       %
\usepackage{amsfonts}       %
\usepackage{nicefrac}       %
\usepackage{microtype}      %
\usepackage{xcolor}         %

\usepackage{amssymb}
\usepackage{mathtools}
\usepackage{amsthm}
\usepackage{algorithm}
\usepackage{algorithmic}  
\usepackage{subcaption}
\usepackage{enumitem}
\usepackage{longtable}
\usepackage{array}
\usepackage{comment}
\usepackage{fontawesome5}

\theoremstyle{plain}
\newtheorem{theorem}{Theorem}[section]

\newtheorem{lemma}[theorem]{Lemma}

\theoremstyle{definition}

\newtheorem{assumption}[theorem]{Assumption}
\theoremstyle{remark}

\newcommand{\EE}{{\mathbb E}}
\newcommand{\PP}{{\mathbb P}}
\newcommand{\R}{{\mathbb R}}

\newcommand{\N}{{\mathbb N}}

\def\xx{\mbox{\boldmath $x$}}
\def\yy{\mbox{\boldmath $y$}}
\def\zz{\mbox{\boldmath $z$}}
\def\uu{\mbox{\boldmath $u$}}

\def\calG{{\cal G}}

\def\calM{{\cal M}}
\def\calN{{\cal N}}

\def\calW{{\cal W}}

\def\q{\quad}

\def\ria{\rightarrow}

\def\bze{\mathbf{0}}
\def\nn{\mathbf{n}}
\def\tt{\mathbf{t}}
\def\tr{\mathrm{tr}}
\def\op{\mathrm{op}}
\newcommand{\wh}[1]{\widehat{#1}}

\newcommand{\sd}[1]{{\scriptsize $\pm\,#1$}}
\newcommand{\M}{\mathcal{M}}

\DeclareMathOperator{\argmin}{argmin}

\DeclareMathOperator{\Memo}{Memo}

\definecolor{mygreen}{rgb}{0.1,0.75,0.2}

\definecolor{dred}{rgb}{0.6,0,0}
\definecolor{lgray}{rgb}{0.5,0.5,0.5}
\definecolor{dgrn}{rgb}{0.1,0.6,0.1}

\newcommand{\exphead}[2]{\par\smallskip\noindent\phantomsection\textbf{#2.}\label{#1}\ }
\title{Learning the Projection onto the Data Support: Distribution Learning that  Avoids Memorization}

\author{%
  Lin Chen\\
  Department of Mathematical Sciences\\
  Carnegie Mellon University\\
  Pittsburgh, PA 15213 \\
  \texttt{lc2@andrew.cmu.edu} \\
  \And
  Dejan Slep{\v{c}}ev \\
  Department of Mathematical Sciences\\
  Carnegie Mellon University\\
  Pittsburgh, PA 15213 \\
  \texttt{slepcev@andrew.cmu.edu} \\
}

\begin{document}

\maketitle 

\begin{abstract}
We introduce a framework for learning distributions supported on manifolds based on estimating the nearest-point projection onto the data support. Specifically, we define an objective functional over a class of functions whose minimizer recovers this projection. The proposed framework automatically adapts to the intrinsic dimension and allows for data supported on unions of manifolds with varying dimensions. The resulting objective can be optimized using standard neural network architectures. 
We present population-level results establishing recovery of the manifold projection and prove finite-sample Wasserstein generalization bounds.
Furthermore, we validate the approach empirically across a range of settings. \href{https://github.com/linchen-00/ProjectionLearning}{\textcolor{blue}{\faGithub}}

In contrast to widely used generative models such as diffusion models, whose objectives can be minimized by memorizing training samples, our formulation excludes such behavior. We also introduce a direct memorization metric to verify that the learned projection generates new samples rather than reproducing the training set.
\end{abstract}

\section{Introduction}

In many settings of interest, high-dimensional data are locally well represented, or at least well approximated, by low-dimensional manifolds or unions of manifolds. This observation has motivated a long line of work on learning and exploiting the geometric structure of data, including Isomap \citep{tenenbaum2000isomap}, locally linear embedding \citep{roweis2000LLE}, Laplacian eigenmaps \citep{belkin2003laplacian}, Hessian eigenmaps \citep{donoho2003hessian}, diffusion maps \citep{coifman2006diffusion}, and principal curves and manifolds \citep{hastie1989principal,zhang2004LTSA}; see also \citet{meila2024manifold} for a recent overview. These methods construct transformations that respect the underlying geometry, but do not directly recover the manifold itself. Manifold reconstruction methods do address this problem and have strong theoretical guarantees \citep{Aamari2018tangential,Aamari2019nonasymptotic,fefferman2020reconstruction}, but they often rely on representations such as coordinate charts or local polynomial patches that are cumbersome to use in high-dimensional learning pipelines.

In contrast, practical representation and generative learning methods often use flexible neural parameterizations, such as denoising autoencoders and diffusion-based generative models. For manifold-supported data, such methods can recover important aspects of the underlying geometry \citep{bengio2013representation,alain2014DAE,azangulov2024convergence,debortoli2022convergence,oko2023diffusion,tang2024adaptivity}. However, they do not directly provide a representation of the manifold, and in finite-sample regimes their objectives may favor maps or scores that concentrate around the training samples rather than recover the geometric support itself. This tension motivates the central question of this work: \emph{Can one use a flexible neural representation while designing an objective whose ideal optimizer recovers the underlying manifold structure rather than the empirical training set?}

\begin{figure}[ht]
    \centering
    \includegraphics[width=0.9\linewidth]{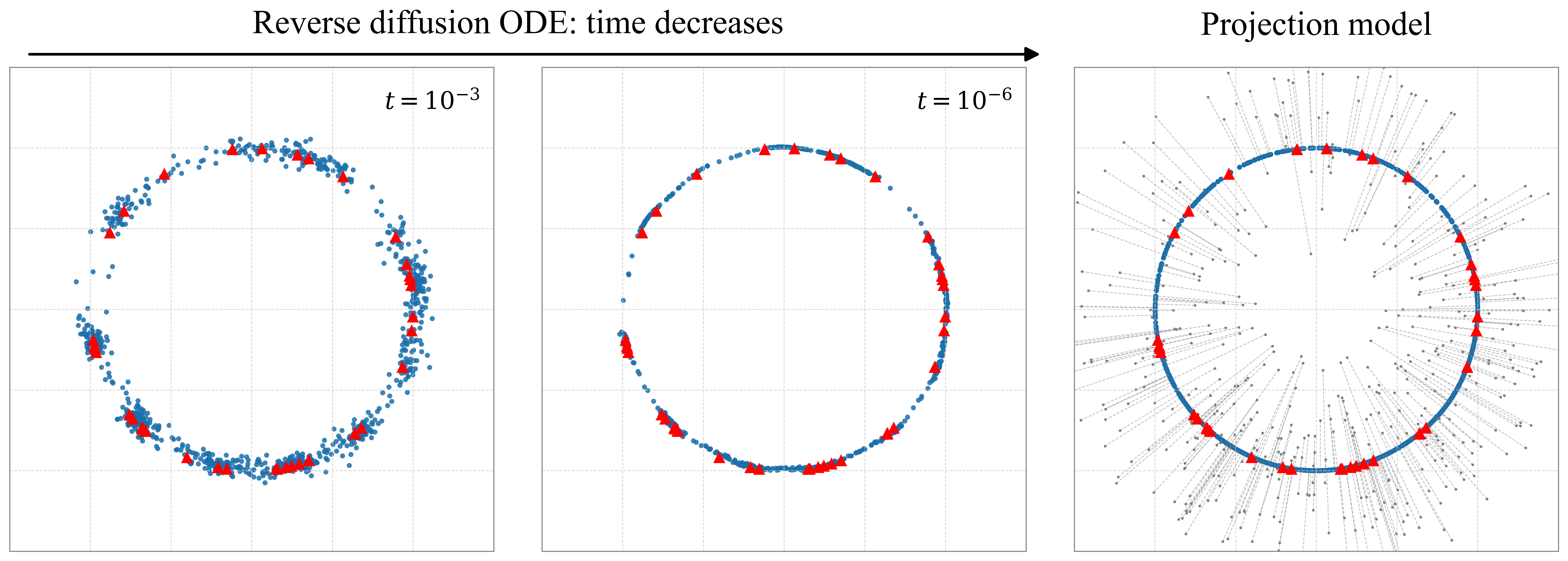}
    \caption{Denoising diffusion versus projection learning for the uniform measure on the unit circle in $\mathbb R^2$, trained on $40$ points (red triangles) with $1000$ generated samples (blue). \textbf{Left:} Reverse-ODE diffusion samples at $t=10^{-3},10^{-6}$; smaller time improves geometry but increases memorization near training points. \textbf{Right:} Learned projection of $1000$ noised samples ($\delta=0.7$, gray) identifies the circle precisely with negligible memorization.}
    \label{fig:circleprojdiff}
\end{figure}

We answer this question by learning the projection map onto the data manifold. Our approach is variational: given samples from a manifold, or from a union of manifolds, we introduce an objective functional whose minimizer is the projection onto sufficiently regular manifold(s) of the lowest dimension containing the data. This representation has several useful features. It is global, can be implemented with standard neural network architectures, does not rely on estimating tangent planes or coordinate charts, automatically and implicitly discovers the intrinsic dimension, and naturally accommodates unions of manifolds with different dimensions. Figure~\ref{fig:circleprojdiff} illustrates the distinction between projection-based generation and diffusion behavior on a simple example. More broadly, we view the learned projection as a practical geometric component that can also be combined with diffusion models as a correction in the final stage of generation. 

\textbf{Our contribution.} We make three main contributions. First, we introduce a variational characterization of manifold projection maps: the population objective recovers the manifold, and the full objective recovers the orthogonal projection. Second, we develop a finite-sample neural implementation and metrics for projection error and memorization. Third, we use the learned projection as a generator: noised training samples are mapped to the learned support, in the spirit of denoising autoencoders \citep{vincent2008DAE}. In contrast to denoising objectives, however, returning the training data does not minimize our objective (see Section~\ref{sec:finiteNstructure}).

\textbf{Geometry and memorization in diffusion models.}
Diffusion models, introduced by \citet{sohl2015deep} and developed through denoising diffusion and score-based generative modeling \citep{ho2020denoising,song2019generative,song2021score}, are now central tools in generative learning and achieve state-of-the-art performance in high-dimensional generation tasks \citep{dhariwal2021diffusion,rombach2022high,saharia2022photorealistic}. At a high level, these methods learn a score field from noised training data and then generate samples by following reverse stochastic or deterministic dynamics. Under the manifold hypothesis, several works show that diffusion-based methods can recover substantial geometric and distributional information.

\textbf{Manifold geometry in diffusion models.}  \citet{debortoli2022convergence} proves convergence of the reverse diffusion process to the data distribution when data are supported on a low-dimensional manifold, assuming that the score of the diffusion started from the full data distribution is learned sufficiently well. Building on \citet{oko2023diffusion}, \citet{tang2024adaptivity} shows that, for sufficiently smooth manifolds and densities, diffusion models can recover the density at nearly minimax rates; related intrinsic-dimension-dependent convergence guarantees are studied by \citet{azangulov2024convergence,potaptchik2025linearconvergencediffusionmodels}. These results require accurate score learning at the manifold scale. In finite samples, however, learning the empirical score too accurately at small noise can lead to memorization. We argue that a preferable model improves as its objective is optimized more accurately.

\textbf{Finite-sample memorization.} Despite their empirical success, score-based methods have a fundamental modeling limitation. Since empirical scores are learned from diffusions initiated at the training samples, increasingly accurate score estimation in the small-noise regime can steer reverse dynamics back toward those samples. Empirical studies show that diffusion models can replicate training images or produce samples highly similar to individual training examples \citep{somepalli2023diffusion,somepalli2023understanding}, and extraction attacks demonstrate that training data can sometimes be recovered from trained models \citep{carlini2023extracting}. In finite-sample regimes, empirical score estimates can lead to KDE-like or sample-replicating behavior \citep{baptista2025memorization,kamb2025an,li2024goodscoredoeslead,lyu2025resolvingmemorizationempiricaldiffusion}.
This concern is investigated by \citet{bonnaire2026momorization}, who argue that diffusion models often avoid memorization in practice through implicit dynamical regularization. They identify a time $\tau_{\mathrm{gen}}$, at which high-quality generation begins, and a later time $\tau_{\mathrm{mem}}$, beyond which memorization appears.  Since $\tau_{\mathrm{mem}}$ grows approximately linearly with the training set size while $\tau_{\mathrm{gen}}$ remains almost constant, there is a widening intermediate regime in which the model generalizes well. This helps explain the practical success of diffusion models, but it also reinforces our modeling concern: good behavior relies on remaining in a specific training regime.

\textbf{Mitigating memorization.}  Proposed mechanisms to mitigate memorization include early stopping of score matching or reverse dynamics, data augmentation \citep{somepalli2023understanding}, architectural bias or limited model capacity \citep{kadkhodaie2024generalization,kamb2025an}, implicit regularization in the  training dynamics \citep{bonnaire2026momorization,yuhan2025taking}, and explicit modifications of the final sampling dynamics. For example, \citet{lyu2025resolvingmemorizationempiricaldiffusion} modify the terminal stage of the reverse ODE to suppress tangential drift toward training samples. Such methods can be effective, but they still treat memorization as something to be controlled by training protocols, architectural constraints, or post-processing. Our goal is different: we seek an objective whose ideal optimizer represents the underlying geometric support itself.

 \textbf{Projection structure in diffusion models.} \citet{shen2026manifold} show that, for diffusion on manifold-supported data, coarse scores capture support geometry before fine-scale density structure. Their representation is related to ours through the potential $\eta_{\mathcal M}(x)=\frac12 d_{\mathcal M}^2(x)$, for which $\nabla \eta_{\mathcal M}(x)=x-\Pi_{\mathcal M}(x)$ near $\calM$. The main difference is where the projection structure resides: their recovery theorem restricts the learned score of diffusions to gradients of squared-distance potentials --- a class encoding the eikonal equation, the intrinsic dimension, and tangent-plane estimates at the samples --- so the projection structure is imposed on the hypothesis class. In our framework the admissible maps carry no structure beyond $C^2$ regularity; being a projection is a derived property of minimizers, and unions of manifolds with unknown, varying dimensions are allowed. The smooth-then-project measure $\Pi_{\calM\#}(\mu\ast\eta_\delta)$ is used in both works; our Theorem~\ref{thm:errorbdd} gives the bound $\calW_2(\mu,\Pi_{\calM\#}(\mu\ast\eta_\delta))\lesssim\delta^2$, complementing their one-sided coverage guarantee.
The interpretation of denoising as approximate projection and of sampling as descent on the squared distance is developed quantitatively by \citet{permenter2024interpreting}, who analyze DDIM under an assumed relative-error bound on a trained denoiser. There the projection property is a hypothesis about the trained denoiser; here it is a conclusion: the population minimizer is the orthogonal projection (Theorem~\ref{thm:miniT}), and every finite-sample minimizer maps onto a manifold near $\calM$, not onto the training set (Theorem~\ref{thm:finiteN}).

 \textbf{Projection-based learning.} The term ``projection learning'' has a different meaning in projection-based learning (PBL) \citep{elser2021learning,bergmeister2025projection}, where projection operators act on the network's own variables as the training dynamics, in place of gradient descent. Here, instead, the projection onto the data support is itself the object being learned, by ordinary gradient training, and serves as the generator. A neural precedent for the latter is \citet{chang2017projection}, which trains a network, adversarially, to project onto the set of natural images for use as a proximal operator in linear inverse problems, with no variational characterization of the projection and no recovery guarantee. A point of contact is \citet{elser2026machinelearnable}, where the learned set is likewise characterized as the fixed-point set of a map and memorizing solutions are excluded by bounding the map's complexity; in our formulation they are excluded by the geometric term $E_2$, with no capacity restriction (Theorem~\ref{thm:finiteN}).

\textbf{Outline.}
Section~\ref{sec:projpopulation} introduces the population-level variational problems whose minimizers recover the manifold and the orthogonal projection.
It also proves a Wasserstein error bound connecting projection accuracy to distribution recovery.
Section~\ref{sec:finite} develops the finite-sample version of the objective, states a result of companion work \citep{slepcev2026learning} showing that the image of every minimizer is a manifold approximating the support more accurately than the sample resolves it,
describes the neural-network training procedure, and introduces metrics used to quantify projection error and memorization. Section~\ref{sec:experiments} presents experiments on synthetic manifolds, unions of manifolds with different dimensions, sample-complexity scaling, and MNIST.

\section{Learning projections: population level}\label{sec:projpopulation}

We now formulate the population-level variational problem whose minimizer is the projection onto the data manifold. We assume that the data distribution is supported on a compact,  smooth embedded manifold $\calM\subset\mathbb R^D$ with positive reach. The framework extends naturally to unions of disjoint manifolds with different intrinsic dimensions, but we present the single-manifold case to keep the notation simple. Indeed, the reach of a disjoint union is at most half of the smallest distance between its connected components, so for $\delta$ below the reach, the $\delta$-tubes of the components are disjoint and the arguments apply to each component separately.

Projection is natural from the geometry of manifold-supported diffusion models: if $\mu$ is supported on $\calM$ with positive differentiable density and $p(x,t)$ is the heat evolution from $p_0=\mu$, then in a tubular neighborhood, as $t\to0$,
\[
\nabla \log p_t(x) = \frac{\Pi_\calM x - x}{t} + O(1),
\]
where $\Pi_\calM x$ is the closest-point projection onto $\calM$ \citep{Hino03diffusion,liu2026diffusion_manifold,shen2026manifold,stanczuk2024diffusion_models_intrinsic_dimension}. Thus for moderately small noise the dominant component of the score encodes the projection direction. With only finitely many samples, however, accurate empirical scores at very small noise  point toward individual training points. Thus, our goal is to learn the projection map directly, without relying on diffusion dynamics.

\subsection{Projection to manifold as a variational problem} \label{sec:population}

We first give the continuum objective, assuming the manifold $\calM$ is known.

Let $\calM$ be a compact $d$-dimensional $C^\infty$ manifold embedded in $\R^D$, with $d<D$ and positive reach $r_\calM$. Let $\calM_\delta:=\{x \in \R^D:\; d(x, \calM) < \delta\}$ be its $\delta$-tubular neighborhood. For $0 < \delta <r_\calM$, the orthogonal projection $\Pi_\calM: \overline{\calM_\delta} \ria \calM$ is well-defined, and we characterize it variationally.

Let $T \in C^\infty(\overline{\calM}_\delta; \R^D)$. Define the following loss terms:
\begin{equation}\label{eqn:lossdef}
\begin{alignedat}{2}
E_1(T) &:= \int_{\calM} |T(x) - x|^2 \, d\mathrm{vol}_\calM(x)
&\qquad
E_2(T) &:= \int_{\calM_\delta} |DT(x)(T(x) - x)|^2 \, dx \\
E_3^{\underline \delta}(T) &:= \int_{\calM_\delta} \left( d(T(x), \calM) - \underline \delta \right)_+ \, dx
&\qquad
E_4(T) &:= \int_{\calM_\delta} \left( |T(x) - x | - \delta \right)_+\, dx.
\end{alignedat}
\end{equation}
Here $\mathrm{vol}_\calM$ denotes the volume form of $\calM$ and $0<\underline \delta<\delta$; in practice we recommend choosing $\underline{\delta}$ to be the maximum nearest-neighbor distance between training samples, which reflects the covering scale of the training data. 
The term $E_1$ enforces the fixed-point property on $\calM$. The key term $E_2$ makes $T$ constant along the segment from $x$ to $T(x)$, matching the behavior of $\Pi_\calM$ along normal fibers: for $x$ near $\calM$ and $s\in[0,1]$, $T((1-s)x+sT(x))=T(x)$. Since $E_1$ and $E_2$ also vanish for the identity map, or for projections onto higher-dimensional sets containing $\calM$, $E_3$ nudges points far from $\calM$ back toward it. As discussed in Section~\ref{sec:experiments}, these three terms already work well in practice. Theoretically, however, they also allow for slanted projections; $E_4$ enforces orthogonality.
 
We define two loss functions 
\begin{equation} \label{eq:losses}
    E(T):= \sum_{i=1}^3 E_i(T) \qquad \text{and} \qquad E'(T):= \sum_{i=1}^4 E_i(T).  
\end{equation}
While $E'(\Pi_\M)=0$ is immediate, the converse requires proving that the projection is the only  minimizer. We do this in joint work with Anton Ullrich \citep{slepcev2026learning}.
We  also show that $E(T)=0$ identifies the manifold through $T(\overline \M_\delta)=\M$. Here we state the result and sketch the proof in Appendix~\ref{apx:proofminiT}.

\begin{theorem}\label{thm:miniT}
Assume $0<\delta < r_\M/3$ and $\underline \delta < \delta/2$.
    If $T \in C^\infty(\overline{\calM}_\delta; \R^D)$ is a minimizer of $E(T)$, then the range of $T$ is $\calM$, i.e.~$T(\overline{\calM}_\delta)=\calM$. Moreover, if $T$ is also a minimizer of $E'(T)$, then it is the orthogonal projection onto the manifold $\calM$, i.e.~$T=\Pi_\calM$.
\end{theorem}

\subsection{Error bound for distribution learning}\label{sec:errorbound}
We next quantify the distributional error from projecting noisy samples back to the manifold; in applications the learned map $T$ replaces the exact projection.
Let $\mu$ be a distribution supported on $\calM$ with density $\rho$ with respect to $\mathrm{vol}_\calM$, and let $\eta:\R \ria [0, \infty)$ be a smooth mollifier supported on $[-1, 1]$. Define the associated noise kernel by $\eta_\delta:=\delta^{-D}\eta(|\cdot| / \delta)$ for some bandwidth $\delta>0$, and let $\bar{\mu}_\delta := \Pi_{\calM \#} (\mu \ast \eta_\delta)$ be the pushforward of the noised measure $\mu \ast \eta_\delta$ by the orthogonal projection $\Pi_\calM$, with density $\bar\rho_\delta$ with respect to $\mathrm{vol}_\calM$. We derive an explicit bound for $\|\bar{\rho}_\delta - \rho\|_{L^\infty(\calM)}$ and $\calW_2(\mu, \bar{\mu}_\delta)$, showing that projection of noisy samples recovers the target distribution up to $\delta^2$-scale. 

Before going into further detail, we state the following assumption. 

\begin{assumption}\label{assumpmu}
(i) $\mu$ has a density $\rho \in C^3(\calM)$ with respect to $\mathrm{vol}_\calM$, with $0<1/ \rho_0 \le \rho \le \rho_0 <\infty$ for some $\rho_0>0$. (ii) The magnitude of noise satisfies $0<\delta < \min\{i_0, K^{-1/2}, r_\calM\} / 2$, where $i_0$ is the injectivity radius of $\calM$ and $K$ is the maximum of the absolute value of its sectional curvature.
\end{assumption}

\begin{theorem}\label{thm:rhoerrbdd}
    Under Assumption~\ref{assumpmu}, we have the following density bias estimate for $\bar{\rho}_\delta$:
    \begin{equation*}
        \|\bar{\rho}_\delta - \rho\|_{L^\infty(\calM)} \le \frac{d}{2D} \delta^2 \|\nabla_\calM^2 \rho\|_{\op}  + O(\delta^3),
    \end{equation*}
    where $\nabla_\calM^2 \rho$ is the Riemannian Hessian of $\rho$ on $\calM$ and $\|\nabla_\calM^2 \rho\|_{\op}:=\sup_{x\in \calM} \|\nabla_\calM^2 \rho(x)\|_{\mathrm{op}}$.

    Moreover, suppose that $\delta \le C / d $ with $C$ independent of $d$ and $D$, we have
    \begin{equation*}
        \|\bar{\rho}_\delta - \rho\|_{L^\infty(\calM)} \le
        C_{\calM, \rho}\frac dD\,\delta^2,
    \end{equation*}
    where
    $C_{\calM, \rho}$ is a constant depending on the geometry of $\calM$ and $\rho$, and is independent of $d$ and $D$.
\end{theorem}

\begin{theorem}\label{thm:errorbdd}
Under Assumption~\ref{assumpmu}, the following  bound in $2$-Wasserstein distance holds:
\[ \calW_2(\mu, \bar{\mu}_\delta) \le  \frac{d}{D} \delta^2 \mathrm{vol}(\calM)^{1/2}\rho_0^{1/2}\lambda_1^{-1/2}   \|\nabla_\calM^2 \rho\|_{\op}  + O(\delta^3), \]
where $\lambda_1$ is the smallest positive eigenvalue of $-\Delta_\calM$.
Furthermore, if $\delta \le C/ d$, we have $\calW_2(\mu, \bar{\mu}_\delta) \le C_{\calM, \rho}\frac dD\,\delta^2$, with constants $C$ and $C_{\calM, \rho}$ independent of $d$ and $D$.
\end{theorem}

\begin{proof}
    The full proof of this theorem can be found in Appendix~\ref{app:errorbdd}.
\end{proof}
These estimates quantify the bias introduced by projecting noised measure and reveal an explicit $d/D$ factor at sufficiently small noise scales. This ratio has a geometric explanation: $\calM$ locally resembles its $d$-dimensional tangent plane, and projection map removes normal displacements to first order while preserving tangential motion. For a fixed noise radius, isotropic noise distributes its variance across the $D$ ambient directions, placing a fraction $d/D$ in tangent directions.

\section{Learning projections from finite data} \label{sec:finite}

The population analysis gives a projection objective and stability bound for projected noisy samples. We now derive the empirical objective  for finite data, and parameterize the map by a neural network.

\subsection{Finite-sample optimization problem} \label{sec:discrete}
Let $X_{1:n}=\{X_i\}_{i=1}^n$ be i.i.d.~samples of $\mu$, let $\calM_{n,\delta}:=\bigcup_{i=1}^n B(X_i,\delta)$ be the tubular neighborhood of the data, and let $Y_{1:N}=\{Y_j\}_{j=1}^N$ be samples from $\calM_{n,\delta}$. Following~\eqref{eqn:lossdef}, define
\begin{equation*}
\setlength{\jot}{1pt}
\begin{alignedat}{2}
& E_{1,n}(T) := \frac{1}{n} \sum_{i=1}^n |T(X_i) - X_i|^2
&\quad & E_{2,n}(T) :=  s\, A(T) + (1-s)\, B(T) \\
& E_{3,n}^{\underline \delta}(T) := \frac{1}{N}\sum_{j=1}^N \left( |T(Y_j)- \Pi_n Y_j| - \underline \delta \right)_+
&\quad & E_{4,n}(T) := \frac{1}{N}\sum_{j=1}^N \left( |T(Y_j) - Y_j | - \delta \right)_+ \\
& A(T)  := \frac{1}{N}\sum_{j=1}^N |DT(T(Y_j))(T(Y_j) - Y_j)|^2
&\quad & B(T) := \frac{1}{N}\sum_{j=1}^N |DT(Y_j)(T(Y_j) - Y_j)|^2
\end{alignedat}
\end{equation*}
where $s \in  [0,1]$ and $\Pi_n(Y) \in \argmin_{X_i} |Y - X_i|$ is the nearest-neighbor projection onto $X_{1:n}$.

For $E_{2,n}$, the choice $s=0$ is the direct empirical analogue of the continuum term, while $s=1$ is also consistent with the objective. In practice we use an adaptive weight: initialize $s_0=1/2$, and after epoch $k$, we update $s_k = B(T_k)/\big(A(T_k)+B(T_k)\big)$, where $T_k$ is the current network, so that the two terms in $E_{2,n}$ are balanced, i.e., $s_k A(T_k) = (1 - s_k) B(T_k)$. The fixed choice $s_k\equiv1$ can give more stable training and better performance in practice; we use it for torus sample-complexity scaling and MNIST experiments in Sections~\ref{sec:torus_scaling}, and adaptive weights for all other experiments.

The term $E_{3,n}^{\underline \delta}$ is stronger and easier to compute than the direct analogue of $E_3^{\underline \delta}$: indeed,
$E_{3,n}^{\underline \delta}(T) \geq N^{-1}\sum_{j=1}^N \left( d(T(Y_j), X_{1:n}) - \underline \delta \right)_+$, and $E_{3,n}^{\underline \delta}(\Pi_\M)=0$.
We define 
\begin{equation} \label{def:En}
    E_n:= E_{1,n}+E_{2,n}+E_{3,n}^{\underline\delta} \quad\text{and}\quad E'_n:= E_n+E_{4,n}.
\end{equation}

\subsection{Structure of finite-sample minimizers} \label{sec:finiteNstructure}

We now state a finite-sample counterpart of Theorem~\ref{thm:miniT}, proved in companion work \citep{slepcev2026learning}, describing \emph{every} minimizer of the four-term objective built from the training data alone. Consider the semi-discrete energy
\begin{equation}\label{eqn:semidiscrete}
\widetilde{E}'_n(T):= E_{1,n}(T)+\int_{\calM_{n,\delta}} \!\! |DT(x)(T(x)-x)|^2+\bigl(d(T(x),X_{1:n})-\underline\delta\bigr)_+ +\bigl(|T(x)-x|-\delta\bigr)_+\,dx,
\end{equation}
the analogue of $E'_n$ with the averages over the noised points $Y_j$ replaced by the integral over $\calM_{n,\delta}$, with $s=0$ in $E_{2,n}$. Like $E'_n$, it depends on the data alone. 

\begin{theorem}[Finite-sample manifold recovery; informal]\label{thm:finiteN}
Let $\calM$ and the training sample $X_{1:n}\subset\calM$ be as in Sections~\ref{sec:population} and~\ref{sec:discrete}, with $\calM$ connected, and let $h:=\max_{z\in\calM}d(z,X_{1:n})$ be the covering radius of the sample. Assume that $\delta$ is a sufficiently small fraction of the reach of $\calM$, and that $h\le\underline\delta$, with $\underline\delta$ a sufficiently small fraction of $\delta$. Then for every minimizer $T_n\in C^2(\overline{\calM_{n,\delta}};\R^D)$ of $\widetilde{E}'_n$, the learned manifold $\wh\calM_n:=T_n\bigl(\overline{\calM_{n,\underline\delta}}\bigr)$ is a compact $C^2$ embedded $d$-dimensional submanifold of $\R^D$ containing $X_{1:n}$, $\Pi_\calM$ maps it onto $\calM$, and
\[
d_{\mathcal H}\bigl(\wh\calM_n,\calM\bigr)\lesssim\frac{h^2}{\delta}.
\]
\end{theorem}

For i.i.d.\ samples under Assumption~\ref{assumpmu} one has $h\lesssim(\log n/n)^{1/d}$ with high probability, so $d_{\mathcal H}(\wh\calM_n,\calM)\lesssim\delta^{-1}(\log n/n)^{2/d}$.  Note that pure memorization is impossible for dimensional reasons: a map returning training points has image concentrated on the $0$-dimensional set $X_{1:n}$, while the image of every minimizer is a $d$-dimensional manifold. Moreover, the learned manifold is \emph{more accurate than the data it is learned from}: the sample locates $\calM$ only to within $h$, while every minimizer recovers $\calM$ to within $h^2/\delta\ll h$. Optimizing the objective more accurately thus produces an object closer to the truth than the training data.

\subsection{Two-stage training procedure} \label{sec:two_stage}

Because $E_{2,n}$ is nonconvex and initialization-sensitive, we use two-stage neural training. First, we minimize the simpler pretraining loss
\begin{equation}\label{eqn:losspre}
    E_{\mathrm{pre}}(T) = E_{1,n}(T) + E_{3,n}^0(T).
\end{equation}
With $\underline \delta=0$, this loss maps tubular-neighborhood points toward nearby training samples and by itself would memorize the data. In the second stage, we train with the full objective $E_n$ and $\underline \delta>0$. The full-batch pseudocode is given in Algorithm~\ref{alg:twostage} in Appendix~\ref{app:algorithm-details}; large data sets can be handled with batches of noised points.

\subsection{Iterated projection learning (IPL) for improved resolution}\label{sec:iterated_projection}
In high ambient dimension, $\delta$ must be large enough for finitely many noisy samples to cover the manifold, but a projection trained at this scale may lack fine resolution. We therefore use an iterated version of the two-stage procedure (summarized in Algorithm~\ref{alg:ipl}): train a coarse projection at scale $\delta_1$, apply it to $\delta_1$-noised training samples, and use the projected points to train the next stage at a smaller scale $\delta_2$. Repeating with $\delta_1>\delta_2>\cdots>\delta_K$ gives maps $T_1,\ldots,T_K$ and final generator
\begin{equation}\label{eq:cascade}
  T := T_K\circ T_{K-1}\circ\cdots\circ T_1.
\end{equation}

\subsection{Measuring memorization and manifold recovery} \label{sec:memorization}
We use task-specific metrics to test whether the trained network $T_\theta:\R^D\ria\R^D$ recovers the manifold and distribution, or instead concentrates near the training set. FID~\citep{heusel2017gans} and KID~\citep{binkowski2018demystifying} are useful for image benchmarks, but with ground truth or a large hold-out set we can measure geometry and memorization more directly.

Let $\mu_{T_\theta}:=N^{-1}\sum_{j=1}^N\delta_{T_\theta(Y_j)}$ be the empirical measure of generated samples $\{T_\theta(Y_j)\}_{j=1}^N$. For $r>0$, denote the $r$-neighborhood of the training set by
$B^r_{X_{1:n}}:=\{x\in\R^D:\; d(x,X_{1:n})\le r\}$. Define
$P_r^{\mathrm{true}}:=\mu(B^r_{X_{1:n}})$ and $P_r^{\mathrm{gen}}:=\mu_{T_\theta}(B^r_{X_{1:n}})$.
The \emph{memorization score} is defined by
\begin{equation}\label{eqn:memoscore}
    \Memo(T_\theta):= \frac{\int_{0}^{\infty}\bigl(P_r^{\mathrm{gen}}-P_r^{\mathrm{true}}\bigr)_+\,dr}
     {\int_{0}^{\infty}\bigl(1-P_r^{\mathrm{true}}\bigr)\,dr}.
\end{equation}
The positive gap $(P_r^{\mathrm{gen}}-P_r^{\mathrm{true}})_+$ measures excess generated mass near the training data, across all radii. The normalization makes the score dimensionless: values near $0$ indicate no detectable excess concentration, while value $1$ corresponds to outputting only training points. In practice, $N$ needs to be chosen large enough that Monte Carlo error in $P_r^{\mathrm{gen}}$ is negligible relative to the observed gap.

When $\calM$ and $\Pi_\calM$ are known, we also measure the normalized projection error
\begin{equation}\label{eqn:perr}
    \text{PErr}(T_\theta) := \frac{1}{\mathrm{tr}(\widehat{\Sigma}_n) N}\sum_{j=1}^N|T_\theta(Y_j) - \Pi_\calM(Y_j)|^2 ,
\end{equation}
where $\widehat{\Sigma}_n$ is the unbiased covariance matrix of $X_{1:n}$. Finally, the \emph{manifold identification error (MIE)} measures mean distance to the target manifold:
\begin{equation}\label{eqn:MIE}
    \mathrm{MIE}(T_\theta):= \frac{1}{N}\sum_{j=1}^N|T_\theta(Y_j) - \Pi_\calM\circ T_\theta (Y_j)|.
\end{equation}

\subsection{Generalization bound} \label{sec:genbdd}
The population results show that the ideal objective recovers $\Pi_\calM$ and that projecting $\mu\ast\eta_\delta$ incurs only an $O(\delta^2)$ bias in $\calW_2$. We now give the finite-sample analogue: a generalization bound, in the spirit of distribution-learning estimates, that converts empirical projection error and sampling error into a $2$-Wasserstein guarantee between the generated law $\mu_{T_\theta}$ and the target law $\mu$.

\begin{theorem}\label{thm:wassbdd}
Let $X_i \overset{i.i.d.}{\sim}\mu$ with $i=1,\cdots,n$ and $Y_j \overset{i.i.d.}{\sim} \mu_n\ast \eta_\delta$ with $j=1,\cdots,N$, where $\mu_n$ is the empirical measure of $X_{1:n}$. Assume that $T_\theta:\R^D \mapsto \R^D$ is measurable on $\calM_\delta$ and only depends on $X_{1:n}$. Let $\mu_{T_\theta}$ be the empirical measure of $\{T_\theta(Y_j)\}_{j=1}^N$. Assume Assumption~\ref{assumpmu} holds and there exists some $r_0 \in (0, 1)$ such that $\inf_{|x|\le r_0} \eta(x) > 0$. For every $\alpha>0$, if $\delta \ge C_\alpha n^{-1/d}(\log n)^{1/d}$ for some $C_\alpha>0$, then there exists some $C>0$ such that for every $\beta>1$
\begin{equation*}
    \calW_2(\mu, \mu_{T_\theta}) \lesssim \left\{\begin{array}{ll}
         n^{-1/2}(\log n)^{1/2} \log(\delta^{-1}) + N^{-1/2} (\log N)^{3/4} + \delta^2  + \sqrt{\mathrm{PErr}(T_\theta)}, & d=2, \\
         n^{-1/2}(\log n)^{1/2} \delta^{1-d/2} +  N^{-1/d} (\log N)^{1/d} + \delta^2 + \sqrt{\mathrm{PErr}(T_\theta)}, & d\ge 3,
    \end{array}\right.
\end{equation*}
with probability at least $1 - 2n^{-\alpha}- C N^{-\beta}$.
\end{theorem}

The proof is given in Appendix~\ref{app:wassbdd}. It relies on the following auxiliary estimates, which may be of independent interest.
The first lemma controls the Monte Carlo error from approximating $\Pi_{\mathcal{M}\#}(\mu_n \ast \eta_\delta)$ by $\{\Pi_\calM(Y_j)\}_{j=1}^N$. Because the projected measures live on the $d$-manifold $\calM$, the rate is the covering scale, $(\log N/N)^{1/d}$ for $d\ge 3$.
\begin{lemma}\label{lem:empratewinf}
    Under the same assumptions as Theorem~\ref{thm:wassbdd}, for every $\alpha>0$, if $\delta \ge C_\alpha n^{-1/d}(\log n)^{1/d}$ for some $C_\alpha>0$, there exists some  $C>0$ such that for every $\beta>1$
    \[ \calW_\infty\left( \Pi_{\calM \#}(\mu_n \ast \eta_\delta), \frac{1}{N}\sum_{j=1}^N \delta_{\Pi_{\calM}(Y_j)} \right) \lesssim \left\{\begin{array}{ll}
        N^{-1/2} (\log N)^{3/4}, & d=2, \\
         N^{-1/d} (\log N)^{1/d}, & d\ge 3,
    \end{array}\right. \]
    holds with probability at least $ 1- n^{-\alpha}- C N^{-\beta}$.
\end{lemma}

The second lemma gives the empirical rate for the $\delta$-smoothed $2$-Wasserstein distance between projected measures on $\calM$ (formal version in Appendix Lemma~\ref{lem:w2smoothPi}). For $d\ge3$, it improves on the unsmoothed $n^{-1/d}$ rate; the factor $\delta^{1-d/2}$ reflects the smoothing by the noise kernel.
\begin{lemma}[informal]\label{lem:w2smooth}
Under the same assumptions as Theorem~\ref{thm:wassbdd}, for every $\alpha>0$, there is a constant $C_\alpha>0$ such that with probability at least $1-n^{-\alpha}$
\[\calW_2\Big(\Pi_{\calM \#}(\mu\ast\eta_\delta), \Pi_{\calM \#}(\mu_n\ast\eta_\delta)\Big) \le C_\alpha \left\{\begin{array}{ll}
      n^{-1/2}(\log n)^{1/2} \log(\delta^{-1}), &  d =2, \\
      n^{-1/2}(\log n)^{1/2} \delta^{1-d/2}, & d\ge 3.
  \end{array}\right. \]
\end{lemma}

\section{Experiments}\label{sec:experiments}

Throughout the experiments we optimize the three-term objective $E_n$: we have found that it already gives very accurate results, with learned projections close to orthogonal, so the term $E_{4,n}$ is not needed in practice. We take $\delta\gtrsim n^{-1/d}$, so that $\calM_{n,\delta}$ covers $\calM_\delta$.

\begin{figure}[ht]
    \vspace*{-10pt}
    \centering
        \begin{subfigure}[b]{0.4\textwidth}
\includegraphics[width=\linewidth,height=2.5in,keepaspectratio]{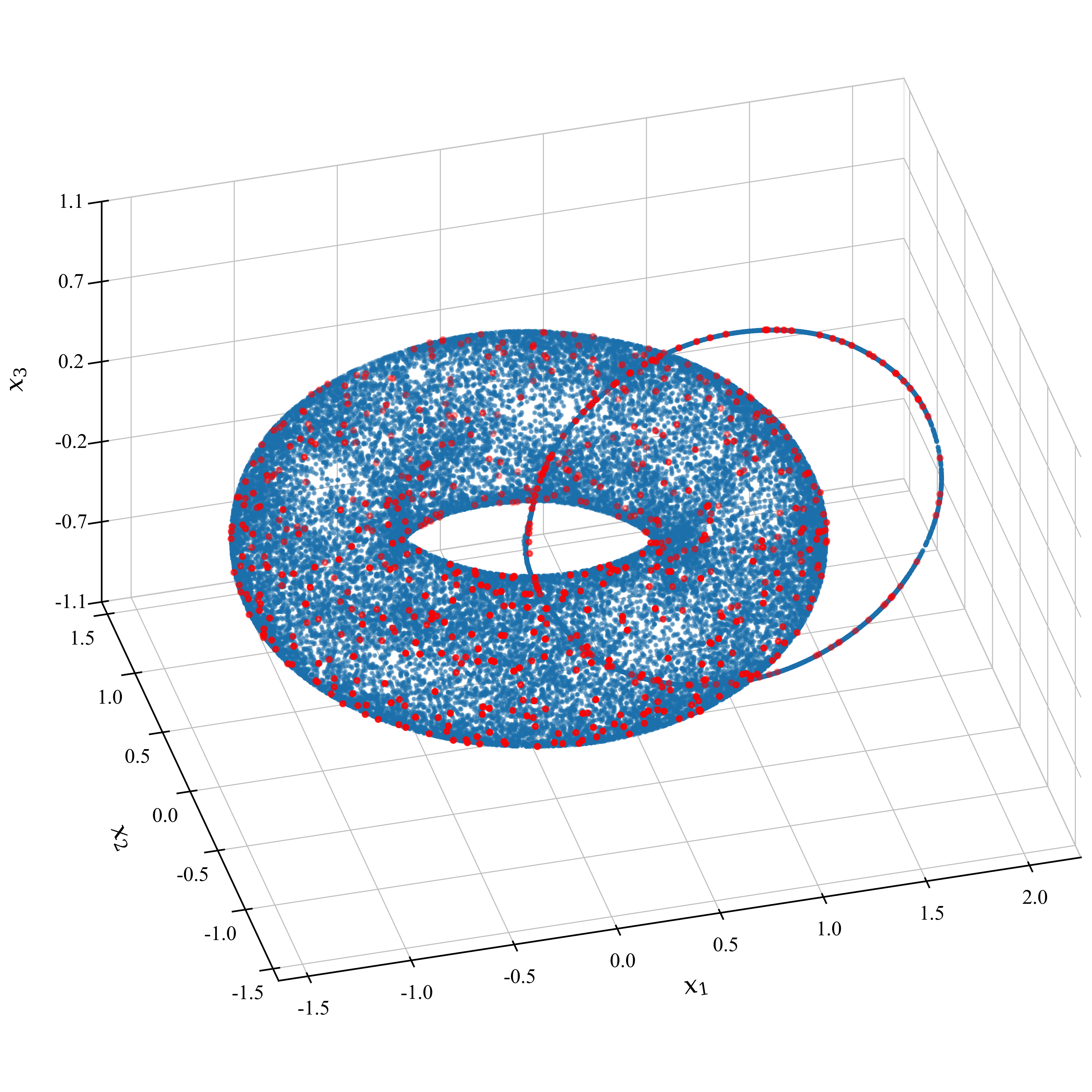}
        \end{subfigure}
        \hspace*{30pt}
    \begin{subfigure}[b]{0.4\textwidth}
\includegraphics[width=\linewidth,height=2.5in,keepaspectratio]{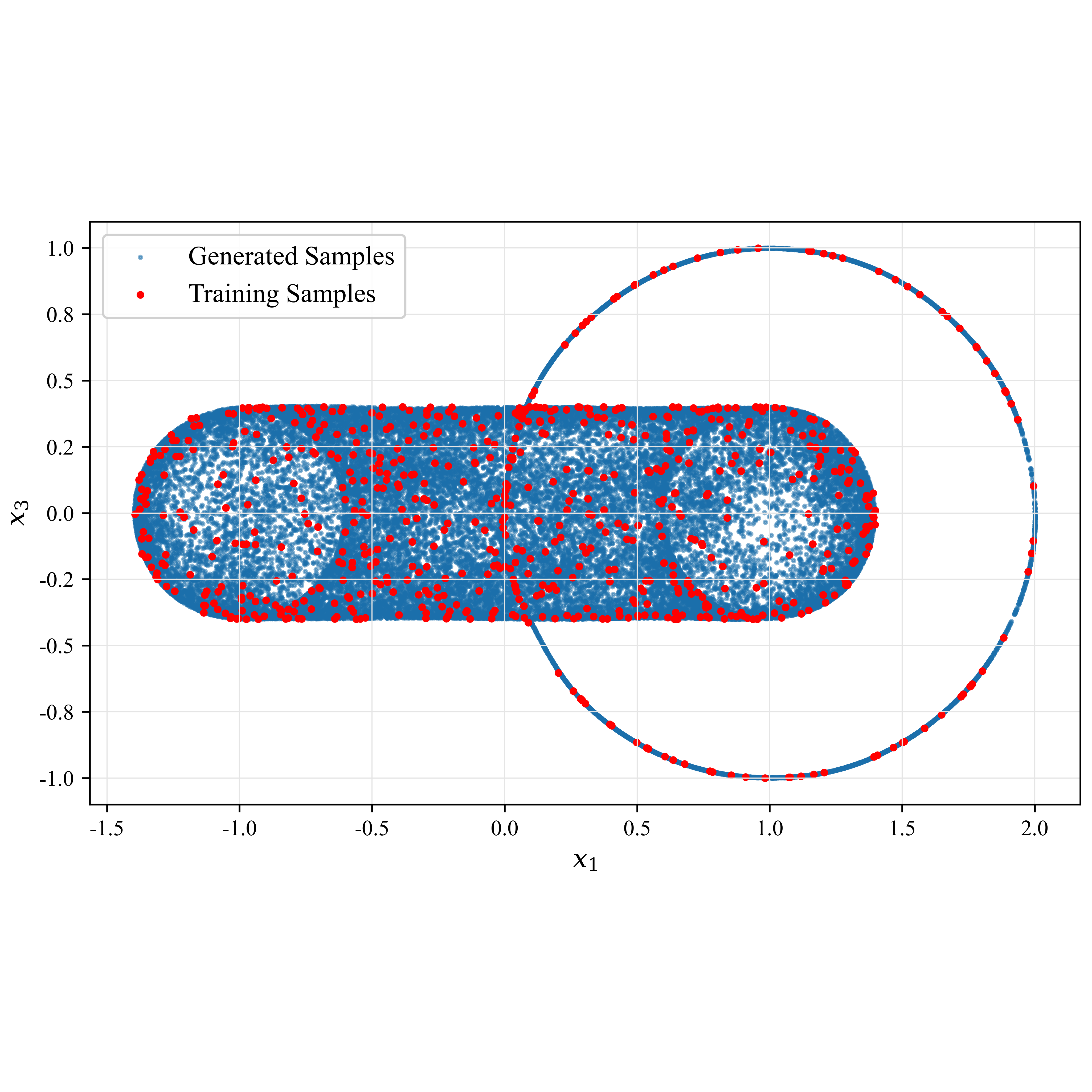}
        \end{subfigure}
        \vspace*{-5pt}
    \caption{Torus linked with a circle in $\R^3$. The model is trained on $600+100$ points (red) and generated $35,000$ samples (blue).}
    \label{fig:torus_circle}
\end{figure}

\exphead{sec:torus_circle}{Manifolds of different dimensions}
We first illustrate the method, see Figure~\ref{fig:torus_circle}, on data supported on a union of manifolds with different intrinsic dimensions. The target set is $\mathcal{M}=\mathcal{T}\cup\mathcal{C}\subset\R^3$, where $\mathcal{T}$ is a torus with major radius $R_1=1$ and tube radius $R_2=0.4$, and $\mathcal{C}$ is a unit circle in the $xz$-plane centered at $(1,0,0)$. This is a challenging geometry because the two components are linked and the reach is small ($0.3$). We sample $600$ points on the torus and $100$ points on the circle, use tubular radius $\delta=0.25$, and train with the same two-stage procedure for a total of $5,000$ epochs.

\exphead{sec:torus_scaling}{Sample complexity scaling for torus identification}
The aim of this experiment is twofold: to illustrate how the required sample size for accurate projection learning scales with the intrinsic dimension of the manifold, and to investigate the effects of the ambient dimension. We consider $\mathbb{T}^d$, the $d$-dimensional torus, embedded in $\R^{2d}$ via the standard product-of-circles embedding. For each $d\in\{1,\dots,5\}$, we first determine the smallest sample size $n^\star(d)$ for which the projection model with $\delta=1$ (reach of $\mathbb{T}^d$) achieves $\mathrm{MIE}<0.02$. While, as expected, the number of points required grows exponentially with $d$, the constant is surprisingly small; we find that $n^\star(d)\approx 9.0 \cdot 3.6^d$.  Next, with the same $n^\star(d)$ we consider the same torus embedded in ambient dimension $D=256$, by adding zero coordinates. To keep the noise per added coordinate at the same scale as the original $2d$-dimensional torus, we set $\delta_d=\sqrt{D/d}$. 
 In Table~\ref{tab:torus_partb}, for $d\in\{1,\cdots,5\}$ we show the scaling of $n^\star$ with dimension, but also the surprising fact that the model recovered the manifold better in high dimension (256D) than in $\R^{2d}$, even though the noising bandwidth $\delta_d$ becomes large and is much larger than the reach of the torus. 
\begin{table}[H]
\centering
\vspace{-4pt}
\caption{Comparing recovery error of $d$-torus in $\R^{2d}$ and in $\R^{256}$ with sample count $n^\star(d)$.}
\label{tab:torus_partb}
\renewcommand{\arraystretch}{1.10}
\setlength{\tabcolsep}{8pt}
\begin{tabular}{rrrcc}
\toprule
$d$ & $n^\star(d)$ & $\delta_d$ & MIE in $\R^{2d} (\delta=1)$ & MIE in $\R^{256} (\delta = \delta_d)$ \\
\midrule
1 & 30   & 16.00 & 0.0042          & 0.0057 \\
2 & 120  & 11.31 & 0.0165           & 0.0118 \\
3 & 450   & 9.24  & 0.0177          & 0.0169 \\
4 & 1616 & 8.00  & 0.0176           & 0.0131 \\
5 & 4829 & 7.16 & 0.0186   & 0.0120 \\
\bottomrule
\end{tabular}
\vspace{-2pt}
\end{table}

\begin{figure}[ht]
\vspace*{-10pt}
  \centering
  \begin{subfigure}[b]{0.45\textwidth}
    \centering
    \includegraphics[width=0.9\textwidth]{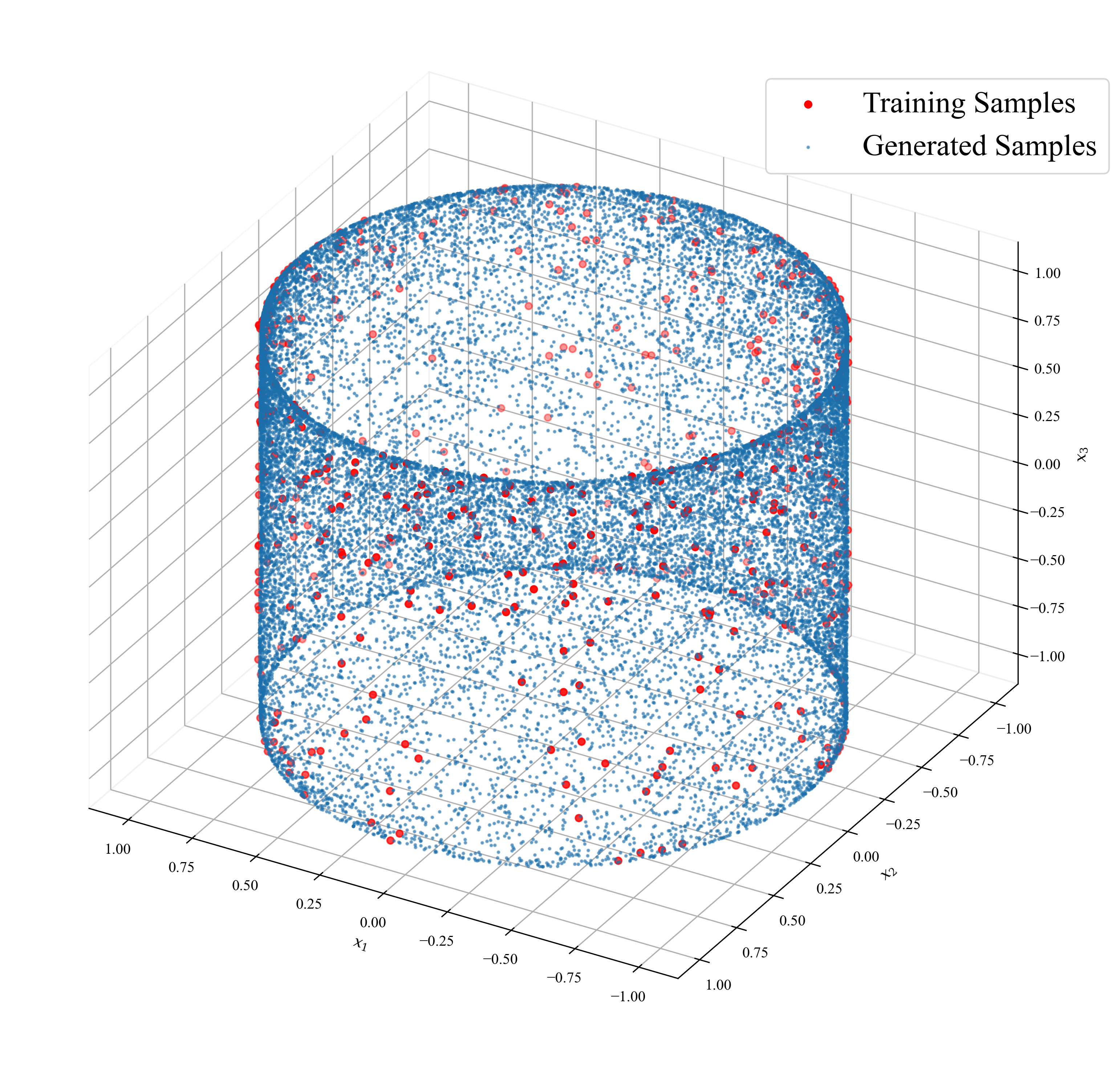}
    \vspace*{-3pt}
    \caption{Generated samples (blue) and training samples (red) in coordinates $(x_1, x_2, x_3)$.}
    \label{fig:torus_3d}
  \end{subfigure}
  \hspace*{25pt}
  \begin{subfigure}[b]{0.42\textwidth}
    \centering
    \includegraphics[width=0.8\textwidth]{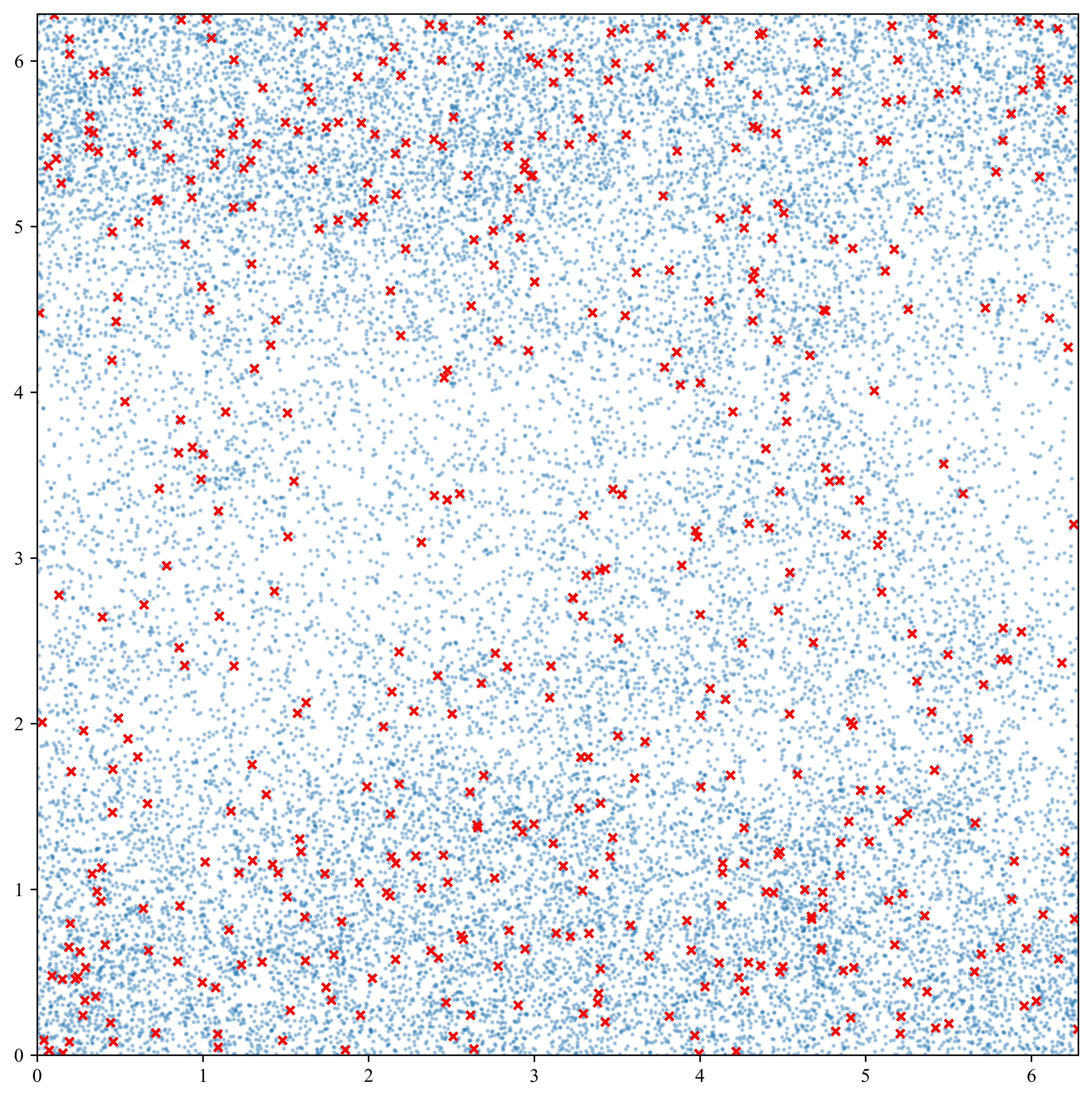}
    \caption{Angle coordinates $(\theta_1, \theta_2)$ of generated samples (blue) and training points (red crosses).}
    \label{fig:torus_angles}
  \end{subfigure}
  \caption{Generating points near embedded torus $S^1 \times S^1 \hookrightarrow \R^{256}$.}
  \label{fig:torus_experiment}
\end{figure}

\exphead{sec:embedded_torus}{Embedded torus}
We next test the method in high ambient dimension over five independent seeds. Let $\mathcal{M}=S^1\times S^1\subset\R^{256}$ be the standard product-of-circles embedding
$(\theta_1,\theta_2)\mapsto (\cos\theta_1,\sin\theta_1,\cos\theta_2,\sin\theta_2,0,\ldots,0)$.
We draw $n=480$ training points from the nonuniform density $\rho(\theta_1,\theta_2)\propto 1+0.5\cos(\theta_2)$, use noising bandwidth $\delta=7$ with 50 noisy copies per point, and train the projection network with the two-stage training Algorithm~\ref{alg:twostage}.
Figure~\ref{fig:torus_experiment} shows that generated samples lie close to the torus. Averaged over seeds, the MIE is $0.006$ and PErr is $0.0006$. By average 1-Wasserstein distances (EMD), the generated samples are closer to true samples than the training set is ($W_1=0.099$ vs.~$0.192$; generated--training $W_1=0.158$, true--true $W_1=0.045$), indicating that the learned projection generates new samples rather than memorizing the training points.

\begin{figure}[ht]
\vspace*{-10pt}
  \centering
  \begin{subfigure}[t]{0.51\textwidth}
    \centering
    \includegraphics[height=4.8cm]{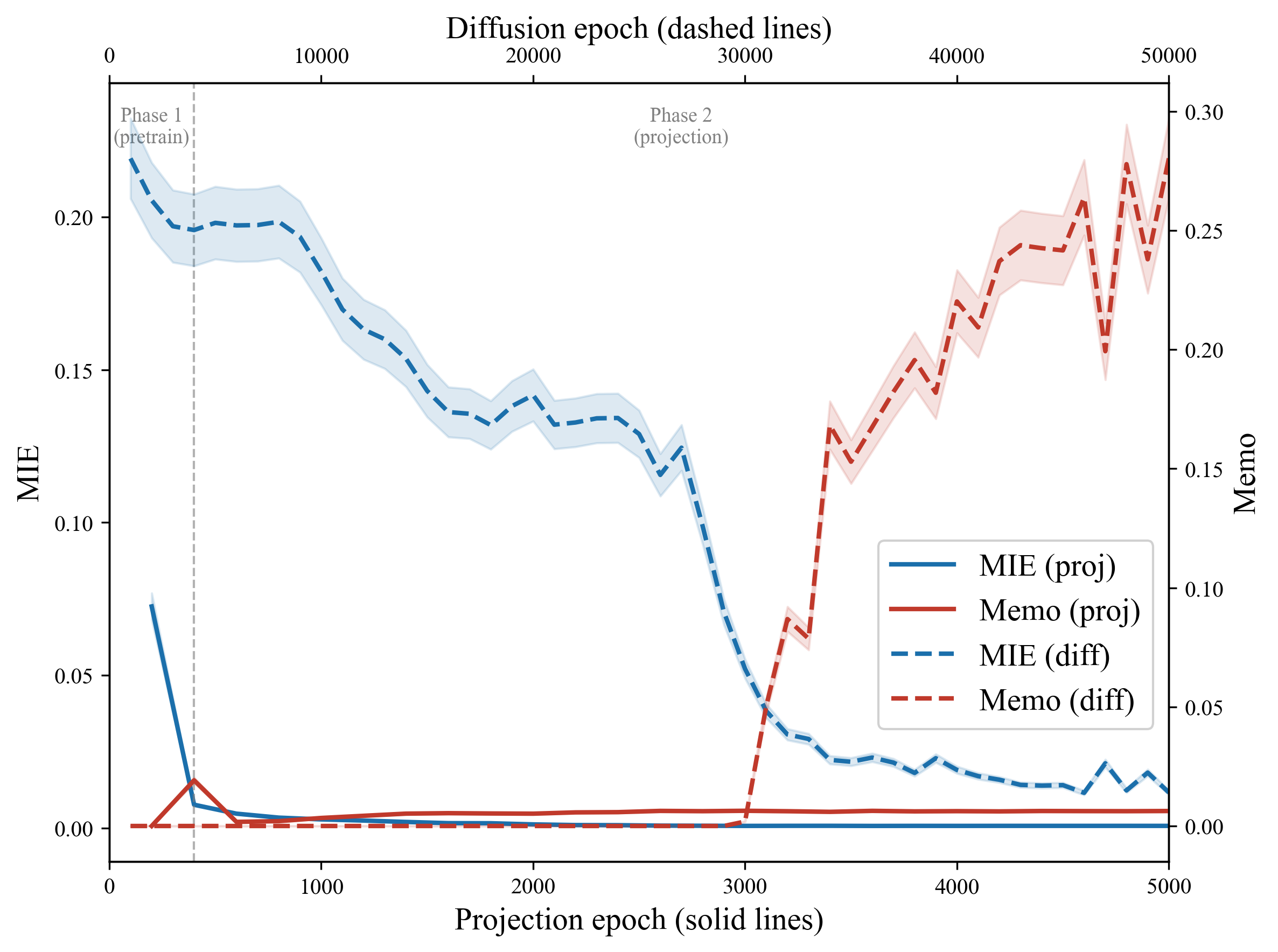}
    \caption{Training metrics over epochs.}
    \label{fig:circle-metrics}
  \end{subfigure}
  \begin{subfigure}[t]{0.42\textwidth}
    \centering
    \includegraphics[height=4.6cm]{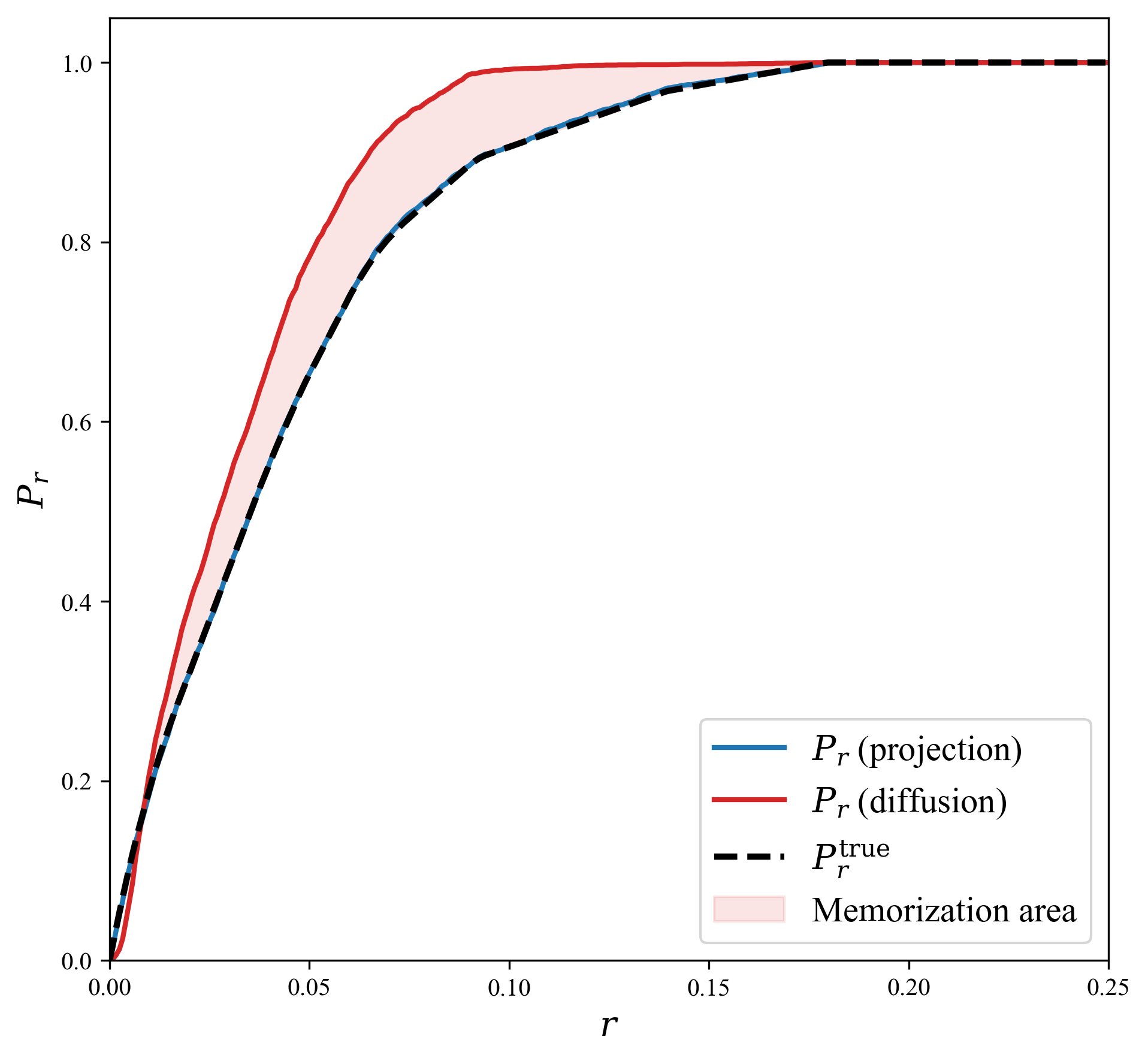}
    \caption{$P_r$ coverage curves.}
    \label{fig:circle-pr}
  \end{subfigure}
  \caption{%
    \textbf{Circle in image space.}
    ($n=30, \, D=64,\, \delta=3$).
    \emph{Left:} MIE (left axis) and Memo (right axis) during
    training; solid = projection model, dashed = diffusion model.
    The vertical dashed line marks the pretrain/projection phase
    boundary.
    \emph{Right:} $P_r$ curves; the projection model closely
    matches $P_r^{\mathrm{true}}$ (black dashed), while the
    diffusion model exhibits memorization (pink area).
  }
  \label{fig:circle-images-main}
\end{figure}

\exphead{sec:exp-circle-images}{\texorpdfstring{$\mathcal S^1$}{S1} in space of images}
We construct a one-dimensional manifold $\mathcal{M}\subset\mathbb{R}^{64}$
of $8\times 8$ images parameterized by angle $t\in[0,2\pi)$. Each image has intensity
\begin{equation}\label{eq:It}
  I_t(x,y)
  = \tfrac{1}{2}
    + \tfrac{1}{2\sqrt{2}}\bigl(x\cos t + y\sin t\bigr),
  \qquad (x,y)\in[-1,1]^2,
\end{equation}
evaluated on a uniform $8\times 8$ grid and vectorized to give
$\gamma(t)\in\mathbb{R}^{64}$.
We draw $n=30$ training samples
$X_i = \gamma(t_i)$ with $t_i\sim\mathrm{Unif}[0,2\pi)$ and set the
tubular-neighborhood radius $\delta=3$.
We compare our two-stage projection MLP, trained for $5,000$ epochs, with a score-based diffusion MLP trained for $50,000$ epochs and sampled by the probability-flow ODE.

Figure~\ref{fig:circle-images-main} summarizes the comparison.
The projection model reaches near-zero manifold identification error (MIE) within roughly $2,000$ epochs and keeps the memorization score $\Memo$ of Section~\ref{sec:memorization} close to zero. The diffusion model reaches comparable MIE only after about $30,000$ epochs, after which memorization rises rapidly. The $P_r$ curve, defined in Section~\ref{sec:memorization}, confirms the distinction: $P_r^{\mathrm{proj}}$ closely tracks the analytic $P_r^{\mathrm{true}}$, while $P_r^{\mathrm{diff}}$ lies above it, with the pink region measuring the memorization gap. Qualitative samples in Figure~\ref{fig:circle-illustration-coefficients-combined} of Appendix~\ref{app:exp-circle-images} show the same pattern.

\begin{figure}[H]
\vspace*{-10pt}
\centering
\begin{subfigure}[b]{0.45\linewidth}
  \centering
  \includegraphics[height=4.5cm]{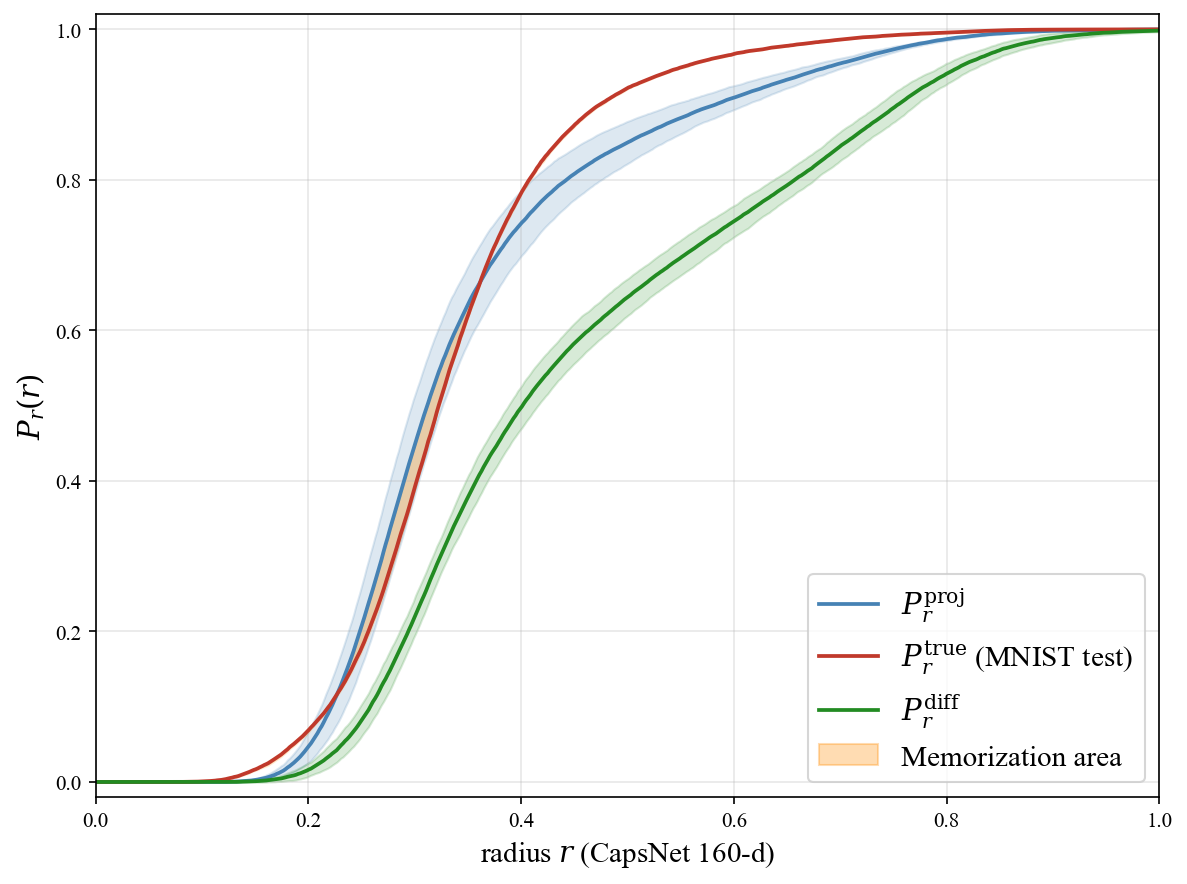}
  \caption{Aggregate $P_r$ curves.}
  \label{fig:agg_pr_test}
\end{subfigure}\hfill
\begin{subfigure}[b]{0.55\linewidth}
  \centering
  \includegraphics[height=4.5cm]{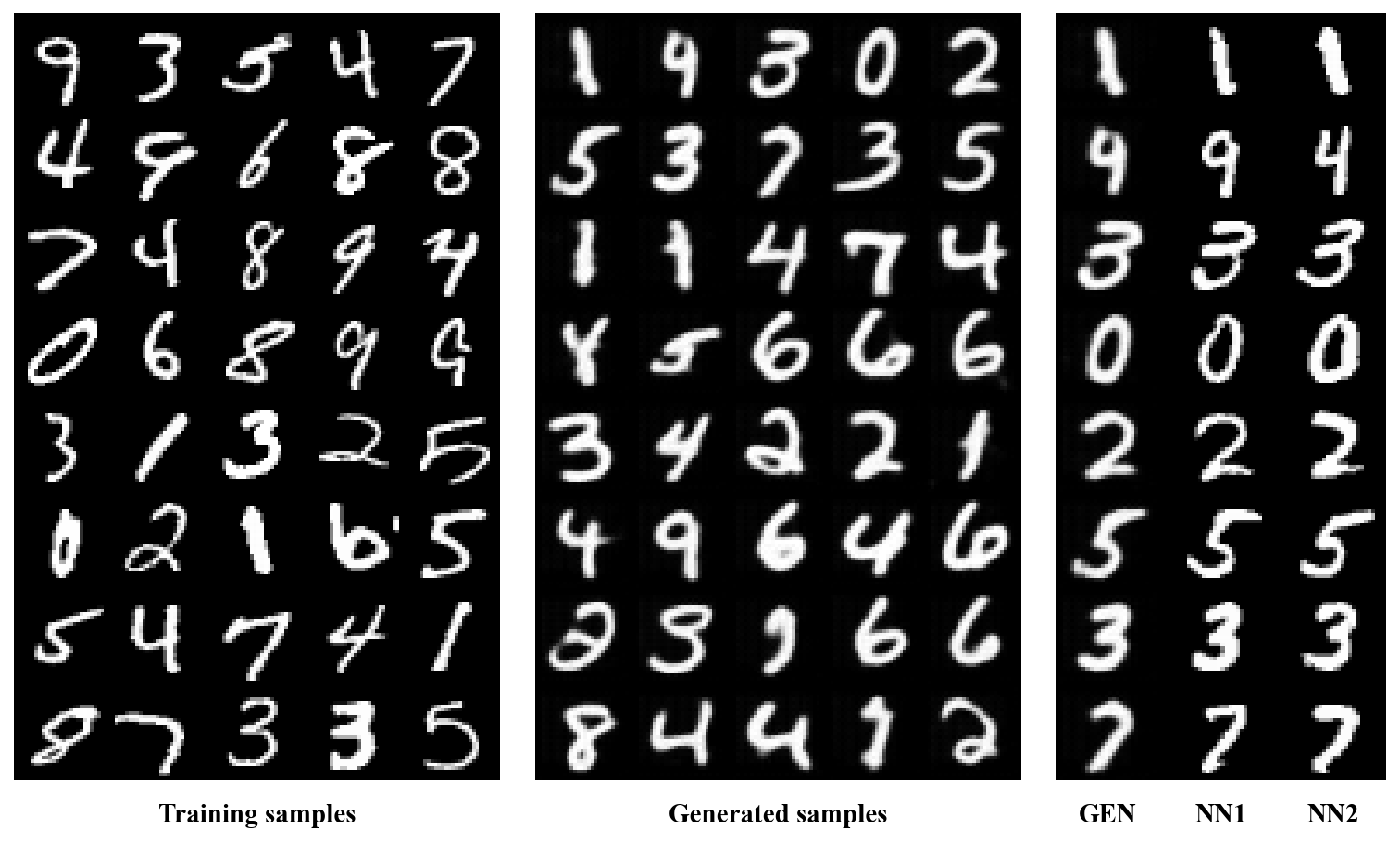}
  \caption{IPL generated images and nearest neighbors.}
  \label{fig:nn_proj_seed0}
\end{subfigure}
\caption{\textbf{$P_r$ curve and generated samples from IPL.}
  \textbf{(a)}~Mean $P_r$ curve (shaded: $\pm$1\,std) over five seeds against the MNIST test reference.
  Blue: IPL ($P_r^{\mathrm{proj}}$); green: VE-SDE diffusion
  ($P_r^{\mathrm{diff}}$); orange: empirical $P_r^{\mathrm{true}}$ using MNIST test samples.
  \textbf{(b)}~\emph{Left:} training MNIST images, \emph{middle}: generated images from IPL, along with the first and second nearest neighbors in the training set (\emph{right}).}
\label{fig:mnist_qual}
\vspace{-8pt}
\end{figure}

\exphead{sec:exp}{MNIST}
We next test whether the projection approach remains applicable to MNIST \citep{lecun1998gradient}. We use the iterated projection procedure of Section~\ref{sec:iterated_projection} with a U-Net architecture. The cascade, architecture, training protocol, and diffusion baseline are described in detail in Appendix~\ref{app:mnist}. For each of five random seeds, we train on $10,000$ MNIST images and use the remaining training images as a held-out reference pool. Both Iterated Projection Learning (IPL) and the VE-SDE diffusion baseline~\citep{song2021scorebased} generate $10,000$ images per seed. Table~\ref{tab:main} reports FID, KID and the $P_r$ memorization score from Section~\ref{sec:memorization} in the 160-dimensional CapsNet~\citep{sabour2017dynamic} embedding space, and $\Memo_{\mathrm{pix}}=N^{-1}\sum_j\mathbf{1}\{d_1(\wh X_j)/d_2(\wh X_j)<\frac{1}{3}\}$ defined by \citet{yoon2023diffusion}, where $d_1,d_2$ are the nearest and second-nearest Euclidean distances to the training set in pixel space. Although $\Memo(\text{VE-SDE})=0$ for each seed, the $P_r$ curve for the diffusion model in Figure~\ref{fig:agg_pr_test} is well below the empirical $P_r^{\mathrm{true}}$ approximated by the MNIST test dataset, indicating that the generated distribution of the diffusion model does not follow the true distribution well. 

\begin{table}[H]
\vspace*{-10pt}
\centering
\caption{\textbf{Multi-seed MNIST evaluation.}
    Each cell reports the across-seed mean with sample standard deviation. FID, KID, Memo, $\Memo_{\mathrm{pix}}$ are computed using the held-out training pool (subsample to $10,000$ samples) as the reference set.}
\label{tab:main}
\renewcommand{\arraystretch}{1.20}
\begin{tabular}{lcccc}
\toprule
\textbf{Model}
   & FID $\downarrow$ & KID $\downarrow$
    & $\Memo\;\downarrow$ & $\Memo_{\mathrm{pix}}\;\downarrow$ \\
\midrule
IPL
   & $0.0565$ \sd{0.0224}
    & $1.40\!\times\!10^{-4}$ \sd{0.50\!\times\!10^{-4}}
   & $0.011$ \sd{0.012}
   & $0.000$ \sd{0.000} \\
VE-SDE
   & $0.0567$ \sd{0.0180}
    & $2.00\!\times\!10^{-4}$ \sd{0.70\!\times\!10^{-4}}
   & $0.000$  \sd{0.000}
   & $0.000$  \sd{0.000} \\
\bottomrule
\end{tabular}
\end{table}

\section{Conclusion}\label{sec:conclusion}

We introduced a variational framework for learning the projection onto the manifold containing the data. The population minimizer of the objective is the orthogonal projection; our distributional analysis quantifies the bias introduced by projecting noised samples; finite-sample Wasserstein bounds relate generation error to projection accuracy and sampling error; and by the companion work \citep{slepcev2026learning} stated in Section~\ref{sec:finiteNstructure}, every minimizer recovers the support to accuracy finer than the covering radius of the data, excluding memorization. We also developed a neural implementation using two-stage training and iterated algorithm to refine the learned map across decreasing noise scales. The implementation uses standard architectures, requires no intrinsic-dimension input, and handles unions of manifolds with different dimensions.  In experiments, the learned maps identify the target manifolds and cover the underlying distribution without collapsing to the training set. 

\paragraph{Open questions and future work.} The current theoretical analysis requires noise below the reach, whereas effective bandwidths in our high-dimensional experiments often exceed it. Understanding this regime could guide the selection of the noise radius and the bandwidth schedule in the iterated algorithm. A further practical goal is to improve training and scale this method to larger image datasets and investigate using the learned projections as a final-stage correction to diffusion models.

\newpage
\begin{ack}
We are grateful to Anton Ullrich for many valuable discussions and to Nick Boffi for advice on training the networks. LC and DS were supported by the NSF grants DMS-2206069 and DMS-2511684. DS also acknowledges the support of the NSF RTG grant DMS-2342349.
\end{ack}

\bibliographystyle{plainnat}
\bibliography{NeurIPS_Final}

@misc{li2024goodscoredoeslead,
  title         = {A Good Score Does not Lead to A Good Generative Model},
  author        = {Sixu Li and Shi Chen and Qin Li},
  year          = {2024},
  archiveprefix = {arXiv},
  eprint        = {2401.04856},
  primaryclass  = {cs.LG},
  url           = {https://arxiv.org/abs/2401.04856}
}

@misc{bergmeister2025projection,
  title         = {A Projection-Based Framework for Gradient-Free and Parallel Learning},
  author        = {Bergmeister, Andreas and Lal, Manish Krishan and Jegelka, Stefanie and Sra, Suvrit},
  year          = {2025},
  archiveprefix = {arXiv},
  eprint        = {2506.05878},
  note          = {arXiv:2506.05878}
}

@book{bishop2011geometry,
  title     = {Geometry of manifolds: Geometry of Manifolds},
  author    = {Bishop, Richard L and Crittenden, Richard J},
  year      = {2011},
  volume    = {15},
  publisher = {Academic press}
}

@misc{elser2026machinelearnable,
  title         = {Machine-learnable Sets},
  author        = {Elser, Veit and Lal, Manish Krishan},
  year          = {2026},
  archiveprefix = {arXiv},
  eprint        = {2606.28947},
  note          = {arXiv:2606.28947},
  primaryclass  = {cs.LG}
}

@misc{shen2026manifold,
  title         = {Manifold Generalization Provably Proceeds Memorization in Diffusion Models},
  author        = {Zebang Shen and Ya-Ping Hsieh and Niao He},
  year          = {2026},
  archiveprefix = {arXiv},
  eprint        = {2603.23792},
  primaryclass  = {cs.LG},
  url           = {https://arxiv.org/abs/2603.23792}
}

@misc{lyu2025resolvingmemorizationempiricaldiffusion,
  title         = {Resolving Memorization in Empirical Diffusion Model for Manifold Data in High-Dimensional Spaces},
  author        = {Yang Lyu and Tan Minh Nguyen and Yuchun Qian and Xin T. Tong},
  year          = {2025},
  archiveprefix = {arXiv},
  eprint        = {2505.02508},
  primaryclass  = {stat.ML},
  url           = {https://arxiv.org/abs/2505.02508}
}

@book{aubin1998some,
  title     = {Some nonlinear problems in Riemannian geometry},
  author    = {Aubin, Thierry},
  year      = {1998},
  publisher = {Springer Science \& Business Media}
}

@unpublished{slepcev2026learning,
  title  = {Variational Approach to Manifold Projection Learning: Analysis and Error Estimates},
  author = {Slep{\v{c}}ev, Dejan and Ullrich, Anton},
  year   = {2026},
  note   = {In preparation}
}

@article{ho2020denoising,
  title   = {Denoising diffusion probabilistic models},
  author  = {Ho, Jonathan and Jain, Ajay and Abbeel, Pieter},
  journal = {Advances in neural information processing systems},
  year    = {2020},
  volume  = {33},
  pages   = {6840--6851}
}

@article{dhariwal2021diffusion,
  title   = {Diffusion models beat gans on image synthesis},
  author  = {Dhariwal, Prafulla and Nichol, Alexander},
  journal = {Advances in neural information processing systems},
  year    = {2021},
  volume  = {34},
  pages   = {8780--8794}
}

@article{sabour2017dynamic,
  title   = {Dynamic routing between capsules},
  author  = {Sabour, Sara and Frosst, Nicholas and Hinton, Geoffrey E},
  journal = {Advances in neural information processing systems},
  year    = {2017},
  volume  = {30}
}

@article{heusel2017gans,
  title   = {Gans trained by a two time-scale update rule converge to a local nash equilibrium},
  author  = {Heusel, Martin and Ramsauer, Hubert and Unterthiner, Thomas and Nessler, Bernhard and Hochreiter, Sepp},
  journal = {Advances in neural information processing systems},
  year    = {2017},
  volume  = {30}
}

@article{song2019generative,
  title   = {Generative modeling by estimating gradients of the data distribution},
  author  = {Song, Yang and Ermon, Stefano},
  journal = {Advances in neural information processing systems},
  year    = {2019},
  volume  = {32}
}

@article{saharia2022photorealistic,
  title   = {Photorealistic text-to-image diffusion models with deep language understanding},
  author  = {Saharia, Chitwan and Chan, William and Saxena, Saurabh and Li, Lala and Whang, Jay and Denton, Emily L and Ghasemipour, Kamyar and Gontijo Lopes, Raphael and Karagol Ayan, Burcu and Salimans, Tim and others},
  journal = {Advances in neural information processing systems},
  year    = {2022},
  volume  = {35},
  pages   = {36479--36494}
}

@article{somepalli2023understanding,
  title   = {Understanding and mitigating copying in diffusion models},
  author  = {Somepalli, Gowthami and Singla, Vasu and Goldblum, Micah and Geiping, Jonas and Goldstein, Tom},
  journal = {Advances in Neural Information Processing Systems},
  year    = {2023},
  volume  = {36},
  pages   = {47783--47803}
}

@article{Hino03diffusion,
  title      = {Small-time {G}aussian behavior of symmetric diffusion
                semigroups},
  author     = {Hino, Masanori and Ram\'irez, Jos\'e{} A.},
  journal    = {Ann. Probab.},
  year       = {2003},
  volume     = {31},
  number     = {3},
  pages      = {1254--1295},
  doi        = {10.1214/aop/1055425779},
  fjournal   = {The Annals of Probability},
  issn       = {0091-1798,2168-894X},
  mrclass    = {60J60 (31C25 58J65)},
  mrnumber   = {1988472},
  mrreviewer = {David\ Nualart},
  url        = {https://doi.org/10.1214/aop/1055425779}
}

@article{Aamari2019nonasymptotic,
  title      = {Nonasymptotic rates for manifold, tangent space and curvature
                estimation},
  author     = {Aamari, Eddie and Levrard, Cl\'ement},
  journal    = {Ann. Statist.},
  year       = {2019},
  volume     = {47},
  number     = {1},
  pages      = {177--204},
  doi        = {10.1214/18-AOS1685},
  fjournal   = {The Annals of Statistics},
  issn       = {0090-5364,2168-8966},
  mrclass    = {62G05 (62C20)},
  mrnumber   = {3909931},
  mrreviewer = {Michael\ Stolz},
  url        = {https://doi.org/10.1214/18-AOS1685}
}

@article{gine2001consistency,
  title   = {On consistency of kernel density estimators for randomly censored data: rates holding uniformly over adaptive intervals},
  author  = {Gin{\'e}, Evarist and Guillou, Armelle},
  journal = {Annales de l'IHP Probabilit{\'e}s et statistiques},
  year    = {2001},
  volume  = {37},
  number  = {4},
  pages   = {503--522}
}

@article{meila2024manifold,
  title    = {Manifold learning: what, how, and why},
  author   = {Meil\u{a}, Marina and Zhang, Hanyu},
  journal  = {Annu. Rev. Stat. Appl.},
  year     = {2024},
  volume   = {11},
  pages    = {393--417},
  doi      = {10.1146/annurev-statistics-040522-115238},
  fjournal = {Annual Review of Statistics and its Application},
  issn     = {2326-8298,2326-831X},
  mrclass  = {62R07 (62-02 62R30)},
  mrnumber = {4742290},
  url      = {https://doi.org/10.1146/annurev-statistics-040522-115238}
}

@article{coifman2006diffusion,
  title      = {Diffusion maps},
  author     = {Coifman, Ronald R. and Lafon, St\'ephane},
  journal    = {Appl. Comput. Harmon. Anal.},
  year       = {2006},
  volume     = {21},
  number     = {1},
  pages      = {5--30},
  doi        = {10.1016/j.acha.2006.04.006},
  fjournal   = {Applied and Computational Harmonic Analysis. Time-Frequency
                and Time-Scale Analysis, Wavelets, Numerical Algorithms, and
                Applications},
  issn       = {1063-5203,1096-603X},
  mrclass    = {60J70 (60J60 68P30)},
  mrnumber   = {2238665},
  mrreviewer = {G\"otz\ Kersting},
  url        = {https://doi.org/10.1016/j.acha.2006.04.006}
}

@article{binkowski2018demystifying,
  title   = {Demystifying mmd gans},
  author  = {Bi{\'n}kowski, Miko{\l}aj and Sutherland, Danica J and Arbel, Michael and Gretton, Arthur},
  journal = {arXiv preprint arXiv:1801.01401},
  year    = {2018}
}

@article{azangulov2024convergence,
  title   = {Convergence of diffusion models under the manifold hypothesis in high-dimensions},
  author  = {Azangulov, Iskander and Deligiannidis, George and Rousseau, Judith},
  journal = {arXiv preprint arXiv:2409.18804},
  year    = {2024}
}

@article{baptista2025memorization,
  title   = {Memorization and regularization in generative diffusion models},
  author  = {Baptista, Ricardo and Dasgupta, Agnimitra and Kovachki, Nikola B and Oberai, Assad and Stuart, Andrew M},
  journal = {arXiv preprint arXiv:2501.15785},
  year    = {2025}
}

@article{bousquet2002bennett,
  title     = {A Bennett concentration inequality and its application to suprema of empirical processes},
  author    = {Bousquet, Olivier},
  journal   = {Comptes Rendus Mathematique},
  year      = {2002},
  volume    = {334},
  number    = {6},
  pages     = {495--500},
  publisher = {Elsevier}
}

@article{Aamari2018tangential,
  title      = {Stability and minimax optimality of tangential {D}elaunay
                complexes for manifold reconstruction},
  author     = {Aamari, Eddie and Levrard, Cl\'ement},
  journal    = {Discrete Comput. Geom.},
  year       = {2018},
  volume     = {59},
  number     = {4},
  pages      = {923--971},
  doi        = {10.1007/s00454-017-9962-z},
  fjournal   = {Discrete \& Computational Geometry. An International Journal
                of Mathematics and Computer Science},
  issn       = {0179-5376,1432-0444},
  mrclass    = {53C23 (55N35 60D05 62E20 62H12 62H25)},
  mrnumber   = {3802310},
  mrreviewer = {Jean-Charles\ Pinoli},
  url        = {https://doi.org/10.1007/s00454-017-9962-z}
}

@article{peyre2018,
  title   = {Comparison between W 2 distance and $\dot{H}^{-1}$ norm, and Localization of Wasserstein distance},
  author  = {Peyre, R{\'e}mi},
  journal = {ESAIM: Control, Optimisation and Calculus of Variations},
  year    = {2018},
  volume  = {24},
  number  = {4},
  pages   = {1489--1501}
}

@article{elser2021learning,
  title   = {Learning without loss},
  author  = {Elser, Veit},
  journal = {Fixed Point Theory and Algorithms for Sciences and Engineering},
  year    = {2021},
  volume  = {2021},
  number  = {12},
  doi     = {10.1186/s13663-021-00697-1}
}

@article{fefferman2020reconstruction,
  title      = {Reconstruction and interpolation of manifolds. {I}: {T}he
                geometric {W}hitney problem},
  author     = {Fefferman, Charles and Ivanov, Sergei and Kurylev, Yaroslav
                and Lassas, Matti and Narayanan, Hariharan},
  journal    = {Found. Comput. Math.},
  year       = {2020},
  volume     = {20},
  number     = {5},
  pages      = {1035--1133},
  doi        = {10.1007/s10208-019-09439-7},
  fjournal   = {Foundations of Computational Mathematics. The Journal of the
                Society for the Foundations of Computational Mathematics},
  issn       = {1615-3375,1615-3383},
  mrclass    = {53C21 (35R01 35R30 41A05 62G05 62R30)},
  mrnumber   = {4156994},
  mrreviewer = {Carlos\ Mudarra},
  url        = {https://doi.org/10.1007/s10208-019-09439-7}
}

@article{garcia2020error,
  title     = {Error estimates for spectral convergence of the graph Laplacian on random geometric graphs toward the Laplace--Beltrami operator},
  author    = {Garc{\'\i}a Trillos, Nicol{\'a}s and Gerlach, Moritz and Hein, Matthias and Slep{\v{c}}ev, Dejan},
  journal   = {Foundations of Computational Mathematics},
  year      = {2020},
  volume    = {20},
  number    = {4},
  pages     = {827--887},
  publisher = {Springer}
}

@article{song2021score,
  title   = {Score-based generative modeling through stochastic differential equations},
  author  = {Song, Yang and Sohl-Dickstein, Jascha and Kingma, Diederik P and Kumar, Abhishek and Ermon, Stefano and Poole, Ben},
  journal = {ICLR},
  year    = {2021}
}

@article{bengio2013representation,
  title     = {Representation learning: A review and new perspectives},
  author    = {Bengio, Yoshua and Courville, Aaron and Vincent, Pascal},
  journal   = {IEEE transactions on pattern analysis and machine intelligence},
  year      = {2013},
  volume    = {35},
  number    = {8},
  pages     = {1798--1828},
  publisher = {IEEE},
  doi       = {10.1109/TPAMI.2013.50}
}

@article{hastie1989principal,
  title     = {Principal curves},
  author    = {Hastie, Trevor and Stuetzle, Werner},
  journal   = {Journal of the American statistical association},
  year      = {1989},
  volume    = {84},
  number    = {406},
  pages     = {502--516},
  publisher = {Taylor \& Francis}
}

@article{belkin2003laplacian,
  title     = {Laplacian eigenmaps for dimensionality reduction and data representation},
  author    = {Belkin, Mikhail and Niyogi, Partha},
  journal   = {Neural computation},
  year      = {2003},
  volume    = {15},
  number    = {6},
  pages     = {1373--1396},
  publisher = {MIT Press}
}

@article{liu2026diffusion_manifold,
  title    = {Improving the Euclidean Diffusion Generation of Manifold Data by Mitigating Score Function Singularity},
  author   = {Zichen Liu and Wei Zhang and Tiejun Li},
  journal  = {NeurIPS 2025},
  year     = {2025},
  arxiv    = {http://arxiv.org/abs/2505.09922},
  language = {en}
}

@article{donoho2003hessian,
  title      = {Hessian eigenmaps: locally linear embedding techniques for
                high-dimensional data},
  author     = {Donoho, David L. and Grimes, Carrie},
  journal    = {Proc. Natl. Acad. Sci. USA},
  year       = {2003},
  volume     = {100},
  number     = {10},
  pages      = {5591--5596},
  doi        = {10.1073/pnas.1031596100},
  fjournal   = {Proceedings of the National Academy of Sciences of the United
                States of America},
  issn       = {0027-8424,1091-6490},
  mrclass    = {62H05 (94A08)},
  mrnumber   = {1981019},
  mrreviewer = {Kotik\ K.\ Lee},
  url        = {https://doi.org/10.1073/pnas.1031596100}
}

@article{lecun1998gradient,
  title     = {Gradient-based learning applied to document recognition},
  author    = {LeCun, Yann and Bottou, L{\'e}on and Bengio, Yoshua and Haffner, Patrick},
  journal   = {Proceedings of the IEEE},
  year      = {1998},
  volume    = {86},
  number    = {11},
  pages     = {2278--2324},
  publisher = {IEEE},
  doi       = {10.1109/5.726791}
}

@article{tenenbaum2000isomap,
  title     = {A global geometric framework for nonlinear dimensionality reduction},
  author    = {Tenenbaum, Joshua B and Silva, Vin de and Langford, John C},
  journal   = {science},
  year      = {2000},
  volume    = {290},
  number    = {5500},
  pages     = {2319--2323},
  publisher = {American Association for the Advancement of Science}
}

@article{roweis2000LLE,
  title   = {Nonlinear Dimensionality Reduction by Locally Linear Embedding},
  author  = {Roweis, Sam T. and Saul, Lawrence K.},
  journal = {Science},
  year    = {2000},
  volume  = {290},
  number  = {5500},
  pages   = {2323-2326},
  doi     = {10.1126/science.290.5500.2323},
  eprint  = {http://www.sciencemag.org/content/290/5500/2323.full.pdf},
  url     = {http://www.sciencemag.org/content/290/5500/2323.abstract}
}

@article{zhang2004LTSA,
  title      = {Principal manifolds and nonlinear dimensionality reduction via
                tangent space alignment},
  author     = {Zhang, Zhenyue and Zha, Hongyuan},
  journal    = {SIAM J. Sci. Comput.},
  year       = {2004},
  volume     = {26},
  number     = {1},
  pages      = {313--338},
  doi        = {10.1137/S1064827502419154},
  fjournal   = {SIAM Journal on Scientific Computing},
  issn       = {1064-8275,1095-7197},
  mrclass    = {65D10 (15A18 65F15 65F50)},
  mrnumber   = {2114346},
  mrreviewer = {Fabio\ Di Benedetto},
  url        = {https://doi.org/10.1137/S1064827502419154}
}

@article{mcdiarmid1989method,
  title     = {On the method of bounded differences},
  author    = {McDiarmid, Colin and others},
  journal   = {Surveys in combinatorics},
  year      = {1989},
  volume    = {141},
  number    = {1},
  pages     = {148--188},
  publisher = {Norwich}
}

@article{massart2000constants,
  title     = {About the constants in Talagrand's concentration inequalities for empirical processes},
  author    = {Massart, Pascal},
  journal   = {The Annals of Probability},
  year      = {2000},
  volume    = {28},
  number    = {2},
  pages     = {863--884},
  publisher = {Institute of Mathematical Statistics}
}

@article{alain2014DAE,
  title     = {What regularized auto-encoders learn from the data-generating distribution},
  author    = {Alain, Guillaume and Bengio, Yoshua},
  journal   = {The Journal of Machine Learning Research},
  year      = {2014},
  volume    = {15},
  number    = {1},
  pages     = {3563--3593},
  publisher = {JMLR. org}
}

@article{Federer1959,
  title    = {Curvature measures},
  author   = {Federer, Herbert},
  journal  = {Trans. Am. Math. Soc.},
  year     = {1959},
  volume   = {93},
  pages    = {418--491},
  doi      = {10.2307/1993504},
  fjournal = {Transactions of the American Mathematical Society},
  issn     = {0002-9947},
  language = {English},
  zbl      = {0089.38402},
  zbmath   = {3146500}
}

@article{debortoli2022convergence,
  title   = {Convergence of denoising diffusion models under the manifold hypothesis},
  author  = {Valentin De Bortoli},
  journal = {Transactions on Machine Learning Research},
  year    = {2022},
  issn    = {2835-8856},
  note    = {Expert Certification},
  url     = {https://openreview.net/forum?id=MhK5aXo3gB}
}

@inproceedings{carlini2023extracting,
  title     = {Extracting training data from diffusion models},
  author    = {Carlini, Nicolas and Hayes, Jamie and Nasr, Milad and Jagielski, Matthew and Sehwag, Vikash and Tramer, Florian and Balle, Borja and Ippolito, Daphne and Wallace, Eric},
  booktitle = {32nd USENIX security symposium (USENIX Security 23)},
  year      = {2023},
  pages     = {5253--5270}
}

@inproceedings{sriperumbudur2012consistency,
  title        = {Consistency and rates for clustering with dbscan},
  author       = {Sriperumbudur, Bharath and Steinwart, Ingo},
  booktitle    = {Artificial Intelligence and Statistics},
  year         = {2012},
  pages        = {1090--1098},
  organization = {PMLR}
}

@inproceedings{kamb2025an,
  title     = {An analytic theory of creativity in convolutional diffusion models},
  author    = {Mason Kamb and Surya Ganguli},
  booktitle = {Forty-second International Conference on Machine Learning},
  year      = {2025}
}

@inproceedings{yoon2023diffusion,
  title     = {Diffusion Probabilistic Models Generalize when They Fail to Memorize},
  author    = {TaeHo Yoon and Joo Young Choi and Sehyun Kwon and Ernest K. Ryu},
  booktitle = {ICML 2023 Workshop on Structured Probabilistic Inference {\&} Generative Modeling},
  year      = {2023},
  url       = {https://openreview.net/forum?id=shciCbSk9h}
}

@inproceedings{song2021scorebased,
  title     = {Score-Based Generative Modeling through Stochastic Differential Equations},
  author    = {Yang Song and Jascha Sohl-Dickstein and Diederik P Kingma and Abhishek Kumar and Stefano Ermon and Ben Poole},
  booktitle = {International Conference on Learning Representations},
  year      = {2021},
  url       = {https://openreview.net/forum?id=PxTIG12RRHS}
}

@inproceedings{sohl2015deep,
  title        = {Deep unsupervised learning using nonequilibrium thermodynamics},
  author       = {Sohl-Dickstein, Jascha and Weiss, Eric and Maheswaranathan, Niru and Ganguli, Surya},
  booktitle    = {International conference on machine learning},
  year         = {2015},
  pages        = {2256--2265},
  organization = {pmlr}
}

@inproceedings{oko2023diffusion,
  title        = {Diffusion models are minimax optimal distribution estimators},
  author       = {Oko, Kazusato and Akiyama, Shunta and Suzuki, Taiji},
  booktitle    = {International Conference on Machine Learning},
  year         = {2023},
  pages        = {26517--26582},
  organization = {PMLR}
}

@inproceedings{vincent2008DAE,
  title     = {Extracting and composing robust features with denoising autoencoders},
  author    = {Vincent, Pascal and Larochelle, Hugo and Bengio, Yoshua and Manzagol, Pierre-Antoine},
  booktitle = {International Conference on Machine Learning},
  year      = {2008},
  pages     = {1096-1103},
  publisher = {Association for Computing Machinery},
  address   = {New York, NY, USA},
  doi       = {10.1145/1390156.1390294},
  location  = {Helsinki, Finland},
  numpages  = {8},
  series    = {ICML '08},
  url       = {https://doi.org/10.1145/1390156.1390294}
}

@inproceedings{kim2019uniform,
  title        = {Uniform convergence rate of the kernel density estimator adaptive to intrinsic volume dimension},
  author       = {Kim, Jisu and Shin, Jaehyeok and Rinaldo, Alessandro and Wasserman, Larry},
  booktitle    = {International Conference on Machine Learning},
  year         = {2019},
  pages        = {3398--3407},
  organization = {PMLR}
}

@inproceedings{tang2024adaptivity,
  title     = {Adaptivity of Diffusion Models to Manifold Structures},
  author    = {Tang, Rong and Yang, Yun},
  booktitle = {Proceedings of The 27th International Conference on Artificial Intelligence and Statistics},
  year      = {2024},
  volume    = {238},
  pages     = {1648--1656},
  publisher = {PMLR},
  editor    = {Dasgupta, Sanjoy and Mandt, Stephan and Li, Yingzhen},
  month     = {02--04 May},
  series    = {Proceedings of Machine Learning Research},
  url       = {https://proceedings.mlr.press/v238/tang24a.html}
}

@inproceedings{stanczuk2024diffusion_models_intrinsic_dimension,
  title     = {Diffusion Models Encode the Intrinsic Dimension of Data Manifolds},
  author    = {Stanczuk, Jan Pawel and Batzolis, Georgios and Deveney, Teo and Sch\"{o}nlieb, Carola-Bibiane},
  booktitle = {Proceedings of the 41st International Conference on Machine Learning},
  year      = {2024},
  volume    = {235},
  pages     = {46412--46440},
  publisher = {PMLR},
  editor    = {Salakhutdinov, Ruslan and Kolter, Zico and Heller, Katherine and Weller, Adrian and Oliver, Nuria and Scarlett, Jonathan and Berkenkamp, Felix},
  month     = {21--27 Jul},
  series    = {Proceedings of Machine Learning Research},
  url       = {https://proceedings.mlr.press/v235/stanczuk24a.html}
}

@inproceedings{permenter2024interpreting,
  title     = {Interpreting and Improving Diffusion Models from an Optimization Perspective},
  author    = {Permenter, Frank and Yuan, Chenyang},
  booktitle = {Proceedings of the 41st International Conference on Machine Learning (ICML)},
  year      = {2024}
}

@inproceedings{chang2017projection,
  title     = {One Network to Solve Them All --- Solving Linear Inverse Problems Using Deep Projection Models},
  author    = {Chang, J. H. Rick and Li, Chun-Liang and P{\'o}czos, Barnab{\'a}s and Vijaya Kumar, B. V. K. and Sankaranarayanan, Aswin C.},
  booktitle = {Proceedings of the IEEE International Conference on Computer Vision (ICCV)},
  year      = {2017},
  pages     = {5888--5897}
}

@inproceedings{somepalli2023diffusion,
  title     = {Diffusion art or digital forgery? investigating data replication in diffusion models},
  author    = {Somepalli, Gowthami and Singla, Vasu and Goldblum, Micah and Geiping, Jonas and Goldstein, Tom},
  booktitle = {Proceedings of the IEEE/CVF conference on computer vision and pattern recognition},
  year      = {2023},
  pages     = {6048--6058}
}

@inproceedings{rombach2022high,
  title     = {High-resolution image synthesis with latent diffusion models},
  author    = {Rombach, Robin and Blattmann, Andreas and Lorenz, Dominik and Esser, Patrick and Ommer, Bj{\"o}rn},
  booktitle = {Proceedings of the IEEE/CVF conference on computer vision and pattern recognition},
  year      = {2022},
  pages     = {10684--10695}
}

@inproceedings{potaptchik2025linearconvergencediffusionmodels,
  title     = {Linear Convergence of Diffusion Models Under the Manifold Hypothesis},
  author    = {Potaptchik, Peter and Azangulov, Iskander and Deligiannidis, George},
  booktitle = {Proceedings of Thirty Eighth Conference on Learning Theory},
  year      = {2025},
  volume    = {291},
  pages     = {4668--4685},
  publisher = {PMLR},
  editor    = {Haghtalab, Nika and Moitra, Ankur},
  month     = {30 Jun--04 Jul},
  series    = {Proceedings of Machine Learning Research},
  url       = {https://proceedings.mlr.press/v291/potaptchik25a.html}
}

@inproceedings{yuhan2025taking,
  title     = {Taking a Big Step: Large Learning Rates in Denoising Score Matching Prevent Memorization},
  author    = {Wu, Yu-Han and Marion, Pierre and Biau, G\'erard and Boyer, Claire},
  booktitle = {Proceedings of Thirty Eighth Conference on Learning Theory},
  year      = {2025},
  volume    = {291},
  pages     = {5718--5756},
  publisher = {PMLR},
  editor    = {Haghtalab, Nika and Moitra, Ankur},
  month     = {30 Jun--04 Jul},
  series    = {Proceedings of Machine Learning Research}
}

@inproceedings{bonnaire2026momorization,
  title     = {Why Diffusion Models Don{\textquoteright}t Memorize:  The Role of Implicit Dynamical Regularization in Training},
  author    = {Tony Bonnaire and Rapha{\"e}l Urfin and Giulio Biroli and Marc Mezard},
  booktitle = {The Thirty-ninth Annual Conference on Neural Information Processing Systems},
  year      = {2026},
  url       = {https://openreview.net/forum?id=BSZqpqgqM0}
}

@inproceedings{kadkhodaie2024generalization,
  title     = {Generalization in diffusion models arises from geometry-adaptive harmonic representations},
  author    = {Zahra Kadkhodaie and Florentin Guth and Eero P Simoncelli and St{\'e}phane Mallat},
  booktitle = {The Twelfth International Conference on Learning Representations},
  year      = {2024}
}

\newpage
\appendix

\section{Technical appendices and supplementary material}

\subsection{Theoretical results}\label{app:theoretical}
\subsubsection{Notation}

\begin{longtable}{>{\arraybackslash}p{0.16\textwidth} p{0.76\textwidth}}
$\mathcal{M}$  & Compact manifold without boundary embedded in $\mathbb{R}^D$. 
Riemannian metric on $\mathcal{M}$ is the one inherited from $\mathbb{R}^D$. \\

$d$  & Intrinsic dimension of $\mathcal{M}$. \\
$D$  & Ambient dimension of $\mathcal{M}$. \\
$\calM_\delta$ & $\delta$-tubular neighborhood around $\calM$. \\

$\mathrm{Vol}(A)$  & Volume of $A \subset \mathcal{M}$ with respect to the Riemannian volume form. \\

$d_{\mathcal{M}}(x,y)$ 
& Geodesic distance between points $x,y\in\mathcal{M}$. \\

$B_{\mathcal{M}}(x,r)$ 
& Ball in $\mathcal{M}$ with respect to geodesic distance on $\mathcal{M}$. \\

$B_{\R^D}(x,r)$ 
& Ball in $\mathbb{R}^D$ of radius $r$ centered at $x$. \\

$\mathrm{diam}(\mathcal{M})$ & Geodesic diameter of $\mathcal{M}$. \\

$\mu$ & Probability measure supported on $\mathcal{M}$ describing the data distribution. \\

$\rho$  & Density of $\mu$ with respect to the Riemannian volume form on $\mathcal{M}$. \\

$\rho_0$  & Positive constant such that $\rho_0^{-1}\le \rho\le \rho_0$ on $\mathcal{M}$. \\

$\eta$ & Smooth function supported on $[0, 1]$ satisfying $\int_{\R^D} \eta(|\xx|) \,d\xx =1$. The associated noise kernel with width $\delta$ is  $\eta_\delta(x):=\delta^{-D}\eta(|x| / \delta)$. \\

$r_0$ & Positivity radius of the kernel $\eta$ such that $\inf_{|x|\le r_0}\eta(x)>0$. \\

$X_{1:n}$ & Point cloud $X_{1:n}=\{X_1,\ldots,X_n\}\subset\mathcal{M}$ drawn from the distribution $\mu$. \\

$Y_{1:N}$ & Point cloud $Y_{1:N}=\{Y_1,\ldots,Y_N\}\subset\mathcal{M}_\delta$ drawn from the distribution $\mu \ast \eta_\delta$. \\

$\mu_n$  & Empirical measure associated with the sample $X_{1:n}$. \\

$r_\calM$ & Reach of the manifold $\mathcal{M}$. \\

$\omega_d$ &  Volume of unit ball in $\R^d$. \\

$i_0$ & Injectivity radius of $\calM$ endowed with the induced Riemannian metric. \\ 

$K$ & Maximum of the absolute value of sectional curvature of $\calM$. \\

$\kappa_{\mathrm{max}}$ & Maximum absolute principal curvature of $\calM$. \\

$C_{\mathrm{sub}}$ & Generic constant depending on the quantities listed in its subscript; values may differ between occurrences. 

\end{longtable}

\subsubsection{Proof of Theorem~\ref{thm:miniT}}\label{apx:proofminiT}

\begin{theorem}[Theorem~\ref{thm:miniT}]
Assume $0<\delta < r_\M/3$ and $\underline \delta < \delta/2$,
    if $T \in C^\infty(\overline{\calM}_\delta; \R^D)$ is a minimizer of $E(T)$, then the range of $T$ is $\calM$, i.e. ~$T(\overline{\calM}_\delta)=\calM$. Moreover, if $T$ is also a minimizer of $E'(T)$, then it is the orthogonal projection onto the manifold $\calM$, i.e.~$T=\Pi_\calM$.
\end{theorem}
\begin{proof}[Sketch of proof]
    First, we note that %
    $\Pi_\calM$ is a minimizer of both $E$ and $E'$ with $E(\Pi_\calM)=E'(\Pi_\calM)=0$. Thus for all minimizers of $E$, we immediately have $T(x) = x$ for $x\in \calM$, $\mathrm{range}(T)\subset \overline{\calM}_{\underline \delta}$ is connected and 
    \begin{equation}\label{eqn:DTcon}
        DT(x)(T(x) - x) =0 \q \text{ for all } x\in \calM_\delta.
    \end{equation}
    Using \eqref{eqn:DTcon}, one can show that $T$ is constant on the line connecting $x$ and $T(x)$. More precisely, we can show 
    \begin{equation}\label{eqn:rayprop}
        T(x + s(T(x) - x)) = T(x)
    \end{equation}
    for all $s \in \R$ such that the line segment from $x$ to $x + s(T(x)-x)$ is contained in $\overline{\calM}_\delta$. Thus $T$ is idempotent by \eqref{eqn:rayprop}.
    
    Furthermore, the rank of $DT$ is constant in $\calM_\delta$. Then it follows from the constant rank theorem that $\mathrm{range}(T)$ is a smooth embedded submanifold. Moreover, we can prove that the level sets of $T$ are nonempty and star-shaped. That is, for any $y \in \mathrm{range}(T)$, there exists some $ x\in \overline{\calM}_\delta \setminus \mathrm{range}(T)$ such that $T(x) = y$, and $T^{-1}(y)$ consists of lines connecting $x$  and $y$ in $\calM_\delta$ and hence is star-shaped with center $y$. 
    
    Therefore, the continuous map $H(x, t):=(1-t)x + tT(x)$ is a deformation retract of $\calM_\delta$ onto $\mathrm{range}(T)$. This implies that $\calM_\delta$ and $\mathrm{range}(T)$ are homotopy equivalent. Similarly, using the orthogonal projection $(1-t)x+t\Pi_\calM(x)$ is a deformation retract from $\calM_\delta$ onto $\calM$, one can find $\calM$ and $\mathrm{range}(T)$ have the same dimension. Since $\calM$ is closed and $\mathrm{range}(T)$ is connected, we can conclude that $\mathrm{range}(T)=\calM$ as desired.

    On the other hand, if $T$ is also a minimizer of $E'$, we have $E_4(T)=0$ and hence $|T(x) - x| \le \delta$ for all $x\in \overline{\calM}_\delta$. Then it is straightforward to obtain $T(x)=\Pi_\calM(x)$ for $x\in \partial\calM_\delta$. Since $\calM$ is compact, we can apply the ray property \eqref{eqn:rayprop} to show that $T(x)=\Pi_\calM(x)$ for all $x\in \calM_\delta$ as well. This completes the argument.
\end{proof}

\subsubsection{Proofs in Section~\ref{sec:errorbound}}\label{app:errorbdd}

Let $T_x\calM$ be the tangent space to $\calM$ at $x\in\calM$, and $N_x\calM$ be the orthogonal complement of $T_x\calM$ in $\R^D$. For any $v \in T_x\calM$ and $N\in N_x\calM$, define the shape operator $S_N|_x:T_x\calM\ria T_x\calM$ by 
    \[ S_N|_x(v) := -\Pi_{T_x\calM}(\bar{\nabla}_v \bar{N}), \]
    where $\bar{N}$ is a local extension of $N$ such that $\bar{N}(p)\in N_p\calM$ for all $p\in \calM$ near $x$. Since the shape operator $S_N|_x$ is self-adjoint, it has real eigenvalues $\{\kappa_i^N(x)\}_{i\in [d]}$. Define the global curvature bound by
    \[ \kappa_{\max}:=\sup_{x\in \calM} \sup_{\substack{N\in N_x\calM \\ \|N\|=1}}\|S_N|_x\|_{\mathrm{op}} = \sup_{\substack{x\in \calM, \; i \in [d] \\ N\in N_x\calM,\; \|N\|=1 }} |\kappa_i^N(x)|. \]
Moreover, we have $\kappa_{\max} \le 1/ r_\calM$ by Theorem 4.18 in \citet{Federer1959}.

Let $\bar{\mu}_\delta := \Pi_{\calM \#} (\mu \ast \eta_\delta)$ be the pushforward of the noised measure $\mu \ast \eta_\delta$ by the projection map $\Pi_\calM$. We can show that $\bar{\mu}_\delta$ is absolutely continuous with respect to the volume form of $\calM$, and denote its density by $\bar{\rho}_\delta$. The following theorem gives a bias estimate for $\bar{\rho}_\delta$ compared to the original density $\rho$. Then we can convert the $L^\infty$ bound on the density error into the $\calW_2$ bound in Theorem~\ref{thm:errorbdd}.

\begin{theorem}[Theorem~\ref{thm:rhoerrbdd}]
    Under Assumption~\ref{assumpmu}, we have the following density bias estimate for $\bar{\rho}_\delta$:
    \begin{equation}\label{eqn:rhoerrbdd}
        \|\rho - \bar{\rho}_\delta\|_{L^\infty(\calM)} \le \frac{d}{2D} \delta^2 \|\nabla_\calM^2 \rho\|_{\op}  + O(\delta^3),
    \end{equation}
    where $\nabla_\calM^2 \rho$ is the Riemannian Hessian of $\rho$ on $\calM$ and $\|\nabla_\calM^2 \rho\|_{\op}:=\sup_{x\in \calM} \|\nabla_\calM^2 \rho(x)\|_{\mathrm{op}}$.

    Moreover, suppose that $\delta \le C / d $ with $C$ independent of $d$ and $D$, we have
    \begin{equation}\label{eqn:rhoerrbdd2}
        \|\rho - \bar{\rho}_\delta\|_{L^\infty(\calM)} \le
        C_{\calM, \rho}\frac dD\,\delta^2,
    \end{equation}
    where
    $C_{\calM, \rho}$ is a constant depending on the geometry of $\calM$ and $\rho$, and is independent of $d$ and $D$.
\end{theorem}
\begin{proof}
    In the following, we divide the proof into five steps, in the first step we derive the density function $\bar{\rho}_\delta$, in the second step we rewrite it in a graph chart, in the third step we expand some integrand factors, in the fourth step we derive the second-order expansion to get \eqref{eqn:rhoerrbdd}, and finally we bound the higher-order terms to get \eqref{eqn:rhoerrbdd2}.

    \paragraph{1. Derive projected noisy density formula.}
    By definition, the pushforward measure $\bar{\mu}_\delta = \Pi_{\calM \#}(\mu \ast \eta_\delta)$ satisfies, for any measurable $V\subset \calM$,
    \begin{equation}\label{eqn:barmusigma}
        \bar{\mu}_\delta(V)= \int_V \bar{\rho}_\delta(\zz) \, d\mathrm{vol}_\calM(\zz) =\int_{\Pi_\calM^{-1}(V)} \rho\ast \eta_\delta(\yy) \,d\yy.
    \end{equation}

    Consider any local chart $(U, \psi)$ of $\calM$ with
    \[ \psi: U \subset \mathbb{R}^d \ria V \subset \calM\subset \mathbb{R}^D, \qquad \uu=(u^1, \cdots, u^d) \mapsto \psi(\uu).   \]

    Then the vectors $\psi_{,i}:=\frac{\partial \psi}{\partial u^i} (\uu) $ with $i=1,\cdots,d$ span the tangent space $T_{\psi(u)}\calM$. Let $\{N_j(\uu)\}_{j=1,\cdots, D-d}$ be a smooth orthonormal normal frame on $N\calM$ along $\psi(U)$.

    Use the chart coordinates $\uu=(u^1, \cdots, u^d)$ on the base and normal coordinates $\tt=(t_1,\dots,t_{D-d})$ on the normal fiber. We can define the normal-coordinate map as follows:
    \[\Phi(\uu, \tt):=\psi(\uu)+\sum_{j=1}^{D-d} t_j N_j(\uu) \in  \mathbb{R}^D,\]
    and $(\uu, \tt)$ can be regarded as a coordinate system for points in a tubular neighborhood.

    In order to perform the change-of-variables $\yy=\Phi(\uu, \tt)$ for $\yy \in \Pi_\calM^{-1}(V)$, we need to compute the Jacobian determinant $\det D\Phi(\uu, \tt)$. By straightforward computation, we have
    \[D\Phi(u^1, \cdots, u^d, t_1,\dots,t_{D-d}) = \left [ \left( \frac{\partial \psi}{\partial u^i} (\uu) +\sum_{j=1}^{D-d} t_j \frac{\partial N_j}{\partial u^i}(\uu)\right)_{i=1,\cdots, d}, \, \left( N_j(\uu) \right)_{j=1,\cdots,D-d}  \right].\]
    Decomposing $\frac{\partial N_j}{\partial u^i}$ into its tangential and normal constituents gives
    \[\frac{\partial N_j}{\partial u^i} = -S_{N_j}\left( \frac{\partial \psi}{\partial u^i} \right) + \sum_{\ell=1}^{D-d}\omega_{ij\ell} N_\ell, \]
    where $\omega_{ij\ell}=\frac{\partial N_j}{\partial u^i} \cdot N_\ell$ are the normal-connection coefficients.

    Denote $t =  |\tt| =\sqrt{\sum_{j=1}^{D-d} t_j^2}$ and $N=t^{-1}\sum_{j=1}^{D-d}t_j N_j$. We can rewrite
    \[ D\Phi(\uu, \tt)=\Bigg[ \left( \frac{\partial \psi}{\partial u^i} (\uu) -t S_N\left(\frac{\partial \psi}{\partial u^i}\right) + \sum_{j,\ell =1}^{D-d}t_j \omega_{ij\ell} N_\ell \right)_{i=1,\cdots, d}, \, \left( N_j(\uu) \right)_{j=1,\cdots,D-d}  \Bigg]. \]

    According to elementary determinant properties, one can find
    \[\det D\Phi(\uu, \tt) = \det\left( \left [ \left( \frac{\partial \psi}{\partial u^i} (\uu) -t S_N\left(\frac{\partial \psi}{\partial u^i}\right) \right)_{i=1,\cdots, d}, \left( N_j(\uu) \right)_{j=1,\cdots,D-d}  \right]\right):= \det B.\]
    By the orthogonality $\frac{\partial \psi}{\partial u^i} \perp N_j$ for any $i,j$, we can compute
    \[ |\det D\Phi(\uu, \tt)| = |\det B| = \sqrt{\det (B^T B)} = \det (I_d-t S_N) \sqrt{\det (D \psi^T D \psi)},\]
    where $\det (I_d-t S_N)>0$ since we have $t \le \delta \le r_\calM / 2  \le (2 \kappa_{\max})^{-1}$.

    Therefore, applying change-of-variables $\yy=\Phi(\uu, \tt)$ and $\zz = \psi(\uu) $ gives
    \begin{align}
              & \int_{\Pi_\calM^{-1}(V)} \rho\ast \eta_\delta(\yy) \,d\yy \notag                                                                                       \\
        = \;  & \int_U \int_{\R^{D-d}} \rho\ast \eta_\delta \left( \psi(\uu)+\sum_{j=1}^{D-d} t_j N_j(\uu) \right) |\det D\Phi(\uu, \tt)| \, d \tt d\uu  \notag        \\
        = \;  & \int_U \int_{\R^{D-d}} \rho\ast \eta_\delta \left( \psi(\uu)+ tN(\uu)\right) \det (I_d-t S_N) \sqrt{\det (D \psi^T D \psi)} \, d \tt d\uu \notag       \\
        =  \; & \int_V \int_{\R^{D-d}} \rho\ast \eta_\delta \left( \zz+ tN|_z \right) \det (I_d-t S_N|_{z})\, d \tt d \mathrm{vol}_\calM(\zz), \label{eqn:barmusigma2}
    \end{align}
    where $N|_z$ and $S_N|_{z}$ denote the normal displacement vector and shape operator at $\zz=\psi(\uu)$, respectively.

    Combining \eqref{eqn:barmusigma} and \eqref{eqn:barmusigma2} yields the noisy density formula:
    \begin{align}
        \bar{\rho}_\delta(\zz) & = \int_{\R^{D-d}} \rho\ast \eta_\delta \left( \zz+ tN|_z \right) \det (I_d-t S_N|_{z})\, d \tt \notag                                                                                                        \\
                               & = \int_\calM \int_{\R^{D-d}} \frac{1}{\delta^D} \eta \left( \frac{\left|\zz+ tN|_z - \xx \right|}{\delta} \right) \det (I_d-t S_N|_{z}) \rho(\xx)\, d \tt d \mathrm{vol}_\calM(\xx)  \label{eqn:barrhosigma}
    \end{align}

    \paragraph{2. Express projected noised density in a graph chart.}
    Next we proceed to estimate the error bound for $\bar{\rho}_\delta(\zz)-\rho(\zz)$. After a rigid motion of $\mathbb R^{D}$, we can assume that
    \begin{enumerate}[leftmargin=*, label=\roman*.]
        \item $\zz = \mathbf{0}$ and $T_0 \calM =\mathrm{span}\{e_1, \cdots, e_d\}$, where $[e_1, \cdots, e_D]$ are the coordinate unit vectors in $\mathbb{R}^D$. $\{N_j\}_{j=1,\cdots, D-d}$ is a smooth orthonormal normal frame on $N\calM$ along $\psi(U)$. And $[N_1, \cdots, N_{D-d}] = [e_{d+1}, \cdots, e_{D}]$ at $\zz=\bze$.
        \item Around $\mathbf{0} \in \calM$, we parametrize the manifold in a graph chart using the same notation $(U, \psi)$ with
        \begin{equation}\label{eqn:Fdef}
            \psi: U \subset \mathbb{R}^d \ria V\subset \calM \subset \mathbb{R}^D, \q \uu=(u^1, \cdots, u^d) \mapsto  \psi(\uu)=(\uu, F(\uu)).
        \end{equation}
              Here the vector-valued map $F(\uu)=(F^1(\uu), \cdots, F^{D-d}(\uu)):\mathbb{R}^d \mapsto \mathbb{R}^{D-d}$ satisfies $F(\mathbf{0})=\bze$ and $DF(\mathbf{0})=\mathbf{0}$.
    \end{enumerate}
    Evaluating the noisy density formula \eqref{eqn:barrhosigma} at $\zz=\bze$ gives
    \[ \bar{\rho}_\delta(\zz) = \int_\calM \int_{\R^{D-d}} \frac{1}{\delta^D} \eta \left( \frac{\left| tN|_\bze - \xx \right|}{\delta} \right) \det (I_d-t S_N|_{\bze}) \rho(\xx)\, d \tt d \mathrm{vol}_\calM(\xx).\]
    We can parametrize $\xx = \psi(\delta \uu)=(\delta \uu, F(\delta \uu))$ and $\nn=\tt / \delta$. Note that $[N_1, \cdots, N_{D-d}] = [e_{d+1}, \cdots, e_{D}]$ at $\zz=\bze$, we have $t=|\tt| = \delta|\nn|$ and
    \[ tN|_\bze = \sum_{j=1}^{D-d} t_j N_j =  \sum_{j=1}^{D-d} \delta n_j N_j = (\bze, \delta \nn).\]

    Then the kernel argument becomes
    \[ \frac{\left| tN|_\bze - \xx \right|}{\delta} = \frac{\left| (\bze, \delta \nn) - (\delta \uu, F(\delta \uu)) \right|}{\delta} = \left| \left(-\uu, \nn -\delta^{-1} F(\delta \uu)\right) \right|. \]

    Write $S_j:= S_{N_j}|_{\bze}$ for $j=1,\cdots, D-d$ and $B(\nn):=\sum_j n_j S_j$. We have
    \[ t S_N|_{\bze} = \sum_{j=1}^{D-d} t_j S_j = \delta \sum_{j=1}^{D-d} n_j S_j =: \delta B(\nn). \]

    The manifold volume element is given by
    \[  d \mathrm{vol}_\calM(\xx) = \sqrt{\det ((D\psi(\delta \uu))^T D\psi(\delta \uu))} \delta^d d\uu = \sqrt{\det (I_d + (DF(\delta \uu))^T DF(\delta \uu))} \delta^d d\uu.\]
    Denote the graph Jacobian
    \[
        J_\delta(\uu)
        :=
        \sqrt{\det(I_d+(DF(\delta \uu))^TDF(\delta \uu))}.
    \]

    Therefore, we can derive the noisy density formula in the graph chart as follows:
    \begin{align*}
        \bar{\rho}_\delta(\zz) = & \int_\calM \int_{\R^{D-d}} \frac{1}{\delta^D} \eta \left( \frac{\left| tN|_\bze - \xx \right|}{\delta} \right) \det (I_d-t S_N|_{\bze}) \rho(\xx)\, d \tt d \mathrm{vol}_\calM(\xx)                                    \\
        =                        & \int_{\R^{d}} \int_{\R^{D-d}} \eta \left( \left| \left(-\uu, \nn -\delta^{-1} F(\delta \uu)\right) \right| \right) \det (I_d- \delta B(\nn)) \rho(\psi(\delta \uu))\delta^{-d} \, d \nn J_\delta(\uu)\delta^d \, d \uu \\
        =                        & \int_{\R^{D}} \eta \left( \left| \left(-\uu, \nn -\delta^{-1} F(\delta \uu)\right) \right| \right) \det (I_d- \delta B(\nn)) \rho(\psi(\delta \uu)) J_\delta(\uu)\, d \nn d \uu                                        \\
        =                        & \int_{\R^{D}} \eta \left( \left| \left(-\uu, \nn\right) \right| \right) \det \left(I_d- \delta B(\nn + \delta^{-1} F(\delta \uu))\right) \rho(\psi(\delta \uu)) J_\delta(\uu)\, d \nn d \uu,
    \end{align*}
    where we apply change of variables $\nn= \nn -\delta^{-1} F(\delta \uu)$ for the last equality.

    On the other hand, we have
    \begin{equation*}
        \rho(\zz) = \rho(\zz) \int_{\R^D}\eta(|\xx|) \, d\xx =\int_{\R^D}  \eta(|(-\uu, \nn)|) \rho(\zz) \, d\nn d\uu.
    \end{equation*}
    Denote
    \[ P_\delta(\nn) := \det(I_d-\delta B(\nn))= \det\left(I_d-\delta \sum_{j=1}^{D-d} n_j S_j\right). \]

    Since $\eta$ is supported in the unit ball, and the transformation
    $\nn\mapsto-\nn$ preserves $\eta(|(-\uu,\nn)|)$ in $B_{\mathbb R^D}(\bze,1)$, we can average the centered identity with the
    same identity after this transformation. Thus the density bias can be
    written in the symmetrized form
    \begin{equation}\label{eqn:maincenteredbias}
        \begin{aligned}
            \bar\rho_\delta(\bze)-\rho(\bze)
            ={} &
            \int_{B_{\mathbb R^D}(\bze,1)}
            \eta(|(-\uu,\nn)|)
            \Bigg[
                \frac12\Big(
                P_\delta(\nn+\delta^{-1}F(\delta \uu))+P_\delta(-\nn+\delta^{-1}F(\delta \uu))
            \Big)              \\
                & \qquad\qquad
                {}\times\rho(\psi(\delta \uu))J_\delta(\uu)
                -\rho(\bze)
                \Bigg]
            \,d\nn\,d\uu .
        \end{aligned}
    \end{equation}
    This is the starting identity we will expand and analyze.

    \paragraph{3. Pointwise expansion of the integrand factors in \eqref{eqn:maincenteredbias}.} We now expand the three factors in \eqref{eqn:maincenteredbias} up to order $\delta^3$. We first expand the graph function $F$ defined in \eqref{eqn:Fdef}. 
    Since \(F(\mathbf 0)=0\) and \(DF(\mathbf 0)=0\), Taylor's expansion theorem gives
    \[
        F(\delta\uu)
        =
        \frac{\delta^2}{2}D^2F(\mathbf 0)[\uu,\uu]
        +
        \delta^3 r_{\delta}(\uu) \q \text{with} \q
        r_{\delta}(\uu)
        =
        \int_0^1
        \frac{(1-t)^2}{2}
        D^3F(t\delta\uu)[\uu,\uu,\uu]\,dt .
    \]
    Using \(D^2F^j(\mathbf 0)[\uu,\uu]=\uu^T S_j\uu\),
    we obtain
    \[
        F(\delta\uu)
        =
        \frac{\delta^2}{2} (\uu^T S_1\uu, \cdots, \uu^T S_{D-d}\uu)^T +\delta^3 r_{\delta}(\uu).
    \]
    Denote the global graph $C^3$-seminorm by $G_3:=
        \sup_{x\in\calM}\sup_{|v|\le r_{\calM}}\|D^3F_x(v)\|_{\op}$, where $F_x:T_x\calM \ria N_x\calM$ is the local graph function of $\calM$ over $T_x\calM$, we have
    \[
        |r_{\delta}(\uu)|
        \le
        \int_0^1
        \frac{(1-t)^2}{2}
        \|D^3F(t\delta\uu)\|_{\op}
        |\uu|^3\,dt
        \le
        G_3|\uu|^3.
    \]

    To obtain the expansion of each determinant factor in the symmetrized expression, we first use the linearity of $B(\nn)=\sum_j n_j S_j$ in its normal argument to obtain
    \[
        B(\nn+\delta^{-1}F(\delta\uu))
        =
        B(\nn)
        +\frac{\delta}{2}
        \sum_{j=1}^{D-d}(\uu^TS_j\uu)S_j
        +\delta^2 B(r_\delta(\uu)).
    \]

    Hence, for either choice of sign,
    \[
        I_d-\delta B(\pm\nn+\delta^{-1}F(\delta\uu))
        =
        I_d\mp\delta B(\nn)
        -\frac{\delta^2}{2}
        \sum_{j=1}^{D-d}(\uu^TS_j\uu)S_j
        -\delta^3 B(r_\delta(\uu)).
    \]
    We apply the exact second-order determinant decomposition
    \[
        \det(I_d+A)
        =
        1+\tr(A)
        +\frac12\left((\tr A)^2-\tr(A^2)\right)
        +R_{\det}(A),
    \]
    with the remainder term $R_{\det}(A)$ satisfying
    \begin{equation}\label{eqn:Rdetbdd}
        |R_{\det}(A)|
        \le
        \sum_{k=3}^d\frac{\|A\|_*^k}{k!}
        \le
        \frac{e^{\|A\|_*}}{6}\|A\|_*^3 \q \text{with}
        \q
        \|A\|_*\le\sqrt d\,\|A\|_F .
    \end{equation}
    Here $\|A\|_*$ is the nuclear norm of $A$, i.e., the sum of its singular values.

    For the determinant factor with $A=-\delta B(\nn+\delta^{-1}F(\delta\uu))$,  the first-order trace term gives
    \[
        \tr(A)
        =
        -\delta\sum_{j=1}^{D-d}n_j\tr(S_j)
        -\frac{\delta^2}{2}
        \sum_{j=1}^{D-d}\tr(S_j)\,\uu^TS_j\uu
        -\delta^3\tr(B(r_\delta(\uu))).
    \]
    Moreover, the degree-two contribution of
    $\frac12((\tr A)^2-\tr(A^2))$ comes only from the leading part
    $-\delta B(\nn)$ and is
    \[
        \frac{\delta^2}{2}
        \left[
            \left(\tr(B(\nn))\right)^2
            -\tr\left( B(\nn)^2 \right)
            \right].
    \]

    Replacing $\nn$ by $-\nn$ gives the expansion of the second determinant
    factor in \eqref{eqn:maincenteredbias}. Averaging the two expansions
    removes the term that is linear in $\nn$ and gives the following expansion:
    \begin{align}
            & \frac12\Big(
        P_\delta(\nn+\delta^{-1}F(\delta \uu))
        +
        P_\delta(-\nn+\delta^{-1}F(\delta \uu))
        \Big) \notag                     \\
        ={} & 1+\frac{\delta^2}{2}\left[
            \left(\tr(B(\nn))\right)^2
            -\tr\left( B(\nn)^2 \right)
            -\sum_{j=1}^{D-d}\tr(S_j)\,\uu^T S_j\uu
            \right] +R_P(\delta,\uu,\nn), \label{eqn:Pdeltaexpand}
    \end{align}
    where $R_P(\delta,\uu,\nn)$ consists of the higher-order terms in $\delta$ and will be bounded in the step 5.

    Next we proceed to  expand the density term. Using the fact that $\psi(\bze)=\bze$ gives
    \begin{equation}\label{eqn:rhoexpandnew}
        \rho(\psi(\delta\uu))=
        \rho(\bze)+\delta \nabla(\rho\circ\psi)(\bze)^T\uu
        +\frac{\delta^2}{2}\uu^T\nabla^2(\rho\circ\psi)(\bze)\uu
        +R_\rho(\delta,\uu),
    \end{equation}
    where
    \begin{equation}\label{eqn:Rrhobdd}
        |R_\rho(\delta,\uu)|
        \le C_{\calM, \rho}\delta^3|\uu|^3\le C_{\calM, \rho}\delta^3|\uu|^2.
    \end{equation}

    Due to $DF(\bze)=\mathbf{0}$, one can find that
    \begin{equation}\label{eqn:DFdeltaexpand}
        DF(\delta\uu)
        =
        \delta\big(S_1\uu,\ldots,S_{D-d}\uu\big)
        +O_\calM(\delta^2|\uu|^2),
    \end{equation}
    where the constants in the $O_\calM$ notation are bounded by the global graph $C^3$-seminorm $G_3$.

    For small positive semidefinite $A$,
    \[\sqrt{\det(I+A)}=1+\frac12\tr(A)+ \widehat R_{\det}(A) \q \text{with} \q |\widehat R_{\det}(A)|\le C e^{\|A\|_*}\|A\|_*^2.\]
    Choose $A=(DF(\delta \uu))^T DF(\delta \uu)$, then the graph Jacobian expansion satisfies
    \begin{equation}\label{eqn:Jdeltaexpand}
        J_\delta(\uu)= \sqrt{\det(I_d+(DF(\delta \uu))^T DF(\delta \uu))}=
        1+\frac{\delta^2}{2}\sum_{j=1}^{D-d}|S_j\uu|^2
        +R_J(\delta,\uu),
    \end{equation}
    where
    \begin{equation*}
        |R_J(\delta,\uu)| \le C_\calM \delta^3 |\uu|^2 \left(\sum_{j=1}^{D-d}|S_j \uu |^2\right)^{1/2} + C_\calM \delta^4 |\uu|^4 + \widehat R_{\mathrm{det}}(DF(\delta \uu)^T DF(\delta \uu)).
    \end{equation*}
    Moreover, it follows from $\|A\|_* \le d \|A\|_{\op}$ and \eqref{eqn:DFdeltaexpand} that
    \[ \|A\|_* \le d \|A\|_{\op} = d \|DF(\delta \uu)\|_{\op}^2 \le d \left( \delta \kappa_{\mathrm{max}} |\uu| + G_3 \delta^2 |\uu|^2 \right)^2 \le d \delta^2|\uu|^2 (\kappa_{\mathrm{max}} + G_3 r_\calM)^2, \]
    which implies $\widehat R_{\mathrm{det}}(A)\le C_\calM e^{(\kappa_{\mathrm{max}} + G_3 r_\calM)^2 d \delta^2} d^2 \delta^4 |\uu|^4$. If $\delta \le C / d$ with $C$ independent of dimensions, we can further simplify the bound to $\widehat R_{\mathrm{det}}(A)\le C_\calM d^2 \delta^4 |\uu|^4$. Combining with the other terms in $R_J$, we have
    \begin{equation}\label{eqn:RJbdd}
        |R_J(\delta,\uu)| \le C_{\calM}\left(
        \delta^3|\uu|^2\left(\sum_{j=1}^{D-d}|S_j\uu|^2\right)^{1/2}
        +\delta^4|\uu|^4
        +d^2\delta^4|\uu|^4
        \right).
    \end{equation}

    \paragraph{4. Integrated estimates, moment bounds and $d/D$ refinement.} Multiplying \eqref{eqn:Pdeltaexpand}--\eqref{eqn:Jdeltaexpand}, the
    first-order contribution in $\delta$ in \eqref{eqn:maincenteredbias} is
    \(\nabla(\rho\circ\psi)(\bze)^T\uu\). Its integral against the radial kernel is zero by odd
    symmetry in $\uu$. Therefore, the only second-order terms that can
    contribute are
    \[
        \frac{1}{2}\rho(\bze) I(\uu, \nn) +\frac12 \uu^T\nabla^2(\rho\circ\psi)(\bze)\uu.
    \]
    where
    \[I(\uu, \nn):=\left(\sum_{j=1}^{D-d}n_j\tr(S_j)\right)^2
        -\tr(B(\nn)^2)
        -\sum_{j=1}^{D-d}\tr(S_j)\,\uu^T S_j\uu
        +\sum_{j=1}^{D-d}|S_j\uu|^2.\]

    In fact, the curvature part cancels exactly. Denote $k$-th moment of $\eta$ by
    \[
        m_k(\eta):=
        \int_{B_{\mathbb R^D}(\bze,1)}|\xx|^k\eta(|\xx|)\,d\xx
        \le \int_{B_{\mathbb R^D}(\bze,1)}\eta(|\xx|)\,d\xx = 1.
    \]
    Since $\eta(|\cdot|)$ is radial, for all $1\le i,j\le D$,
    \begin{equation}\label{eqn:radialmoments}
        \int_{B_{\mathbb R^D}(\bze,1)}x_i x_j\eta(|\xx|)\,d\xx
        =
        \frac{m_2(\eta)}{D}\mathbf 1_{\{i=j\}}.
    \end{equation}
    In particular,
    \begin{equation}\label{eqn:radialmoment2}
        \int_{B_{\mathbb R^D}(\bze,1)} |\uu|^2\eta(\|(-\uu,\nn)\|)\,d\nn\,d\uu
        = \frac{d}{D}m_2(\eta) \le \frac{d}{D}.
    \end{equation}

    Then we can compute the second-order curvature contributions as follows:
    \begin{align}
        \int_{B_{\mathbb R^D}(\bze,1)}
        \left(\sum_{j=1}^{D-d}n_j\tr(S_j)\right)^2
        \eta(|(-\uu,\nn)|)
        \,d\nn\,d\uu
         & =\frac{m_2(\eta)}{ D}\sum_{j=1}^{D-d}\tr(S_j)^2,  \label{eqn:njtrsjintegral} \\
        \int_{B_{\mathbb R^D}(\bze,1)}
        \tr(B(\nn)^2)
        \eta(|(-\uu,\nn)|)
        \,d\nn\,d\uu
         & =\frac{m_2(\eta)}{D}\sum_{j=1}^{D-d}\|S_j\|_F^2,                             \\
        \int_{B_{\mathbb R^D}(\bze,1)}
        \sum_{j=1}^{D-d}\tr(S_j)\,\uu^T S_j\uu
        \eta(|(-\uu,\nn)|)
        \,d\nn\,d\uu
         & =\frac{m_2(\eta)}{D}\sum_{j=1}^{D-d}\tr(S_j)^2,                              \\
        \int_{B_{\mathbb R^D}(\bze,1)}
        \sum_{j=1}^{D-d}|S_j\uu|^2
        \eta(|(-\uu,\nn)|)
        \,d\nn\,d\uu
         & =\frac{m_2(\eta)}{D}\sum_{j=1}^{D-d}\|S_j\|_F^2. \label{eqn:sjuu2integral}
    \end{align}
    We have $\int I(\uu, \nn)\eta(|(-\uu,\nn)|)\,d\nn\,d\uu=0$, and therefore the curvature contributions vanish in the integrated bias. This is a key cancellation that allows us to obtain the refined $d/D$-dependent estimate.
    The remaining density term is bounded by
    \[
        \left|
        \frac{\delta^2}{2}
        \int_{B_{\mathbb R^D}(\bze,1)}
        \uu^T\nabla^2(\rho\circ\psi)(\bze)\uu\,
        \eta(|(-\uu,\nn)|)
        \,d\nn\,d\uu
        \right|
        \le
        \frac{d}{2D}\|\nabla^2(\rho\circ\psi)(\bze)\|_{\op}\,\delta^2.
    \]
    Hence, the second-order contribution to the integrated bias is bounded by $\frac dD\,\delta^2$, i.e.
    \[ |\bar\rho_\delta(\bze)-\rho(\bze)|
        \le
        \frac{d}{2D}\|\nabla^2(\rho\circ\psi)(\bze)\|_{\op}\,\delta^2 + O(\delta^3).
    \]
    Here the higher order contributions are bounded by the integrated remainder terms, which will be estimated in the next step. Since the point $\zz$ and the chart are arbitrary, we can take the supremum over $\calM$ and all graph charts to obtain the estimate \eqref{eqn:rhoerrbdd} in the theorem statement.

    \paragraph{5. Estimates of $O(\delta^3)$ integrated remainder contributions.}
    For simplicity, we omit the integral domain $B_{\mathbb R^D}(\bze,1)$ in the following calculations unless specified otherwise. We define $R_\delta(\uu,\nn)$ by collecting all terms beyond the first
    and second order terms in the product expansion, namely
    \begin{align}
            & \frac12\Big(
        P_\delta(\nn+\delta^{-1}F(\delta\uu))
        +P_\delta(-\nn+\delta^{-1}F(\delta\uu))
        \Big)
        \rho(\psi(\delta\uu))J_\delta(\uu)
        -\rho(\bze) \notag \\
        ={} &
        \delta\nabla(\rho\circ\psi)(\bze)^T\uu
        +
        \frac{\delta^2}{2}
        \left[
            \uu^T\nabla^2(\rho\circ\psi)(\bze)\uu
            +\rho(\bze)I(\uu,\nn)
            \right]
        +R_\delta(\uu,\nn).
        \label{eqn:RdeltaDefinition}
    \end{align}
    Equivalently, $R_\delta(\uu,\nn)$ is the unique function satisfying
    \eqref{eqn:RdeltaDefinition}. We now show that the integrated contribution of $R_\delta(\uu,\nn)$ is bounded by $C_{\calM, \rho} \frac{d}{D} \delta^2$. Using the expansion formulas and the bounds for the remainder terms,
    the integrated remainder can be written as
    \begin{align}
         & \int R_\delta(\uu,\nn)\eta(|(-\uu,\nn)|)\,d\nn\,d\uu
        = \int \Bigg[ R_\rho  +\rho(\bze)R_P +\rho(\bze)R_J +\delta\nabla(\rho\circ\psi)(\bze)^T\uu (R_P+R_J)   \notag                                                                                                                        \\
         & \qquad +\frac{\delta^2}{2}\uu^T\nabla^2(\rho\circ\psi)(\bze)\uu
        \left(\frac{\delta^2}{2}I+R_P+R_J \right)+R_\rho \left(\frac{\delta^2}{2}I +R_P+R_J\right)   \label{eqn:RdeltaIntegrated}                                                                                                             \\
         & \qquad + \left(\frac{\delta^2}{2}I - \frac{\delta^2}{2} \sum_{j=1}^{D-d}|S_j\uu|^2 +R_P \right)  \left(\frac{\delta^2}{2}\sum_{j=1}^{D-d}|S_j\uu|^2+R_J \right) \rho(\psi(\delta \uu))\Bigg]\eta(|(-\uu,\nn)|)\,d\nn\,d\uu. \notag
    \end{align}

    Before estimating the integrated remainder contributions, we first denote the full squared second fundamental form by
    \begin{equation*}
        |\mathrm{II}_0|^2 := \sum_{j=1}^{D-d}\|S_j\|_F^2  = \sum_{j=1}^{D-d}\sum_{a,b=1}^d |\langle S_je_a, e_b \rangle|^2,
    \end{equation*}
    where $\{e_1, \cdots, e_d\}$ is the coordinate basis of $T_0\calM$. Recall that $\{N_j\}_{j=1,\cdots, D-d}$ is the orthonormal basis of $N_0\calM$, we can write
    \[\mathrm{II}_0(e_a, e_b) = \sum_{j=1}^{D-d} \langle \mathrm{II}_0(e_a, e_b), N_j \rangle N_j = \sum_{j=1}^{D-d} \langle S_je_a, e_b \rangle N_j.\]
    Then the orthonormality of $\{N_j\}$ implies
    \[|\mathrm{II}_0(e_a, e_b)|^2 = \sum_{j=1}^{D-d} |\langle S_je_a, e_b \rangle|^2.\]
    On the other hand, the dual characterization of the Euclidean norm gives
    \[ |\mathrm{II}_0(e_a, e_b)| = \sup_{\substack{v\in N_x\calM \\ \|v\|=1}} |\langle \mathrm{II}_0(e_a, e_b), v \rangle|  = \sup_{\substack{v\in N_x\calM \\ \|v\|=1}} |\langle S_v e_a, e_b \rangle| \le \kappa_{\mathrm{max}}. \]
    Therefore, we have the following bound for the full second fundamental form:
    \begin{equation}\label{eqn:II0bdd}
        |\mathrm{II}_0|^2 = \sum_{j=1}^{D-d}\sum_{a,b=1}^d |\langle S_je_a, e_b \rangle|^2= \sum_{a,b=1}^d |\mathrm{II}_0(e_a, e_b)|^2 \le d^2 \kappa_{\mathrm{max}}^2.
    \end{equation}

    Using the similar idea as above, one can find
    \begin{align*}
         & \left(\sum_{j=1}^{D-d}n_j\tr(S_j)\right)^2 = \left| \langle \nn, \sum_{j=1}^{D-d} \tr(S_j) N_j \rangle \right|^2 \le |\nn|^2 \left|\sup_{\substack{v\in N_x\calM                        \\ \|v\|=1}} |\tr (S_v)| \right|^2 \le d^2 \kappa_{\mathrm{max}}^2 |\nn|^2, \\
         & \tr(B(\nn)^2) \le d \|B(\nn)\|_{\op}^2 = d \left\| \sum_{j=1}^{D-d} n_j S_j \right\|_{\op}^2  = d \| |\nn| S_{\nn/|\nn|} \|_{\op}^2\le d \kappa_{\mathrm{max}}^2 |\nn|^2,               \\
         & \sum_{j=1}^{D-d}\tr(S_j) \uu^T S_j\uu \le \left(\sum_{j=1}^{D-d} |\tr(S_j)|^2\right)^{1/2} \left(\sum_{j=1}^{D-d} |\uu^T S_j \uu|^2\right)^{1/2} \le d \kappa_{\mathrm{max}}^2 |\uu|^2, \\
         & \sum_{j=1}^{D-d}|S_j\uu|^2 = \sum_{j=1}^{D-d}\sum_{a=1}^d |\langle S_j \uu, e_a \rangle|^2  = \sum_{a=1}^d |\mathrm{II}_0(\uu, e_a)|^2 \le d \kappa_{\mathrm{max}}^2 |\uu|^2.
    \end{align*}
    Therefore, by the definition of $I(\uu, \nn)$ and the condition $\delta \le C / d$, we have
    \begin{equation}\label{eqn:Ibdd}
        \frac{\delta^2}{2}|I(\uu, \nn)| \le 2 d^2 \delta^2 \kappa_{\mathrm{max}}^2 \le 2 C^2 \kappa_{\mathrm{max}}^2  \q \text{and} \q \frac{\delta^2}{2} \sum_{j=1}^{D-d}|S_j\uu|^2 \le \frac{\delta^2}{2} d \kappa_{\mathrm{max}}^2  \le \frac{Cr_\calM \kappa_{\mathrm{max}}^2}{2}.
    \end{equation}

    Next we are going to show that under the same condition on $\delta$, the integrated contribution of the remainder terms is bounded by $C_{\calM, \rho} \frac{d}{D} \delta^2$, i.e.,
    \begin{equation}\label{eqn:remainderint1}
        \int \left(|R_\rho(\delta,\uu)|+|R_J(\delta,\uu)|+|R_P(\delta,\uu,\nn)|\right)
        \eta(|(-\uu,\nn)|)\,d\nn\,d\uu
        \le C_{\calM,\rho}\frac dD\,\delta^2.
    \end{equation}
    We first estimate $R_\rho(\delta,\uu)$. Since the kernel is supported on $|(-\uu,\nn)|\le1$, we have
    $|\uu|^3\le|\uu|^2$, and \eqref{eqn:radialmoment2} gives
    \begin{equation}\label{eqn:Rrhointegrated}
        \int|R_\rho(\delta,\uu)|
        \eta(|(-\uu,\nn)|)\,d\nn\,d\uu
        \le
        C_{\calM,\rho}\frac dD\,\delta^3.
    \end{equation}

    We proceed to estimate the Jacobian remainder $R_J(\delta, \uu)$. The fourth-moment of the radial kernel is given by
    \[ \int |\uu|^4\eta(|(-\uu,\nn)|)\,d\nn\,d\uu \le C \left(\frac dD\right)^{2}. \]
    Applying the Cauchy-Schwarz inequality with \eqref{eqn:sjuu2integral} and \eqref{eqn:II0bdd}, we can estimate the first term in \eqref{eqn:RJbdd} as follows:
    \begin{align*}
         & \int |\uu|^2\left(\sum_{j=1}^{D-d}|S_j\uu|^2\right)^{1/2}
        \eta(|(-\uu,\nn)|)\,d\nn\,d\uu                               \\
         & \qquad\qquad\le
        \left(\int |\uu|^4\eta(|(-\uu,\nn)|)\,d\nn\,d\uu\right)^{1/2}
        \left(\int \sum_{j=1}^{D-d}|S_j\uu|^2
        \eta(|(-\uu,\nn)|)\,d\nn\,d\uu\right)^{1/2}                  \\
         & \qquad\qquad\le
        C\frac{d}{D^{3/2}}
        \left(\sum_{j=1}^{D-d}\|S_j\|_F^2\right)^{1/2} \le C_\calM\frac{d^2}{D^{3/2}}.
    \end{align*}
    Therefore, under the condition $\delta \lesssim 1/d$ we have
    \begin{equation}\label{eqn:RJintegrated}
        \int |R_J(\delta,\uu)|
        \eta(|(-\uu,\nn)|)\,d\nn\,d\uu
        \le
        C_{\calM}\left(
        \delta^3\frac{d^2}{D^{3/2}} + \delta^4\frac{d^2}{D^2} + \delta^4\frac{d^4}{D^2}\right) \le C_\calM \frac{d}{D}\delta^2.
    \end{equation}

    We next estimate $R_P$. To avoid repeatedly writing the shifted
    part of the determinant, denote
    \[
        H(\uu):=\frac{\delta^2}{2}
        \sum_{j=1}^{D-d}(\uu^TS_j\uu)S_j
        +\delta^3B(r_\delta(\uu)).
    \]
    By the definition of $R_P$ in \eqref{eqn:Pdeltaexpand} and the exact
    determinant expansion in Step~3, we have
    \begin{align}
        R_P(\delta,\uu,\nn)
        ={} &
        -\delta^3\tr(B(r_\delta(\uu)))
        +\frac12\left[(\tr H)^2-\tr(H^2)\right] \notag \\
            & \quad
        +\frac12R_{\det}(-\delta B(\nn)-H)
        +\frac12R_{\det}(\delta B(\nn)-H). \label{eqn:RPexactdecomposition}
    \end{align}
    The $\nn\mapsto-\nn$ symmetrization cancels all determinant terms
    containing an odd number of factors $\delta B(\nn)$.
    First we can estimate the first trace term by
    \begin{equation}\label{eqn:graphtraceRP}
        \left|
        \tr(B(r_\delta(\uu)))
        \right|
        = \left|\langle r_\delta(\uu), \sum_{j=1}^{D-d} \tr(S_j)N_j \rangle \right|
        \le  C_\calM |\uu|^3 \left| \sum_{i=1}^d \mathrm{II}_0(e_i,e_i) \right| \le C_\calM  |\uu|^3 d \kappa_{\max}.
    \end{equation}

    For the quadratic term, we can estimate via the Cauchy-Schwarz inequality and the bound for $r_\delta$ to obtain
    \begin{align}
        \left|\frac12\left[(\tr H)^2-\tr(H^2)\right]\right|
         & \le d\|H\|_F^2 \le d\delta^4
        \left\|\sum_{j=1}^{D-d}(\uu^TS_j\uu)S_j\right\|_F^2
        +C_{\calM}d\delta^6|\uu|^6
        |\mathrm{II}_0|. \label{eqn:HquadraticRP}
    \end{align}

    Since the average in \eqref{eqn:RPexactdecomposition} is even in
    $\delta B(\nn)$, its pure-normal contribution begins at order four. Applying the bound for $R_{\det}$ in \eqref{eqn:Rdetbdd} gives
    \begin{align}
         & \left|
        \frac12R_{\det}(-\delta B(\nn)-H)
        +\frac12R_{\det}(\delta B(\nn)-H)
        \right| \notag             \\
         & \qquad \qquad \qquad\le
        C e^{\|\delta B(\nn)\|_*+\|H\|_*}
        \left(\|\delta B(\nn)\|_*^4
        +\|\delta B(\nn)\|_*^2\|H\|_*
        +\|H\|_*^3\right). \label{eqn:symdetRemainder}
    \end{align}
    The last term in \eqref{eqn:symdetRemainder} records the pure shifted
    determinant terms of degree at least three.
    Recall that $B(\nn) = \sum_j n_j S_j$, then we have $\|B(\nn)\|_{\op} \le \kappa_{\max}|\nn|$ for all $\nn \in \mathbb{R}^{D-d}$. Hence, combining with the fact $\|A\|_* \le d \|A\|_{\op}$ for any $d\times d$ matrix $A$ gives $\|\delta B(\nn)\|_*\le d\delta\kappa_{\max}|\nn|$ and
    \[
        \|H\|_*=\|B(F(\delta\uu))\|_* \le d\kappa_{\max}|F(\delta\uu)| \le d\kappa_{\max} \left( \delta^2 \kappa_{\max} |\uu|^2 + \delta^3 G_3 |\uu|^3 \right) \le C_{\calM}d\delta^2|\uu|^2.
    \]
    Since $|\uu|,|\nn|\le1$ on the kernel support and
    $\delta\le r_{\calM}$, the exponential factor is bounded by $e^{C_{\calM}d\delta}$.
    Since $\|A\|_*\le\sqrt d\,\|A\|_F$, we can bound the polynomial factor in \eqref{eqn:symdetRemainder} as follows:
    \begin{align}
        & \|\delta B(\nn)\|_*^4
        +\|\delta B(\nn)\|_*^2\|H\|_*
        +\|H\|_*^3 \le  d^2\delta^4\|B(\nn)\|_F^4 + d^{3/2}\delta^2 \|B(\nn)\|_F^2\|H\|_F + \|H\|_*^3 \notag                                           \\
        \le \; & d^2\delta^4\|B(\nn)\|_F^4 + d^{3/2}\delta^4\|B(\nn)\|_F^2
        \left\|\sum_{j=1}^{D-d}(\uu^TS_j\uu)S_j\right\|_F  
         + d^{3/2}\delta^5\|B(\nn)\|_F^2\|B(r_\delta(\uu))\|_F \notag \\
         & \quad + C_\calM d^3 \delta^6 |\uu|^6 \notag                                  \\
         \le \; &  d^2\delta^4\|B(\nn)\|_F^4 + d^{3/2}\delta^4\|B(\nn)\|_F^2
        \left\|\sum_{j=1}^{D-d}(\uu^TS_j\uu)S_j\right\|_F   + C_\calM d^{3/2}\delta^5\|B(\nn)\|_F^2 |\uu|^3 |\mathrm{II}_0| \notag \\
        & \quad + C_\calM d^3 \delta^6 |\uu|^6. \label{eqn:detRemainderPoly}
    \end{align}

    Combining equations \eqref{eqn:graphtraceRP}--\eqref{eqn:detRemainderPoly} gives the pointwise bound
    \begin{align}
        |R_P(\delta,\uu,\nn)|
        \le{}
         & C_{\calM}e^{C_{\calM}d\delta}
        \Bigg[
        \delta^3 d\,|\uu|^3
        +d\delta^4
        \left\|\sum_{j=1}^{D-d}(\uu^TS_j\uu)S_j\right\|_F^2
        +d^2\delta^4\|B(\nn)\|_F^4 \notag                            \\
         & \quad
        +d^{3/2}\delta^4\|B(\nn)\|_F^2
        \left\|\sum_{j=1}^{D-d}(\uu^TS_j\uu)S_j\right\|_F
        +d^{3/2}\delta^5\|B(\nn)\|_F^2|\uu|^3|\mathrm{II}_0| \notag  \\
         & \quad +d\delta^6|\uu|^6|\mathrm{II}_0|+d^3\delta^6|\uu|^6
        \Bigg]. \label{eqn:RPpointwise}
    \end{align}

    We next integrate \eqref{eqn:RPpointwise}. Radial symmetry gives the following moment estimates:
    \begin{align}
        \int |\uu|^3\eta(|(-\uu,\nn)|)\,d\nn\,d\uu
         & =
        m_3(\eta)
        \frac{\Gamma(\frac d2+\frac32)\Gamma(\frac D2)}
        {\Gamma(\frac d2)\Gamma(\frac D2+\frac32)}
        \le
        C \left(\frac dD\right)^{3/2}, \label{eqn:thirdmomentu} \\
        \int |\uu|^6\eta(|(-\uu,\nn)|)\,d\nn\,d\uu
         & =
        m_6(\eta)\frac{d(d+2)(d+4)}
        {D(D+2)(D+4)} \le C \left(\frac dD\right)^3. \label{eqn:sixthmomentu}
    \end{align}
    For any symmetric matrix $A, B\in \R^{d\times d}$, we have the following fourth-moment identity:
    \begin{equation}\label{eqn:fourthmomentid}
        \int (\uu^T A \uu)(\uu^T B \uu)\eta(|(-\uu,\nn)|)\,d\nn\,d\uu
        =\frac{m_4(\eta)}{D(D+2)}\left(2\tr(AB)+\tr(A)\tr(B)\right).
    \end{equation}
    Hence, we can estimate
    \begin{align}
         & \int\left\|\sum_{j=1}^{D-d}(\uu^TS_j\uu)S_j\right\|_F^2\eta(|(-\uu,\nn)|) \,d\nn\,d\uu
        \le\frac{m_4(\eta)(d+2)}{D(D+2)}|\mathrm{II}_0|^2 \le C \frac{d}{D^2}|\mathrm{II}_0|^2, \label{eqn:quadcurvintegrated} \\
         & \int\|B(\nn)\|_F^4\eta(|(-\uu,\nn)|) \,d\nn\,d\uu
        \le \frac{3m_4(\eta)}{D(D+2)}|\mathrm{II}_0|^2 \le C D^{-2}|\mathrm{II}_0|^2. \label{eqn:Bfourintegrated}
    \end{align}
    Applying the Cauchy-Schwarz inequality to estimate the mixture terms to obtain
    \begin{align}
         & \int\|B(\nn)\|_F^2
        \left\|\sum_{j=1}^{D-d}(\uu^TS_j\uu)S_j\right\|_F\eta(|(-\uu,\nn)|) \,d\nn\,d\uu
        \le C \frac{\sqrt{d}}{D^2}|\mathrm{II}_0|^2, \label{eqn:Bquadcurvintegrated} \\
         & \int\|B(\nn)\|_F^2|\uu|^3\eta(|(-\uu,\nn)|) \,d\nn\,d\uu
        \le C \frac{d^{3/2}}{D^{5/2}}|\mathrm{II}_0|. \label{eqn:BEintegrated}
    \end{align}

    Substituting these estimates \eqref{eqn:thirdmomentu}--\eqref{eqn:BEintegrated} in \eqref{eqn:RPpointwise} gives
    \begin{align}
        \int |R_P(\delta,\uu,\nn)|
        \eta(|(-\uu,\nn)|)\,d\nn\,d\uu \notag
         & \le
        C_{\calM}e^{C_{\calM}d\delta}
        \Bigg[
        \delta^3 \frac{d^{5/2}}{D^{3/2}}
        +\delta^4\frac{d^2}{D^2}|\mathrm{II}_0|^2 +\delta^5\frac{d^3}{D^{5/2}}|\mathrm{II}_0|^{2}  \notag \\
         & \qquad + \delta^6|\mathrm{II}_0| \frac{d^4}{D^3} + \delta^6 \frac{d^{6}}{D^3}
        \Bigg]. \label{eqn:RPintegrated}
    \end{align}

    Finally, under the condition $\delta \le C/d$ with $C$ independent of $d$ and $D$, we can further simplify \eqref{eqn:RPintegrated} with \eqref{eqn:II0bdd} to obtain
    \begin{align}
        \int |R_P(\delta,\uu,\nn)|
        \eta(|(-\uu,\nn)|)\,d\nn\,d\uu
         & \le C_{\calM}e^{C_{\calM}d\delta}
        \left(\delta^3 \frac{d^{5/2}}{D^{3/2}}
        +\delta^4\frac{d^4}{D^2}+\delta^5\frac{d^{5}}{D^{5/2}} + \delta^6  \frac{d^{5}}{D^3} + \delta^6 \frac{d^{6}}{D^3} \right) \notag \\
         & \le C_{\calM}
        \left(\delta^2 \frac{d^{3/2}}{D^{3/2}}
        +\delta^2\frac{d^2}{D^2}+\delta^2\frac{d^{2}}{D^{5/2}} + \delta^2  \frac{d}{D^3} + \delta^2 \frac{d^{2}}{D^3} \right) \notag     \\
         & \le C_\calM \frac{d}{D} \delta^2.\label{eqn:RPintegratedII}
    \end{align}

    Combining \eqref{eqn:Rrhointegrated}, \eqref{eqn:RJintegrated}, and \eqref{eqn:RPintegratedII} gives \eqref{eqn:remainderint1}. Therefore, the linear remainder contributions in \eqref{eqn:RdeltaIntegrated} are bounded by
    \[\int \left[ R_\rho  +\rho(\bze)R_P +\rho(\bze)R_J +\delta\nabla(\rho\circ\psi)(\bze)^T\uu (R_P+R_J) \right]\eta(|(-\uu,\nn)|)\,d\nn\,d\uu \le C_{\calM,\rho}\frac dD\,\delta^2.\]

    Note that $|R_\rho|$ and $|R_J|$ are bounded by some dimensional-free constant when $\delta\le C/d$. Combining the bound of $\delta^2 |I|$ in \eqref{eqn:Ibdd} and \eqref{eqn:remainderint1} gives the bound for the following product remainder terms in \eqref{eqn:RdeltaIntegrated}:
    \begin{align*}
            & \int \left[\frac{\delta^2}{2}\uu^T\nabla^2(\rho\circ\psi)(\bze)\uu
        \left(\frac{\delta^2}{2}I+R_P+R_J \right)+R_\rho \left(\frac{\delta^2}{2}I +R_P+R_J\right) \right]\eta(|(-\uu,\nn)|)\,d\nn\,d\uu                               \\
        \le & \int \left[ \delta^2 C_{\calM,\rho} \left(|\uu|^2 + |R_P|+|R_J|\right) + C_{\calM,\rho} (|R_\rho| + |R_P| + |R_J|) \right]\eta(|(-\uu,\nn)|)\,d\nn\,d\uu \\
        \le & \; C_{\calM,\rho}\frac dD\,\delta^2.
    \end{align*}
    In order to estimate the last product remainder term in \eqref{eqn:RdeltaIntegrated}, we first show that the following inequality holds under the condition $\delta\le C/d$:
    \begin{equation}\label{eqn:productremainder}
        \int \left| \frac{\delta^2}{2} I(\uu, \nn) - \frac{\delta^2}{2} \sum_{j=1}^{D-d}|S_j\uu|^2  \right|\eta(|(-\uu,\nn)|)\,d\nn\,d\uu \le \frac{3d^2 \delta^2 \kappa_{\mathrm{max}}^2}{2D} \le \frac{C_\calM}{D}.
    \end{equation}
    It is straightforward to apply \eqref{eqn:fourthmomentid} to compute
    \begin{align*}
        LHS & \le  \frac{\delta^2}{2} \int \left[ \left(\sum_{j=1}^{D-d}n_j\tr(S_j)\right)^2 + \tr(B(\nn)^2)
        + \left|\sum_{j=1}^{D-d}\tr(S_j)\,\uu^T S_j\uu\right| \right]\eta(|(-\uu,\nn)|)\,d\nn\,d\uu                                                                                                                                                   \\
            & \le \frac{\delta^2}{2} \left[ \frac{m_2(\eta)}{D}\left|\sum_{j=1}^{D-d}\tr(S_j)N_j \right|^2 + \frac{m_2(\eta)}{D}|\mathrm{II}_0|^2 + \left|\sum_{j=1}^{D-d}\tr(S_j)N_j \right| \int |\mathrm{II}_0(\uu, \uu)| \eta \,d\nn\,d\uu\right] \\
            & \le \frac{3\delta^2 d^2 \kappa_{\mathrm{max}}^2}{2D} \le \frac{C_\calM}{D},
    \end{align*}
    which gives \eqref{eqn:productremainder}. Combining \eqref{eqn:productremainder} with the bound of $\rho$, $R_J$, \eqref{eqn:Ibdd} and \eqref{eqn:remainderint1} yields
    \begin{align*}
            & \int \left(\frac{\delta^2}{2}I - \frac{\delta^2}{2} \sum_{j=1}^{D-d}|S_j\uu|^2 +R_P \right)  \left(\frac{\delta^2}{2}\sum_{j=1}^{D-d}|S_j\uu|^2+R_J \right) \rho(\psi(\delta \uu))\eta(|(-\uu,\nn)|)\,d\nn\,d\uu                             \\
        \le & \int \left| \frac{\delta^2}{2} I(\uu, \nn) - \frac{\delta^2}{2} \sum_{j=1}^{D-d}|S_j\uu|^2  \right| \frac{\delta^2}{2}\sum_{j=1}^{D-d}|S_j\uu|^2 \rho(\psi(\delta \uu)) \eta(|(-\uu,\nn)|)\,d\nn\,d\uu                                       \\
            & \quad + \int \left(\frac{\delta^2}{2}|I(\uu, \nn)| + \frac{\delta^2}{2} \sum_{j=1}^{D-d}|S_j\uu|^2 \right) |R_J|  \rho(\psi(\delta \uu))\eta(|(-\uu,\nn)|)\,d\nn\,d\uu                                                                       \\
            & \quad + \int |R_P| \left(\frac{\delta^2}{2}\sum_{j=1}^{D-d}|S_j\uu|^2+ |R_J| \right) \rho(\psi(\delta \uu))\eta(|(-\uu,\nn)|)\,d\nn\,d\uu                                                                                                    \\
        \le & \; \frac{C_\calM}{D}  \frac{\delta^2 d \kappa_{\mathrm{max}}^2}{2} \|\rho\|_{L^\infty(\calM)} + C_\calM \|\rho\|_{L^\infty(\calM)} C_{\calM,\rho}\frac{d}{D}\delta^2 + C_\calM \|\rho\|_{L^\infty(\calM)}  C_{\calM,\rho}\frac{d}{D}\delta^2 \\
        \le & \; C_{\calM,\rho}\frac{d}{D}\delta^2.
    \end{align*}
    Thus, collecting all the bounds for the linear and product remainder terms in \eqref{eqn:RdeltaIntegrated} gives that the total contribution of the remainder terms is bounded by $C_{\calM,\rho}\frac dD\,\delta^2$ under the condition $\delta\le C/d$.

    Combining these bounds with the exact cancellation of the
    second-order curvature contributions in Step~4 and the estimate for
    the remaining density term gives
    \[
        |\bar\rho_\delta(\bze)-\rho(\bze)|
        \le
        C_{\calM,\rho}\frac dD\,\delta^2.
    \]
    Since the base point is arbitrary and the bounds are uniform on
    $\calM$, taking the supremum over $\calM$ yields
    \[
        \|\rho-\bar\rho_\delta\|_{L^\infty(\calM)}
        \le
        C_{\calM,\rho}\frac dD\,\delta^2,
    \]
    which proves \eqref{eqn:rhoerrbdd2}.

\end{proof}

\begin{proof}[Proof of Theorem~\ref{thm:errorbdd}]
    Let $\calM_1, \cdots, \calM_\ell$ be disjoint connected components of $\calM$, i.e. $\calM = \bigsqcup_{j=1}^\ell \calM_j$. Define the probability measure on each component by 
    \[ \mu_j = \frac{\mu |_{\calM_j}}{\mu(\calM_j)} \quad \text{and} \quad \bar{\mu}_j = \frac{\bar{\mu}_\delta |_{\calM_j} }{\bar{\mu}_\delta(\calM_j)}.\]
    Note that each noised point projects back onto the component from which it originated, we have $\mu(\calM_j) = \bar{\mu}_\delta(\calM_j)$. Since $\rho$ is bounded below by Assumption~\ref{assumpmu}, the density of $\mu_j$ is also bounded below by $(\rho_0 \mu(\calM_j))^{-1}$ with $\mu(\calM_j) > \mathrm{vol}(\calM_j) \rho_0^{-1} >0 $.
     
    Applying Corollary~2.3 in \citet{peyre2018} for each $\calM_j$ with $j=1,\cdots, \ell$ gives
    \[\calW_2(\mu_j, \bar{\mu}_j) \le 2 \rho_0^{1/2} \|\bar{\mu}_j- \mu_j\|_{\dot{H}^{-1}(\calM_j)} = 2 ( \mu(\calM_j) \rho_0)^{1/2} \sup_{\substack{\varphi \in C^\infty(\calM_j): \\ \int \varphi = 0, \, \|\nabla_{\calM_j} \varphi\|_{L^2} \le 1}} \int_{\calM_j} (\bar{\mu}_j - \mu_j) \varphi \, d\mathrm{vol}_{\calM_j}.\]
    Then we can apply Cauchy-Schwarz inequality and Poincar\'{e}'s inequality to get 
    \[\left|\int_{\calM_j} (\bar{\mu}_j - \mu_j) \varphi \, d\mathrm{vol}_{\calM_j} \right| \le \|\bar{\mu}_j - \mu_j\|_{L^2(\calM_j)} \|\varphi\|_{L^2(\calM_j)} \le \lambda_{j,1}^{-1/2}\|\bar{\mu}_j - \mu_j\|_{L^2(\calM_j)} \|\nabla_{\calM_j} \varphi\|_{L^2(\calM_j)},\]
    where $\lambda_{j, 1}$ is the smallest positive eigenvalue of $-\Delta_{\calM_j}$. Combining the previous two inequalities gives
    \[\calW_2(\mu_j, \bar{\mu}_j) \le 2 (\mu(\calM_j) \rho_0)^{1/2}\lambda_{j,1}^{-1/2}\|\bar{\mu}_j - \mu_j\|_{L^2(\calM_j)}. \]
    Denote  $\lambda_1: = \min_{j=1,\cdots,\ell} \lambda_{j, 1}$. It follows from the joint convexity of squared $2$-Wasserstein distance and $\sum_{j=1}^\ell \mu(\calM_j) = 1$ that
    \begin{align*}
        \calW_2^2(\mu, \bar{\mu}_\delta) & = \calW_2^2\left( \sum_{j=1}^\ell \mu(\calM_j) \mu_j,  \sum_{j=1}^\ell \mu(\calM_j) \bar{\mu}_j \right) \le \sum_{j=1}^\ell \mu(\calM_j)  \calW_2^2(\mu_j, \bar{\mu}_j) \\
        & \le \sum_{j=1}^\ell  4 (\mu(\calM_j))^2 \rho_0\lambda_{j,1}^{-1}\|\bar{\mu}_j - \mu_j\|_{L^2(\calM_j)}^2 = \sum_{j=1}^\ell  4 \rho_0\lambda_{j,1}^{-1}\|\bar{\mu}_\delta - \mu\|_{L^2(\calM_j)}^2 \\
        & \le 4 \rho_0\lambda_{1}^{-1}\|\bar{\mu}_\delta - \mu\|_{L^2(\calM)}^2 \le 4 \rho_0\lambda_{1}^{-1} \mathrm{vol}(\calM)\|\bar{\mu}_\delta - \mu\|_{L^\infty(\calM)}^2 .
    \end{align*}
    Consequently, applying Theorem~\ref{thm:rhoerrbdd} yields the desired results.
\end{proof}

\subsubsection{Proofs for Section~\ref{sec:genbdd}}\label{app:wassbdd}

We first prove the auxiliary estimates used in the generalization bound. The proof of Theorem~\ref{thm:wassbdd} appears at the end of this subsection, after these lemmas.

First, we quote a  bound on the expected suprema of empirical
processes indexed by Vapnik-\v{C}ervonenkis classes of functions, from Proposition 2.1  in \citet{gine2001consistency} and Theorem A.2 in \citet{sriperumbudur2012consistency}.
\begin{lemma}[\citet{gine2001consistency, sriperumbudur2012consistency, kim2019uniform}]\label{lem:vcbdd}
    Let $(X, P)$ be a probability space and let $X_1, \cdots, X_n$ be i.i.d. samples from $P$. Let $\calG$ be a class of functions from $X$ to $\R$. Suppose $\calG$ is a uniformly bounded VC-class with dimension $\nu>0$, i.e., there exist $A, B>0$ such that for all $f\in \calG$, we have $\|f\|_\infty \le B$ and the covering numbers satisfy
    \[ \calN(\calG, L_2(Q), \epsilon) \le \Big( \frac{AB}{\epsilon}\Big)^\nu \]
    for every probability measure $Q$ on $X$ and every $0\le \epsilon \le B$. Let $0 < \sigma \le B$ be a positive constant such that $\EE_P f^2 \le \sigma^2$ for all $f\in \calG$. Then there exists a universal constant $C$ not depending on any parameters such that 
    \[\EE_P\left[ \sup_{f\in \calG}\Big|\frac{1}{n}\sum_{i=1}^n  \big(f(X_i)-\EE_P f(X_1)\big) \Big| \right] \le C\left( \frac{\nu B}{n} \log \Big( \frac{AB}{\sigma} \Big) + \sqrt{\frac{\nu \sigma^2}{n} \log \Big( \frac{AB}{\sigma} \Big)}\right).\]
\end{lemma}

Denote $q_{n, \delta}$ as the density function of $\Pi_{\calM \#}(\mu_n \ast \eta_\delta)$ with respect to the volume form of $\calM$. Using a similar derivation as \eqref{eqn:barrhosigma}, we have 
\begin{align*}
    q_{n, \delta}(x)  & =  \int_{B_{\R^{D-d}}(0,\delta)} \mu_n \ast \eta_\delta \left( x+ tN|_x \right) \det (I_d-t S_N|_{x})\, d \tt \\
    & = \frac{1}{n}\sum_{i=1}^n \int_{B_{\R^{D-d}}(0,\delta)} \eta_\delta \left( x+ tN|_x - X_i \right) \det (I_d-t S_N|_{x})\, d \tt.
\end{align*}
For any $x\in \calM$, denote the expectation of $q_{n, \delta}(x)$ by
\[q_\delta(x) := \EE_{X_{1:n}} q_{n, \delta}(x) = \int_\calM \int_{B_{\R^{D-d}}(0,\delta)} \eta_\delta \left( x+ tN|_x - y \right) \det (I_d-t S_N|_{x})\, d \tt \rho(y)\, d y. \]

Define the projected kernel by
\[ K_\delta(x, y) = \int_{B_{\R^{D-d}}(0,\delta)} \eta_\delta \left( x+ tN|_x - y \right) \det (I_d-t S_N|_{x})\, d \tt \qquad \text{for any} \; x,y\in \calM. \]
Then we have for $x\in \calM$
\begin{equation}\label{eqn:qndeltaqd}
    q_{n, \delta}(x) = \frac{1}{n}\sum_{i=1}^n K_\delta(x, X_i) \q \text{and} \q q_\delta(x) =\EE_\mu[K_\delta(x, X_1)].
\end{equation}

Let $i_0$ be the injectivity radius of $\calM$ endowed with the induced Riemannian metric, and let $K$ be the maximum of the absolute value of sectional curvature of $\calM$. In the following lemma, we derive a uniform stochastic bound between $q_{n,\delta}$ and $q_\delta$.

\begin{lemma}\label{lem:qndeltaconcen}
    Let $X_i \overset{i.i.d.}{\sim}\mu$ with $i=1,\cdots,n$. Define $q_{n,\delta}:\calM \mapsto \R$ and $q_{\delta}:\calM \mapsto \R$ by \eqref{eqn:qndeltaqd}. Then for every $t>0$, there exist some $C_{\calM, d, \eta, \omega_{D-d}}>0$ such that
    \[ \PP\left(\|q_{n, \delta} - q_\delta \|_{L^\infty(\calM)}  \ge C_{\calM, d, \eta, \omega_{D-d}}\left( \frac{\log (\delta^{-d} ) + t}{n \delta^d}  + \sqrt{\frac{\log (\delta^{-d} ) + t}{n \delta^d}}\right) \right) \le e^{-t}.\]
    Moreover, if we choose $\delta\ge n^{-1/d}$, then for every $\gamma>0$,
    \begin{equation}\label{eqn:qndeltabdd}
        \|q_{n,\delta} - q_\delta\|_{L^\infty(\calM)} \le C_{\calM, d, \eta, \omega_{D-d}} (\gamma + 1) \left( \sqrt{\frac{\log n}{n \delta^d}} + \frac{\log n}{n \delta^d} \right)
    \end{equation}
    holds with probability at least $1 - n^{-\gamma}$. 
\end{lemma}
\begin{proof}
    First, we show that kernel $K_\delta$ is bounded with a controlled Lipschitz constant and a variance scale for fixed $\delta$. Let $\{\kappa_i^N(x)\}_{i=1,\cdots,d}$ be real eigenvalues for $S_N|_{x}$, we have
    \[ \det (I_d- t S_N|_{x}) = \Pi_{i=1}^d (1- |\tt| \kappa_i^N(x)). \]
    Since $\sup_{x \in \calM}\sup_{i\in [d]} |\kappa_i^N(x)| \le 1/r_\calM$ and $|\tt|\le \delta \le r_\calM / 2$, it is straightforward to show $|\tt| |\kappa_i^N(x)| \le 1/2$ and hence 
    \[ 2^{-d} \le \det (I_d- t S_N|_{x}) \le (3 / 2)^d.  \]
    
    Note that $\eta_\delta$ is supported on $B_{\R^D}(0, \delta)$, then for any $x, y \in \mathrm{Supp}(K_\delta)$, we have
    \[|x - y| \le |x+ tN|_x - y| + |tN|_x| \le 2 \delta.\]
     For any fixed $0<\delta \le r_\calM / 2$, we can derive the upper bound
    \begin{align}
        K_\delta(x, y) & = \int_{B_{\R^{D-d}}(0,\delta)} \delta^{-D} \eta \left( \frac{|x+ tN|_x - y|}{\delta} \right) \det (I_d-t S_N|_{x})\, d \tt \notag \\
        & \le \delta^{-D} \|\eta\|_{\infty} (3 / 2)^d|B_{\R^{D-d}}(0,\delta)| = \delta^{-d}\|\eta\|_{\infty} (3 / 2)^d\omega_{D-d}. \label{eqn:kdeltaupb}
    \end{align}
    
Since $\calM$ is a compact smooth embedded manifold without boundary, equipped with the induced Riemannian metric. Then it has bounded sectional curvature and hence its Ricci curvature is bounded below. It follows from Bishop–Gromov volume comparison theorem \citep{bishop2011geometry} and Equation (1.35) in \citet{garcia2020error} that there exist some $r_1>0$ and $c_1, C_1 > 0$ depending on $d, K, \omega_d, i_0$ such that
\begin{equation}\label{eqn:Mballbdd}
    c_1 r^d \le \mathrm{Vol}(B_\calM (x, r)) \le C_1 r^d \qquad \text{for all } 0<r<\min\{i_0, K^{-1/2}\} \text{ and } x\in\calM.
\end{equation}

Using the fact that $1 / \rho_0 \le \rho \le \rho_0$, one can find
    \begin{align*}
    \EE_{\mu}|K_\delta(x, X_1)|^2 &  = \int_\calM \left( \int_{B_{\R^{D-d}}(0,\delta)} \delta^{-D} \eta \left(\frac{ |x+ tN|_x - y|}{\delta} \right) \det (I_d-t S_N|_{x})\, d \tt\right)^2\rho(y)\, dy \\
    & \le \delta^{-2D} \|\eta\|^2_\infty (3/2)^{2d} \rho_0 \mathrm{Vol}(B_\calM(x, 2\delta))|B_{\R^{D-d}}(0,\delta)|^2 \\
    &\le \delta^{-d} \|\eta\|^2_\infty 3^{2d} 2^{-d} \rho_0 C_1 \omega_{D-d}^2.
\end{align*}
In order to estimate $\mathrm{Lip}_x K_\delta(x, y)$, we can compute for any $x_1, x_2 \in \calM$
\begin{align*}
    I_K := & K_\delta(x_1, y) - K_\delta(x_2, y) \\
    = \; &  \int_{B_{\R^{D-d}}(0,\delta)} \big( \eta_\delta \left( x_1 + tN|_{x_1} - y \right) - \eta_\delta \left( x_2 + tN|_{x_2} - y \right) \big) \det (I_d-t S_N|_{x_1})\, d \tt \\
    & + \int_{B_{\R^{D-d}}(0,\delta)} \eta_\delta \left( x_2 + tN|_{x_2} - y \right) \big( \det (I_d-t S_N|_{x_1}) - \det (I_d-t S_N|_{x_2}) \big)\, d \tt.
\end{align*}
Since the normal frame is smooth on a compact chart, there exists some constant $C_\calM>0$ such that for any $x_1, x_2 \in \calM$
\begin{align*}
   \big|\eta_\delta \left( x_1 + tN|_{x_1} - y \right) - \eta_\delta \left( x_2 + tN|_{x_2} - y \right) \big| & \le \|\nabla \eta_\delta\|_\infty \big| x_1 + tN|_{x_1} - x_2 - tN|_{x_2} \big| \\
    & \le \delta^{-D-1}\|\eta'\|_\infty C_\calM (1 + t) d_\calM(x_1, x_2) \\
    &\le \delta^{-D-1}\|\eta'\|_\infty C_\calM (1 + r_\calM / 2) d_\calM(x_1, x_2).
\end{align*}
Also, since the determinant is Lipschitz and the shape operator is differentiable, there exists some constant $C_{\calM}>0$ such that for any $x_1, x_2 \in \calM$
\begin{align*}
    & \big| \det (I_d-t S_N|_{x_1}) - \det (I_d-t S_N|_{x_2}) \big| \\
    \le \; & \|\nabla_x\det (I_d-t S_N|_{x})\|_\infty \big| t S_N|_{x_1} - t S_N|_{x_2} \big| \\
    \le \; & d (1 + t \kappa_{\mathrm{max}})^{d-1} t \sup_{x\in \calM} \sup_{\substack{N\in N_x\calM, \, \|N\|=1}}\|\nabla_x S_N|_x\| d_\calM(x_1, x_2)\\
    \le \; & C_\calM d (3 /2)^{d-1} (r_\calM / 2)d_\calM(x_1, x_2).
\end{align*}
Collecting previous estimates yields
\begin{align*}
    |I_K|\le \; &  C_\calM d_\calM(x_1, x_2) \delta^{-D-1} (3/2)^d\int_{B_{\R^{D-d}}(0,\delta)} \|\eta'\|_\infty (1 + r_\calM / 2)  + \delta d\|\eta\|_\infty  r_\calM / 3\, d\tt \\
    \le \; & C_\calM  (3/2)^d \omega_{D-d}(\|\eta'\|_\infty (1 + r_\calM / 2)  + d\|\eta\|_\infty  r_\calM^2 / 3) \delta^{-d-1} d_\calM(x_1, x_2)\\
    =: \; & C_{\calM, D, d, \eta} \delta^{-d-1} d_\calM(x_1, x_2),
\end{align*}
which implies that $\mathrm{Lip}_x K_\delta(x, y)\le C_{\calM, D, d, \eta} \delta^{-d-1}$.

For fixed $\delta>0$, define a class of functions from $\calM$ to $\R$ by
\[ \calG_\delta=\{f_x(\cdot)=K_\delta(x, \cdot) - \EE_\mu K_\delta(x, X_1):\; x\in \calM \}. \]
We claim that $\calG_\delta$ is a VC-type kernel class with envelope $U_\delta \sim \delta^{-d}$.
Applying bounds on $K_\delta$ we have for any $f_x \in \calG_\delta$
\[\|f_x\|_{\infty} \le 2\|K_\delta\|_\infty \le 2\delta^{-d}\|\eta\|_{\infty} (3 / 2)^d\omega_{D-d}=:U_\delta \]
and 
\begin{align*}
    \EE_\mu f_x^2 & = \EE_\mu \big| K_\delta(x, X_1) - \EE_\mu K_\delta(x, X_1) \big|^2 \le \EE_\mu \big| K_\delta(x, X_1) \big|^2 \\
    & \le \delta^{-d} \|\eta\|^2_\infty 3^{2d} 2^{-d} \rho_0 C_1 \omega_{D-d}^2=:\sigma_\delta^2.
\end{align*}

For any probability measure $Q$ on $\calM$, we proceed to show the covering numbers satisfy
\begin{equation}\label{eqn:covernumGdelta}
    \calN(\calG_\delta, L^2(Q), \epsilon U_\delta)\le C_\calN (\epsilon \delta)^{-d} \qquad \text{for every} \; 0<\epsilon \le 1.
\end{equation}
Consider a $r$-net $\{x_1, \cdots, x_\ell\} \subset \calM$ with $r:= \epsilon U_\delta / (2C_{\calM, D, d, \eta} \delta^{-d-1})$. Let $x\in \calM$ and choose $x_i$ in the $r$-net such that $d_\calM(x, x_i)\le r$. Then it follows from the Lipschitz bound $\mathrm{Lip}_x K_\delta(x, y)\le C_{\calM, D, d, \eta} \delta^{-d-1}$ that 
\begin{align*}
    \|f_{x} - f_{x_i}\|_{L^2(Q)} & \le \sup_{y\in \calM}| K_\delta(x, y) - \EE_\mu K_\delta(x, X_1) - K_\delta(x_i, y) + \EE_\mu K_\delta(x_i, X_1)| \\
    & \le 2\sup_{y\in \calM}| K_\delta(x, y) -  K_\delta(x_i, y)| \le 2 C_{\calM, D, d, \eta} \delta^{-d-1} d_\calM(x, x_i)\\
    & \le 2 C_{\calM, D, d, \eta} \delta^{-d-1} r= \epsilon U_\delta.
\end{align*}
Since $\calM$ is a compact $d$-dimensional manifold, there exists some $C_{\mathrm{cov}}>0$ depending on the geometry of $\calM$ such that for small $r>0$, the covering number satisfies $\calN(\calM, d_\calM, r) \le C_{\mathrm{cov}} r^{-d}$. Therefore, we have 
\begin{align*}
    \calN(\calG_\delta, L^2(Q), \epsilon U_\delta)& \le \calN(\calM, d_\calM, r) \le C_{\mathrm{cov}} r^{-d} \\
    & = C_{\mathrm{cov}}  \|\eta\|_{\infty}^{-d} (3 / 2)^{-d^2}\omega_{D-d}^{-d} C_{\calM, D, d, \eta}^d (\epsilon \delta)^{-d}=:C_\calN (\epsilon \delta)^{-d}. 
\end{align*}
Hence, for fixed $\delta>0$, $\calG_\delta$ is a uniformly bounded VC-class with dimension $d$. 

Applying Lemma~\ref{lem:vcbdd} on $\calG_\delta$ with \eqref{eqn:covernumGdelta} and
\[\nu=d, \q A=C_\calN^{1/d} \delta^{-1}, \q B=U_\delta, \q \sigma^2 = \sigma_\delta^2\]
gives
\begin{align}
    \EE_{X_{1:n}}\left[ \|q_{n, \delta} - q_\delta \|_{L^\infty(\calM)} \right] & = \EE_\mu\left[ \sup_{x\in \calM}\Big|\frac{1}{n}\sum_{i=1}^n K_\delta(x, X_i) - \EE_\mu K_\delta(x, X_1) \Big| \right] \notag \\
    & \le C\left( \frac{\nu B}{n} \log \Big( \frac{AB}{\sigma} \Big) + \sqrt{\frac{\nu \sigma^2}{n} \log \Big( \frac{AB}{\sigma} \Big)}\right) \notag\\
    & \le C\left( n^{-1}\delta^{-d} \log (\delta^{-d} ) + \sqrt{n^{-1}\delta^{-d} \log (\delta^{-d} )}\right) \label{eqn:expsupbdd}
\end{align}
where $C>0$ is a constant depending on $\calM, d, \eta, \omega_{D-d}$.

Furthermore, denote \[Z:=\sup_{f_x\in \calG_\delta}\Big|\sum_{i=1}^n f_x(X_i) \Big| = \sup_{x\in \calM}\Big|\sum_{i=1}^n (K_\delta(x, X_i) - \EE_\mu K_\delta(x, X_1)) \Big|\] 
with $\|f_x\|_\infty \le U_\delta$ and $\EE_\mu f_x^2\le \sigma_\delta^2$. We apply Talagrand's concentration inequality \citep{massart2000constants, bousquet2002bennett} for empirical process to obtain for every $t>0$
\[ \PP\left( Z \ge \EE Z + \sqrt{2t(2U_\delta \EE Z + n \sigma_\delta^2)} + \frac{U_\delta t}{3} \right) \le e^{-t}. \]
By a straightforward computation via Young's inequality, we have 
\[ \sqrt{2t(2U_\delta \EE Z + n \sigma_\delta^2)} \le \sqrt{4tU_\delta \EE Z} + \sqrt{2t n \sigma_\delta^2} \le \EE Z + U_\delta t + \sqrt{2t n \sigma_\delta^2} \]
and hence
\begin{equation}\label{eqn:talagrand}
    \PP\left( \frac{Z}{n} \ge \frac{2\EE Z}{n} + \sqrt{\frac{2t \sigma_\delta^2}{n}} + \frac{4U_\delta t}{3n} \right) = \PP\left( Z \ge 2\EE Z + \sqrt{2tn \sigma_\delta^2} + \frac{4U_\delta t}{3} \right) \le e^{-t}.
\end{equation}
Combining \eqref{eqn:expsupbdd}--\eqref{eqn:talagrand} and plugging in $U_\delta, \sigma_\delta^2$ yields there exists some $C>0$ depending on $\calM, d, \eta, \omega_{D-d}$ such that 
\[ \|q_{n, \delta} - q_\delta \|_{L^\infty(\calM)} \ge C\left( \frac{\log (\delta^{-d} )}{n \delta^d}  + \sqrt{\frac{\log (\delta^{-d} )}{n \delta^d}}\right) + C\left(\sqrt{\frac{t }{n \delta^d}} + \frac{t}{n \delta^d} \right) \]
holds with probability at most $e^{-t}$ for every $t>0$. Moreover, we have 
\[ \PP\left(\|q_{n, \delta} - q_\delta \|_{L^\infty(\calM)}  \ge C\left( \frac{\log (\delta^{-d} ) + t}{n \delta^d}  + \sqrt{\frac{\log (\delta^{-d} ) + t}{n \delta^d}}\right) \right) \le e^{-t}\]
as desired. For every $\gamma>0$, we can choose $t=\gamma\log n$ to obtain \eqref{eqn:qndeltabdd}.

\end{proof}

Next, we apply Theorem 2 in \citet{garcia2020error} and Lemma~\ref{lem:qndeltaconcen} to show the empirical rate for the push-forward measure $\Pi_{\calM \#}(\mu_n \ast \eta_\delta)$ in $\infty$-Wasserstein distance.

\begin{lemma}[Lemma~\ref{lem:empratewinf}]\label{lem:empratePi}
    Let $X_i \overset{i.i.d.}{\sim}\mu$ with $i=1,\cdots,n$ and $Y_j \overset{i.i.d.}{\sim} \mu_n\ast \eta_\delta$ with $j=1,\cdots,N$, where $\mu_n$ is the empirical measure of $X_{1:n}$. Assume Assumption~\ref{assumpmu} holds and there exists some $r_0 \in (0, 1)$ such that $\inf_{|x|\le r_0} \eta(x) > 0$. For every $\gamma>0$, if $\delta \ge C_\gamma n^{-1/d}(\log n)^{1/d}$ with $C_\gamma>0$ depending on $\calM, d, \eta, \omega_{D-d}, \gamma$, there exists some  $C_{\eta, \mathrm{vol}(\calM), d, i_0}>0$ such that for every $\beta>1$
    \[ \calW_\infty\left( \Pi_{\calM \#}(\mu_n \ast \eta_\delta), \frac{1}{N}\sum_{j=1}^N \delta_{\Pi_{\calM}(Y_j)} \right) \le A \left\{\begin{array}{ll}
        N^{-1/2} (\log N)^{3/4}, & d=2, \\
         N^{-1/d} (\log N)^{1/d}, & d\ge 3,
    \end{array}\right. \]
    holds with probability at least $ 1- n^{-\gamma}- C_{\eta, \mathrm{vol}(\calM), d, i_0} N^{-\beta}$, where $A>0$ depends on $\eta, \calM, d, \rho_0, \omega_{D-d}, \beta$.
\end{lemma}

\begin{proof}
    First, we claim that $q_\delta$ is bounded from above and below in order to apply results in \citet{garcia2020error}. As we showed in the proof of Lemma~\ref{lem:qndeltaconcen}, for any $x, y \in \mathrm{Supp}(K_\delta)$, we have $|x-y|\le 2 \delta$ and $K_\delta(x, y)\le \delta^{-d}\|\eta\|_{\infty} (3 / 2)^d\omega_{D-d}$. Using the fact that $1 / \rho_0 \le \rho \le \rho_0$ and \eqref{eqn:Mballbdd}, we can derive the upper bound 
    \begin{align*}
        q_\delta(x) & =   \int_\calM K_\delta(x, y) \rho(y)  \, d y =   \int_{B_{\calM}(x, 2\delta)} K_\delta(x, y) \rho(y)  \, d y\\
        & \le \delta^{-d}\|\eta\|_{\infty} (3 / 2)^d\omega_{D-d} \rho_0 \mathrm{Vol}(B_{\calM}(x, 2\delta)) \\
        &\le 3^d \rho_0 C_1 \|\eta\|_{\infty} \omega_{D-d}=:C_q.
    \end{align*}

    On the other hand, recall that there exists some $r_0 \in (0, 1)$ such that $\inf_{|x|\le r_0} \eta(x) > 0$. Choosing $t \le r_0 \delta /2$ and $|x-y|\le r_0 \delta /2$ gives $|x+ tN|_x - y| / \delta \le r_0$, then we can derive the lower bound with \eqref{eqn:Mballbdd}
    \begin{align*}
        q_\delta(x) & = \int_\calM \int_{B_{\R^{D-d}}(0,\delta)} \eta_\delta \left( x+ tN|_x - y \right) \det (I_d-t S_N|_{x})\, d \tt \rho(y)\, d y \\
        & \ge  2^{-d} \rho_0^{-1} \int_\calM \int_{B_{\R^{D-d}}(0,\delta)} \eta_\delta \left( x+ tN|_x - y \right) \, d \tt  \, d y \\
        & \ge  2^{-d} \rho_0^{-1} \delta^{-D} \int_{\calM \cap B_{\calM}(x, r_0 \delta /2)} \int_{B_{\R^{D-d}}(0, r_0 \delta /2)} \eta \left( \frac{x+ tN|_x - y}{\delta} \right) \, d \tt  \, d y \\
        & \ge 2^{-d} \rho_0^{-1} \delta^{-D} \inf_{|x|\le r_0} \eta(x) \mathrm{Vol}(B_{\calM}(x, r_0 \delta /2)) |B_{\R^{D-d}}(0,r_0 \delta /2)|\\
        & \ge 2^{-2d} \rho_0^{-1} c_1 r_0^{D} \inf_{|x|\le r_0} \eta(x) \omega_{D-d}=:c_q .
    \end{align*}

    According to Lemma~\ref{lem:qndeltaconcen}, when $\delta \ge n^{-1/d}$, for every $\gamma>0$
    \[\|q_{n,\delta} - q_\delta\|_{L^\infty(\calM)} \le C_{\calM, d, \eta, \omega_{D-d}} (\gamma + 1) \left( \sqrt{\frac{\log n}{n \delta^d}} + \frac{\log n}{n \delta^d} \right)\]
    holds with probability at least $1 - n^{-\gamma}$. By setting $\delta \ge C_\gamma n^{-1/d}(\log n)^{1/d}$ with 
    \[ C_\gamma = \left( \frac{2C_{\calM, d, \eta, \omega_{D-d}} (\gamma + 1) }{c_q} \right)^{1/d} + \left( \frac{2C_{\calM, d, \eta, \omega_{D-d}} (\gamma + 1) }{c_q} \right)^{2/d}, \]
    we have $\|q_{n,\delta} - q_\delta\|_{L^\infty(\calM)} \le c_q / 2$ holds with probability at least $1 - n^{-\gamma}$, and hence
    \[  \frac{c_q}{2} \le q_{n,\delta}(x) \le C_q + \frac{c_q}{2} \qquad \text{for all} \; x \in \calM. \]
    Then we can apply Theorem 2 in \citet{garcia2020error} on $\Pi_{\calM \#}(\mu_n \ast \eta_\delta)$ to get for every $\beta>1$, there exists some $A>0$ depending on $\eta, \calM, \rho_0, d, \beta$ such that 
    \[ \calW_\infty\left( \Pi_{\calM \#}(\mu_n \ast \eta_\delta), \frac{1}{N}\sum_{j=1}^N \delta_{\Pi_{\calM}(Y_j)} \right) \le A \left\{\begin{array}{ll}
        N^{-1/2} (\log N)^{3/4}, & d=2, \\
         N^{-1/d} (\log N)^{1/d}, & d\ge 3,
    \end{array}\right. \]
    holds with probability at least $ 1 -C_{\eta, \mathrm{vol}(\calM), d, i_0} N^{-\beta}$. 
\end{proof}

Furthermore, we derive the bounds of $\calW_2(\Pi_{\calM \#}(\mu\ast\eta_\delta), \Pi_{\calM \#}(\mu_n\ast\eta_\delta))$ through the weighted homogeneous Sobolev norm between $q_{n, \delta}$ and $q_\delta$, and then estimate it by the Green's function of the Laplacian.

\begin{lemma}[Lemma~\ref{lem:w2smooth}]\label{lem:w2smoothPi}
Let $\mu_n$ be the empirical measure of $X_i \overset{i.i.d.}{\sim}\mu$ with $i=1,\cdots,n$. Assume Assumption~\ref{assumpmu} holds and there exists some $r_0 \in (0, 1)$ such that $\inf_{|x|\le r_0} \eta(x) > 0$. Then there exists some constant $B>0$ depending on $\calM, \eta, \rho_0, d, \omega_{D-d}$ such that 
\begin{equation}\label{eqn:w2smoothexpe}
  \EE \calW_2\left(\Pi_{\calM \#}(\mu\ast\eta_\delta), \Pi_{\calM \#}(\mu_n\ast\eta_\delta)\right) \le B \left\{\begin{array}{ll}
      n^{-1/2} \log(\delta^{-1}), &  d =2, \\
      n^{-1/2}\delta^{1-d/2}, & d\ge 3.
  \end{array}\right.  
\end{equation}
Moreover, for any $\alpha>0$, 
\begin{equation}\label{eqn:w2smoothhigh}
  \calW_2\left(\Pi_{\calM \#}(\mu\ast\eta_\delta), \Pi_{\calM \#}(\mu_n\ast\eta_\delta)\right) \le B (1+ \sqrt{2\alpha}) \left\{\begin{array}{ll}
      n^{-1/2}(\log n)^{1/2} \log(\delta^{-1}), &  d =2, \\
      n^{-1/2}(\log n)^{1/2} \delta^{1-d/2}, & d\ge 3,
  \end{array}\right.  
\end{equation}
holds with probability at least $1-n^{-\alpha}$.
\end{lemma}
\begin{proof}
As we have shown in the proof of Lemma~\ref{lem:empratePi}, there exist 
\[ C_q = 3^d \rho_0 C_1 \|\eta\|_{\infty} \omega_{D-d} \q \text{and} \q c_q = 2^{-2d} \rho_0^{-1} c_1 r_0^{D} \inf_{|x|\le r_0} \eta(x) \omega_{D-d} \]
such that $c_q \le q_\delta(x) \le C_q$ for every $x\in \calM$. Recall that $q_\delta, q_{n,\delta}$ are densities of $\Pi_{\calM \#}(\mu\ast\eta_\delta)$ and $\Pi_{\calM \#}(\mu_n\ast\eta_\delta)$ with respect to the volume form of $\calM$, respectively. It follows from Corollary~2.3 in \citet{peyre2018} that 
\begin{equation}\label{eqn:w2h1bdd}
    \calW_2^2(\Pi_{\calM \#}(\mu\ast\eta_\delta), \Pi_{\calM \#}(\mu_n\ast\eta_\delta)) \le 4 c_q^{-1} \|q_{\delta} - q_{n, \delta}\|_{\dot{H}^{-1}(\calM)}^2,
\end{equation}
where
\[ \|f\|_{\dot{H}^{-1}(\calM)}:= \sup\left\{ \int_{\calM} f \varphi \, d\mathrm{vol}_\calM :\; \varphi \in C^\infty(\calM), \, \int_\calM \varphi \, d\mathrm{vol}_\calM=0,\, \|\nabla \varphi\|_{L^2(\calM)}\le 1\right\}. \]
Fix $y\in \calM$, $K_\delta(\cdot, y)$ is the density of $\Pi_{\calM \#}(\delta_y\ast\eta_\delta)$ w.r.t. $\mathrm{vol}_\calM$ and then satisfies
\begin{equation}\label{eqn:kdeltaint}
     \int_{\calM} K_\delta(x, y) \, d\mathrm{vol}_\calM(x) = 1 \qquad \text{for any} \; y\in \calM. 
\end{equation}
For $i=1,\cdots,n$, denote
\[Z_{i}(x):= K_\delta(x, X_i) - \mathrm{Vol}(\calM)^{-1},\]
then we have $\int_\calM Z_i(x) \, d\mathrm{vol}_\calM(x)=0$ and hence $Z_i\in \dot{H}^{-1}(\calM)$. We can rewrite
\begin{align*}
    \|q_{\delta} - q_{n, \delta}\|_{\dot{H}^{-1}(\calM)} & = \Big\|\frac{1}{n}\sum_{i=1}^n K_\delta(x, X_i) - \EE_\mu K_\delta(x, X_1) \Big\|_{\dot{H}^{-1}(\calM)} \\
    & = \Big\|\frac{1}{n}\sum_{i=1}^n \Big( Z_i(x) - \EE_\mu Z_i(x) \Big) \Big\|_{\dot{H}^{-1}(\calM)}.
\end{align*}
Since $\dot{H}^{-1}(\calM)$ is a Hilbert space and $Z_i - \EE_\mu Z_i$ are i.i.d. centered random variables, one can find 
\begin{align*}
    \EE_{X_{1:n}}\|q_{\delta} - q_{n, \delta}\|_{\dot{H}^{-1}(\calM)}^2 & = \EE_{X_{1:n}}\Big\|\frac{1}{n}\sum_{i=1}^n \Big( Z_i(x) - \EE_\mu Z_i(x) \Big) \Big\|_{\dot{H}^{-1}(\calM)}^2 \\
    & = \frac{1}{n^2} \sum_{i,j =1}^n\EE_{X_{1:n}} \big\langle Z_i - \EE_\mu Z_i, Z_j - \EE_\mu Z_j \big\rangle_{\dot{H}^{-1}(\calM)} \\
    & = \frac{1}{n} \EE_{\mu}\big\|Z_1 - \EE_\mu Z_1 \big\|_{\dot{H}^{-1}(\calM)}^2 \\
    &\le \frac{1}{n} \EE_{\mu}\|Z_1 \|_{\dot{H}^{-1}(\calM)}^2. 
\end{align*}

In order to estimate $\|Z_1 \|_{\dot{H}^{-1}(\calM)}^2$, we introduce $G(x,y): \calM\times \calM\setminus \mathrm{diag}(\calM) \mapsto \R$ as the Green's function of the Laplace-Beltrami operator on $\calM$. It satisfies $G(x,y)=G(y,x)$, 
\[\Delta_x G(x,y)= \delta_y(x)- \mathrm{Vol}(\calM)^{-1} \quad \text{and} \quad \int_\calM G(x,y)\, d\mathrm{vol}_\calM(x)=0 \quad \text{for every}\; y\in\calM. \]
Let $u:C^\infty(\calM)\mapsto \R$ be a solution of 
\[ -\Delta u(x) =  Z_1(x) \q \text{and} \q \int_\calM u(x)\, d\mathrm{vol}_\calM(x)=0, \]
then we have 
\begin{equation}\label{eqn:ux}
   u(x) = -\int_\calM G(x, y) Z_1(y) \, d\mathrm{vol}_\calM(y). 
\end{equation}
For any test function $\varphi\in C^\infty(\calM)$ satisfying $\int_\calM \varphi = 0$ and $\|\nabla \varphi\|_{L^2(\calM)}\le 1$, we can compute by Green's identities and the Cauchy-Schwarz inequality that
\begin{align*}
    \int_\calM Z_1(x) \varphi(x) \, d\mathrm{vol}_\calM(x)& = -\int_\calM \Delta u(x) \varphi(x) \, d\mathrm{vol}_\calM(x) \\
    & = \int_\calM \langle \nabla u, \nabla \varphi \rangle_x \, d\mathrm{vol}_\calM(x) \le \|\nabla u\|_{L^2(\calM)},
\end{align*}
where the equality holds if $\varphi = u / \|\nabla u\|_{L^2(\calM)}$.
Therefore, by the definition of $\dot{H}^{-1}(\calM)$ and \eqref{eqn:ux}, we have
\begin{align*}
    \|Z_1\|_{\dot{H}^{-1}(\calM)}^2 & = \|\nabla u\|_{L^2(\calM)}^2 = -\int_\calM u(x) \Delta u(x) \, d\mathrm{vol}_\calM(x) = \int_\calM u Z_1(x) \, d\mathrm{vol}_\calM(x)\\
    & = -\int_{\calM\times\calM}G(x, y) Z_1(x) Z_1(y)  d\mathrm{vol}_\calM(y)d\mathrm{vol}_\calM(x) \\
    & = -\int_{\calM\times\calM} G(x, y) \big( K_\delta(x, X_1) - \mathrm{Vol}(\calM)^{-1} \big) \big( K_\delta(y, X_1) - \mathrm{Vol}(\calM)^{-1} \big)  d\mathrm{vol}_\calM(y)d\mathrm{vol}_\calM(x) \\
    & = -\int_{\calM\times\calM} G(x, y)  K_\delta(x, X_1) K_\delta(y, X_1) 
    d\mathrm{vol}_\calM(y)d\mathrm{vol}_\calM(x),
\end{align*}
where we use the property $\int_\calM G(x,y)\, d\mathrm{vol}_\calM(x)=0$ for the last equality.

Since $\calM$ is compact and smooth, it follows from Theorem~4.13 in \citet{aubin1998some} that there exists some constant $C>0$ such that 
\[ |G(x,y)| \le C \left\{ \begin{array}{ll}
   (1 + |\log (d_\calM(x,y))|),  & d=2, \\
    d_\calM(x,y)^{2-d}, &  d\ge 3.
\end{array} \right. \]

Using \eqref{eqn:kdeltaupb}, \eqref{eqn:Mballbdd}, and \eqref{eqn:kdeltaint}, we integrate the Green function singularity near the diagonal. The constants $C$ below are independent of $\delta$. For $d\ge 3$, geodesic polar integration gives
\begin{align*}
    \|Z_1\|_{\dot{H}^{-1}(\calM)}^2 & \le C\int_{B_\calM(X_1, 2\delta)}\int_{B_\calM(X_1, 2\delta)} d_\calM(x,y)^{2-d} K_\delta(x, X_1) K_\delta(y, X_1) \, d\mathrm{vol}_\calM(y)d\mathrm{vol}_\calM(x) \\
    & \le C\delta^{-d}\int_{B_\calM(X_1, 2\delta)} K_\delta(x, X_1)\,d\mathrm{vol}_\calM(x)
        \int_0^{4\delta} r^{2-d}r^{d-1}\,dr \\
    & \le C\delta^{-d}\int_0^{4\delta}r\,dr
      \le C\delta^{2-d}.
\end{align*}

On the other hand, for $d=2$, using a similar argument gives
\begin{align*}
    \|Z_1\|_{\dot{H}^{-1}(\calM)}^2 & \le C\int_{(B_\calM(X_1, 2\delta))^2} (1 + |\log (d_\calM(x,y))|) K_\delta(x, X_1) K_\delta(y, X_1) \, d\mathrm{vol}_\calM(y)d\mathrm{vol}_\calM(x) \\
    & \le C\delta^{-2}\int_{B_\calM(X_1, 2\delta)} K_\delta(x, X_1)\,d\mathrm{vol}_\calM(x)
        \int_0^{4\delta}(1+|\log r|)r\,dr \\
    & \le C\delta^{-2}\int_0^{4\delta}(1+|\log r|)r\,dr
      \le C(1+|\log(4\delta)|).
\end{align*}
Denote 
\[ C_\delta := C_{\calM,d,\eta,\omega_{D-d}}\left\{ \begin{array}{ll}
    (1 + |\log(4\delta)|)^{1/2}, & d=2, \\
    \delta^{1-d/2}, & d\ge 3.
\end{array} \right. \]

Then we can apply Jensen's inequality to get 
\begin{equation}\label{eqn:qdeltahomobdd}
    \EE_{X_{1:n}}\|q_{\delta} - q_{n, \delta}\|_{\dot{H}^{-1}(\calM)} \le \big(n^{-1}\EE_{\mu}\|Z_1 \|_{\dot{H}^{-1}(\calM)}^2\big)^{1/2} \le n^{-1/2} C_\delta.
\end{equation}
Combining it with \eqref{eqn:w2h1bdd} gives \eqref{eqn:w2smoothexpe}. 

Next we proceed to show high-probability bound via bounded difference inequality. Denote
\[ F(X_1, \cdots, X_n):= \Big\|\frac{1}{n}\sum_{i=1}^n \Big( Z_i(x) - \EE_\mu Z_i(x) \Big) \Big\|_{\dot{H}^{-1}(\calM)}=\|q_{\delta} - q_{n, \delta}\|_{\dot{H}^{-1}(\calM)}.\]
Let $X'_i$ be a random variable drawn from $\mu$ and independent of $X_{1:n}$. Then we replace $X_i$ by $X'_i$ in $F$ to obtain
\[|F(X_1, \cdots,X_i,\cdots, X_n) - F(X_1, \cdots,X'_i,\cdots, X_n)| 
    \le  \frac{1}{n} \| Z_i - Z'_i\|_{\dot{H}^{-1}(\calM)} \le \frac{2C_\delta}{n}. \]
Then it follows from McDiarmid's inequality \citep{mcdiarmid1989method} that for any $t>0$
\[\PP\left( F(X_1, \cdots, X_n) - \EE F(X_1, \cdots, X_n) \ge t \right) \le \exp\left( -\frac{n t^2}{2C_\delta^2} \right).\]
For any $\alpha>0$, choosing $t=C_\delta\sqrt{\frac{2\alpha \log n}{n}}$ yields
\[ \PP\left( \|q_{\delta} - q_{n, \delta}\|_{\dot{H}^{-1}(\calM)} - \EE\|q_{\delta} - q_{n, \delta}\|_{\dot{H}^{-1}(\calM)} \ge C_\delta\sqrt{\frac{2\alpha \log n}{n}} \right) \le n^{-\alpha}. \]
Combining the above inequality with \eqref{eqn:qdeltahomobdd} gives
\[ \PP\left( \|q_{\delta} - q_{n, \delta}\|_{\dot{H}^{-1}(\calM)}  \le C_\delta (1+ \sqrt{2\alpha}) n^{-1/2} (\log n)^{1/2} \right) \ge 1 - n^{-\alpha}. \]
Combining it with \eqref{eqn:w2h1bdd} gives \eqref{eqn:w2smoothhigh}. This completes the proof.

\end{proof}
\paragraph{Proof of Theorem~\ref{thm:wassbdd}.}
For convenience, we restate the theorem before proving it.
\begin{theorem}[Theorem~\ref{thm:wassbdd}]
Let $X_i \overset{i.i.d.}{\sim}\mu$ with $i=1,\cdots,n$ and $Y_j \overset{i.i.d.}{\sim} \mu_n\ast \eta_\delta$ with $j=1,\cdots,N$, where $\mu_n$ is the empirical measure of $X_{1:n}$. Assume that $T_\theta:\R^D \mapsto \R^D$ is measurable on $\calM_\delta$ and only depends on $X_{1:n}$. Let $\mu_{T_\theta}$ be the empirical measure of $\{T_\theta(Y_j)\}_{j=1}^N$. Assume Assumption~\ref{assumpmu} holds and there exists some $r_0 \in (0, 1)$ such that $\inf_{|x|\le r_0} \eta(x) > 0$. For every $\alpha>0$, if $\delta \ge C_\alpha n^{-1/d}(\log n)^{1/d}$ with $C_\alpha>0$ depending on $\calM, d, \eta, \omega_{D-d}, \alpha$, there exists some $C_{\eta, \mathrm{vol}(\calM), d, i_0}>0$ such that for every $\beta>1$ and $n\in \N$
\begin{equation}\label{eqn:wassbdd}
    \calW_2(\mu, \mu_{T_\theta}) \le C \left\{\begin{array}{ll}
        \delta^2  + n^{-1/2}(\log n)^{1/2} \log(\delta^{-1}) + N^{-1/2} (\log N)^{3/4} + \sqrt{\mathrm{PErr}(T_\theta)}, & d=2, \\
        \delta^2  + n^{-1/2}(\log n)^{1/2} \delta^{1-d/2} +  N^{-1/d} (\log N)^{1/d} + \sqrt{\mathrm{PErr}(T_\theta)}, & d\ge 3,
    \end{array}\right.
\end{equation}
holds with probability at least $1 - 2n^{-\alpha}- C_{\eta, \mathrm{vol}(\calM), d, i_0} N^{-\beta}$, where $C>0$ depends on $\eta, \calM, d, \rho_0$, $\omega_{D-d}, \alpha, \beta$.
\end{theorem}
\begin{proof}
    First we use the triangle inequality to decompose $\calW_2(\mu, \mu_{T_\theta})$ into four parts: 
    \begin{multline*}
        \calW_2(\mu, \mu_{T_\theta})  \lesssim  \; \calW_2(\mu, \Pi_{\calM \#}(\mu\ast\eta_\delta)) + \calW_2(\Pi_{\calM \#}(\mu\ast\eta_\delta), \Pi_{\calM \#}(\mu_n\ast\eta_\delta)) \\
         + \calW_2 \Big(\Pi_{\calM \#}(\mu_n\ast\eta_\delta), \frac{1}{N}\sum_{j=1}^N \delta_{\Pi_{\calM}(Y_j)}\Big) +  \calW_2\Big(\frac{1}{N}\sum_{j=1}^N \delta_{\Pi_{\calM}(Y_j)}, \mu_{T_\theta}\Big). 
    \end{multline*}

    According to Theorem~\ref{thm:errorbdd}, we have 
    \begin{equation}\label{eqn:w2bdd1}
        \calW_2(\mu, \Pi_{\calM \#}(\mu\ast\eta_\delta)) \lesssim \delta^2.
    \end{equation}

    For the second term, it follows from Lemma~\ref{lem:w2smoothPi} that for any $\alpha>0$ and $n\in \N$, there exists some $B_\alpha>0$ depending on $\calM, \eta, \rho_0, d, \omega_{D-d}, \alpha$ such that 
    \begin{equation}\label{eqn:w2bdd2}
      \calW_2(\Pi_{\calM \#}(\mu\ast\eta_\delta), \Pi_{\calM \#}(\mu_n\ast\eta_\delta)) \le B_\alpha \left\{\begin{array}{ll}
          n^{-1/2}(\log n)^{1/2} \log(\delta^{-1}), &  d =2, \\
          n^{-1/2}(\log n)^{1/2} \delta^{1-d/2}, & d\ge 3,
      \end{array}\right.  
    \end{equation}
    holds with probability at least $1-n^{-\alpha}$.

    Moreover, Lemma~\ref{lem:empratePi} shows that for every $\alpha>0$, if $\delta \ge C_\alpha n^{-1/d}(\log n)^{1/d}$ with $C_\alpha>0$ depending on $\calM, d, \eta, \omega_{D-d}, \alpha$, there exists some  $C_{\eta, \mathrm{vol}(\calM), d, i_0}>0$ such that for every $\beta>1$
    \begin{equation}\label{eqn:w2bdd3}
        \calW_2 \Big(\Pi_{\calM \#}(\mu_n\ast\eta_\delta), \frac{1}{N}\sum_{j=1}^N \delta_{\Pi_{\calM}(Y_j)}\Big)\le A \left\{\begin{array}{ll}
        N^{-1/2} (\log N)^{3/4}, & d=2, \\
         N^{-1/d} (\log N)^{1/d}, & d\ge 3,
    \end{array}\right. 
    \end{equation}
    holds with probability at least $ 1- n^{-\alpha}-C_{\eta, \mathrm{vol}(\calM), d, i_0} N^{-\beta}$, where $A>0$ depends on $\eta, \calM, d, \rho_0$, $ \omega_{D-d}, \beta$.
    
    Next, coupling $\Pi_\calM(Y_j)$ with $T_\theta(Y_j)$ for each $j$ gives
    \begin{align*}
        \calW_2^2\Big(\frac{1}{N}\sum_{j=1}^N \delta_{\Pi_{\calM}(Y_j)}, \mu_{T_\theta}\Big)
        & \le \frac{1}{N}\sum_{j=1}^N |\Pi_\calM(Y_j)-T_\theta(Y_j)|^2.
    \end{align*}

    By the definition of $\mathrm{PErr}$, we have 
    \[ \calW_2\Big(\frac{1}{N}\sum_{j=1}^N \delta_{\Pi_{\calM}(Y_j)}, \mu_{T_\theta}\Big) \le \sqrt{\mathrm{PErr}(T_\theta) \mathrm{tr}(\widehat{\Sigma}_n)}. \]
    Since $\widehat{\Sigma}_n$ is the unbiased covariance of training samples $X_{1:n}$, it is straightforward to show that 
    \[  \mathrm{tr}(\widehat{\Sigma}_n) = \frac{1}{2n(n-1)}\sum_{i,j=1}^n\|X_i - X_j\|^2 \le \frac{n^2}{2n(n-1)}\mathrm{diam}(\calM)^2 \le \mathrm{diam}(\calM)^2. \]
    Therefore, we have
    \begin{equation}\label{eqn:w2bdd4}
        \calW_2\Big(\frac{1}{N}\sum_{j=1}^N \delta_{\Pi_{\calM}(Y_j)}, \mu_{T_\theta}\Big) \le \mathrm{diam}(\calM)\sqrt{\mathrm{PErr}(T_\theta)}.
    \end{equation}
    
    Putting these estimates \eqref{eqn:w2bdd1}--\eqref{eqn:w2bdd4} together, we have \eqref{eqn:wassbdd} hold. This completes the proof.
\end{proof}

Alternatively, if we assume $T_\theta:\R^D \mapsto \R^D$ is a Lipschitz map on $\calM_\delta$, we have the following result:
\begin{lemma}
Let $X_i \overset{i.i.d.}{\sim}\mu$ with $i=1,\cdots,n$ and $Y_j \overset{i.i.d.}{\sim} \mu_n\ast \eta_\delta$ with $j=1,\cdots,N$, where $\mu_n$ is the empirical measure of $X_{1:n}$. Assume that $T_\theta:\R^D \mapsto \R^D$ is a Lipschitz map on $\calM_\delta$ and only depends on $X_{1:n}$. Let $\mu_{T_\theta}$ be the empirical measure of $\{T_\theta(Y_j)\}_{j=1}^N$. Assume Assumption~\ref{assumpmu} holds and there exists some $r_0 \in (0, 1)$ such that $\inf_{|x|\le r_0} \eta(x) > 0$. Then we have
\begin{equation}\label{eqn:wassbdd2}
    \mathbb{E}\calW_2(\mu, \mu_{T_\theta}) \lesssim \left\{\begin{array}{ll}
        \delta^2 + n^{-1/2} \log(\delta^{-1}) + \delta + \left(\frac{n}{N}\right)^{1/4} + \big(\mathbb{E}\mathrm{PErr}(T_\theta)\big)^{1/2}, & d=2, \\
        \delta^2 + n^{-1/2}\delta^{1-d/2} + \delta + \left(\frac{n}{N}\right)^{1/4} +\big(\mathbb{E}\mathrm{PErr}(T_\theta)\big)^{1/2}, & d \ge 3.
    \end{array}\right.
\end{equation}
\end{lemma}
\begin{proof}
First we decompose $\calW_2(\mu, \mu_{T_\theta})$ into four parts and get
\begin{align}
\calW_2(\mu, \mu_{T_\theta})   \lesssim  & \; \calW_2(\mu, \Pi_{\calM \#}(\mu\ast\eta_\delta)) + \calW_2(\Pi_{\calM \#}(\mu\ast\eta_\delta), \Pi_{\calM \#}(\mu_n\ast\eta_\delta))  \notag\\
 &  + \calW_2(\Pi_{\calM \#}(\mu_n\ast\eta_\delta), T_{\theta \#}(\mu_n\ast\eta_\delta)) +  \calW_2(T_{\theta \#}(\mu_n\ast\eta_\delta), \mu_{T_\theta}) \label{eqn:wassmuTtheta}
\end{align}

According to Theorem~\ref{thm:errorbdd}, we have 
\begin{equation}\label{eqn:w2bdd21}
    \calW_2(\mu, \Pi_{\calM \#}(\mu\ast\eta_\delta)) \lesssim \delta^2.
\end{equation}
And it follows from Lemma~\ref{lem:w2smoothPi} that 
\begin{equation}\label{eqn:w2bdd22}
  \EE \calW_2(\Pi_{\calM \#}(\mu\ast\eta_\delta), \Pi_{\calM \#}(\mu_n\ast\eta_\delta)) \lesssim \left\{\begin{array}{ll}
      n^{-1/2} \log(\delta^{-1}), &  d =2, \\
      n^{-1/2}\delta^{1-d/2}, & d\ge 3.
  \end{array}\right.  
\end{equation}

Since $T_\theta$ is Lipschitz in $\calM_\delta$ and hence contraction in $\calW_2$, we can compute 
    \begin{align*}
        \calW_2(\mu_{T_\theta}, T_{\theta \#}(\mu_n\ast\eta_\delta)) & =  \calW_2(\left(T_{\theta \#}(\mu_n \ast \eta_\delta)\right)_N, T_{\theta \#}(\mu_n\ast\eta_\delta)) \\
        & = \calW_2(T_{\theta \#}(\mu_n \ast \eta_\delta)_N, T_{\theta \#}(\mu_n\ast\eta_\delta)) \\
        & \le \mathrm{Lip}(T_\theta)\cdot \calW_2((\mu_n \ast \eta_\delta)_N, \mu_n\ast\eta_\delta).
    \end{align*}
    Let $Y_j = X_{I_j} + U_j$, where $I_{1:N}$ are i.i.d indices drawn uniformly from $\{1,\dots,n\}$ and $U_j \overset{i.i.d.}{\sim} \eta_\delta$. Let $(\mu_n)_N$ be the empirical measure of $\{X_{I_j}\}_{j=1}^N$. By the triangle inequality, the couplings $(Y_j,X_{I_j})$ and $(X_{I_1}+U_1,X_{I_1})$ conditionally on $X_{1:n}$, and $\mathrm{Supp}(\eta_\delta)\subset B_{\R^D}(0, \delta)$, we have 
    \begin{align*}
        \calW_2((\mu_n \ast \eta_\delta)_N, \mu_n\ast\eta_\delta) & \le \calW_2((\mu_n \ast \eta_\delta)_N, (\mu_n)_N) +\calW_2(\mu_n, \mu_n\ast\eta_\delta)  + \calW_2((\mu_n)_N, \mu_n) \\
        & = \calW_2\left(\frac{1}{N}\sum_{j=1}^N \delta_{Y_j}, \frac{1}{N}\sum_{j=1}^N \delta_{ X_{I_j}}\right)  +\calW_2(\mu_n, \mu_n\ast\eta_\delta) + \calW_2((\mu_n)_N, \mu_n)\\
        & \le \Big(\frac{1}{N}\sum_{j=1}^N |U_j|^2\Big)^{1/2}+ \sqrt{\EE |U_1|^2} + \calW_2((\mu_n)_N, \mu_n)\\
        & \le 2\delta + \calW_2((\mu_n)_N, \mu_n).
    \end{align*}
    Next we proceed to estimate $\calW_2((\mu_n)_N, \mu_n)$. For each $i=1,\dots,n$, define
\[
\hat c_i := \frac1N \sum_{j=1}^N \mathbf 1_{\{I_j=i\}}.
\]
Note that $I_j \overset{i.i.d.}{\sim} \mathrm{Unif(\{1,\cdots, n\})}$, we then have $(\mu_n)_N=\sum_{i=1}^n \hat c_i\,\delta_{X_i}$ and $\EE_{Y_{1:N} \mid X_{1:n}} \hat c_i = \frac{1}{n}$. Since both measures $(\mu_n)_N$, $\mu_n$ are supported on $X_{1:n}$, one can couple the common mass at each atom and only move the remaining mass. Moving any remaining mass costs at most distance $\mathrm{diam}(\calM)$, and the remaining mass equals the total variation distance. Hence,
\[
W_2\big((\mu_n)_N,\mu_n\big) \le \mathrm{diam}(\calM) \sqrt{\mathrm{TV}\big((\mu_n)_N,\mu_n\big)}
=\frac{\mathrm{diam}(\calM)}{\sqrt{2}}\sqrt{\sum_{i=1}^n \left|\hat c_i-\frac{1}{n}\right|}.
\]

Since the indicators $\mathbf 1_{\{I_j=i\}}$ are i.i.d. Bernoulli$(1/n)$ given $X_{1:n}$,
\[
\mathrm{Var}(\hat c_i)
= \frac{1}{N^2}\sum_{j=1}^N \mathrm{Var}(\mathbf 1_{\{I_j=i\}})
= \frac{1}{N^2}\,N\cdot \frac1n\Big(1-\frac1n\Big)
\le \frac{1}{nN}.
\]
Using the inequality $\mathbb E|Z-\mathbb EZ|\le \sqrt{\mathrm{Var}(Z)}$ gives
\[
\sum_{i=1}^n \EE_{Y_{1:N} \mid X_{1:n}}\Big|\hat c_i-\frac1n\Big| \le \sum_{i=1}^n \sqrt{\mathrm{Var}(\hat c_i)}
\le n\cdot \frac{1}{\sqrt{nN}}
= \sqrt{\frac{n}{N}}.
\]
Applying Jensen's inequality gives
\begin{align*}
    \EE_{X_{1:n}, Y_{1:N}}W_2\big((\mu_n)_N,\mu_n\big) & \le \frac{\mathrm{diam}(\calM)}{\sqrt{2}} \EE_{X_{1:n}, Y_{1:N}} \sqrt{\sum_{i=1}^n \left|\hat c_i-\frac{1}{n}\right|} \\
    & \le \frac{\mathrm{diam}(\calM)}{\sqrt{2}} \EE_{X_{1:n}} \sqrt{\sum_{i=1}^n \EE_{Y_{1:N} \mid X_{1:n}}\Big|\hat c_i-\frac1n\Big|} \\
    & \le \frac{\mathrm{diam}(\calM)}{\sqrt{2}} \left(\frac{n}{N}\right)^{1/4}.
\end{align*}

    Therefore, we have
    \begin{equation}\label{eqn:w2bdd42}
        \EE\calW_2(\mu_{T_\theta}, T_{\theta \#}(\mu_n\ast\eta_\delta))   \le 2 \mathrm{Lip}(T_\theta) \cdot \delta + \frac{\mathrm{diam}(\calM) \mathrm{Lip}(T_\theta)}{\sqrt{2}} \left(\frac{n}{N}\right)^{1/4}.
    \end{equation}

Moreover, considering the coupling $\pi:=(\Pi_\calM, T_\theta)_{\#}(\mu_n\ast \eta_\delta)$ between $\Pi_{\calM \#}(\mu_n\ast\eta_\delta)$ and $T_{\theta \#}(\mu_n\ast\eta_\delta)$ yields 
    \[\calW_2^2(\Pi_{\calM \#}(\mu_n\ast\eta_\delta), T_{\theta \#}(\mu_n\ast\eta_\delta)) \le \int_{\R^D \times\R^D}|x-y|^2\, d\pi(x,y) = \int_{\R^D}|\Pi_\calM(y) - T_\theta(y)|^2\, d\mu_n\ast \eta_\delta(y).\]
    By the definition of $\mathrm{PErr}$ and the fact $Y_j \overset{\text{i.i.d.}}{\sim} \mu_n \ast \eta_\delta$ for $j=1,\cdots,N$, we have
    \[ \mathbb{E}_{Y_{1:N} \mid X_{1:n}} \big(\mathrm{PErr}(T_\theta)\big) = \frac{ \mathbb{E}_{Y_{1:N}\mid X_{1:n}}|T_\theta(Y_j) - \Pi_\calM(Y_j)|^2 }{  \mathrm{tr}(\widehat{\Sigma}_n)} = \frac{ \int_{\R^D}|\Pi_\calM(y) - T_\theta(y)|^2\, d\mu_n\ast \eta_\delta(y) }{ \mathrm{tr}(\widehat{\Sigma}_n)}. \]
    Combining the previous two formulae gives
    \[ \calW_2(\Pi_{\calM \#}(\mu_n\ast\eta_\delta), T_{\theta \#}(\mu_n\ast\eta_\delta)) \le  \left(\mathrm{tr}(\widehat{\Sigma}_n)\mathbb{E}_{Y_{1:N}\mid X_{1:n}} \big(\mathrm{PErr}(T_\theta)\big)\right)^{\frac{1}{2}}.\]
    Since $\widehat{\Sigma}_n$ is the unbiased covariance of training samples $X_{1:n}$, it is straightforward to show that 
    \[  \mathrm{tr}(\widehat{\Sigma}_n) = \frac{1}{2n(n-1)}\sum_{i,j=1}^n\|X_i - X_j\|^2 \le \frac{n^2}{2n(n-1)}\mathrm{diam}(\calM)^2 \le \mathrm{diam}(\calM)^2. \]
    Therefore, using Jensen's inequality gives
    \begin{align}
        \mathbb{E}_{X_{1:n}, Y_{1:N}}\calW_2(\Pi_{\calM \#}(\mu_n\ast\eta_\delta), T_{\theta \#}(\mu_n\ast\eta_\delta)) & \lesssim \mathbb{E}_{X_{1:n}}\sqrt{\mathbb{E}_{Y_{1:N}\mid X_{1:n}} \big(\mathrm{PErr}(T_\theta)\big)} \notag \\
        & \le \sqrt{\mathbb{E}_{X_{1:n}, Y_{1:N}} \big(\mathrm{PErr}(T_\theta)\big)}. \label{eqn:w2bdd32}
    \end{align}
Plugging in \eqref{eqn:w2bdd21}--\eqref{eqn:w2bdd32} into \eqref{eqn:wassmuTtheta} gives \eqref{eqn:wassbdd2}. 
\end{proof}

\subsection{Algorithm details}\label{app:algorithm-details}

Algorithm~\ref{alg:twostage} states details of the training routine described in Section~\ref{sec:two_stage}. At each epoch,  we resample noisy points $\{Y_j\}_{j=1}^N$ around the training set, rather than fixing one noisy dataset in advance.  After training, generation uses fresh tubular noise and applies the learned map $T_\theta$ to obtain generated samples.

The pretraining loss is given by
\[
E_{\mathrm{pre}}(T) =
 \frac{1}{n}\sum_{i=1}^n|T(X_i)-X_i|^2
 + \frac{1}{N}\sum_{j=1}^N |T(Y_j)-\Pi_n Y_j|,
\]
where $\Pi_n$ is the nearest training sample map. And the full projection loss is given by $E_n$ in \eqref{def:En}.
The implementation also supports weighted versions of these terms and
mini-batch training; the numerical choices of each experiment are reported in Appendix~\ref{app:numerical-setup}.

\begin{algorithm*}[tbh]
    \caption{Two-stage training of Projection Learning (PL)}
    \begin{algorithmic}[1]  \label{alg:twostage}
        \STATE \textbf{Input:} Training sample $\{X_i\}_{i=1}^n \sim \mu$
        \STATE {\bfseries Initialize:} neural network $T_\theta$; noise magnitude $\delta$; number of noisy points per sample $N_p$; number of epochs $k_{\mathrm{pre}}, k_{\mathrm{main}}$; learning rates $\alpha_{\mathrm{pre}},  \alpha$

        \vspace{0.3em}
        \STATE {\bfseries Stage I: pretraining}
        \FOR{ $k = 1$ to $k_{\mathrm{pre}}$ }
                \STATE $Y_j = X_{\lfloor (j-1)/N_p +1 \rfloor} + U_j$ with $U_j \sim \eta_\delta$ for $j \in [n N_p]$
                \hfill $\triangleright$ Add i.i.d. noise to each sample
                \STATE Update $\theta \leftarrow \theta - \alpha_{\mathrm{pre}} \nabla_\theta E_{\mathrm{pre}}(T_\theta; \{X_i\}, \{Y_j\})$
                \hfill $\triangleright$ Pretraining loss
        \ENDFOR

        \vspace{0.3em}
        \STATE {\bfseries Stage II: training of the full model}
        \FOR{ $k = 1$ to $k_{\mathrm{main}}$ }
                \STATE Create noised points $\{Y_j\}_{j=1}^N$ as above with $N=n N_p$
                \STATE Update $\theta \leftarrow \theta - \alpha \nabla_\theta E_n(T_\theta; \{X_i\}, \{Y_j\})$
                \hfill $\triangleright$ Projection loss
        \ENDFOR

        \vspace{0.3em}
        \STATE {\bfseries Output:} Trained mapping $T_\theta$
        \STATE \textbf{Generated points:} Create $N_{\textrm{gen}}$ noised points $Y_j$ and set $\widehat{X}_j \leftarrow T_\theta(Y_j)$
    \end{algorithmic}
\end{algorithm*}

Algorithm~\ref{alg:ipl} is the iterated version used when a single tubular radius is too large or the geometry is more complex and described in Section~\ref{sec:iterated_projection}.  It trains a
cascade of projection maps at decreasing radii.  Each stage uses the same
network architecture; the first stage is initialized randomly and later
stages are warm-started from the previous stage.  During generation, noised
points from the largest radius are pushed through the learned cascade.
\begin{algorithm}[tbh]
  \caption{Iterated Projection Learning (IPL)}\label{alg:ipl}
  \begin{algorithmic}[1]
    \STATE \textbf{Input:} Training samples $\{X_i\}_{i=1}^{n}$;
           outer radii $\delta_1\!>\!\cdots\!>\!\delta_K$;
           inner radii
           $\underline{\delta}_1\!\ge\!\cdots\!\ge\!\underline{\delta}_K$
    \STATE \textbf{Initialise:} network $T_{\theta_1}$ at random

    \vspace{0.3em}
    \STATE \textbf{Stage~$1$:}
           Train $T_{\theta_1}$ on $\{X_i\}_{i=1}^{n}$ with noise radius $\delta_1$
           and inner radius $\underline{\delta}_1$ by
           Algorithm~\ref{alg:twostage}

    \vspace{0.3em}
    \FOR{$k=2$ \textbf{to} $K$}
      \STATE Initialise $\theta_k\leftarrow\theta_{k-1}$
             \hfill$\triangleright$ warm-start $T_{\theta_k}$ from
             the previous projection $T_{\theta_{k-1}}$
      \STATE $X_i^{(k)}\leftarrow T_{\theta_{k-1}}(X_i+V_i)$,
             $V_i\sim\eta_{\delta_{k-1}}$, $i\in[n]$
             \hfill$\triangleright$ refined anchors near $\calM$
      \STATE Train $T_{\theta_k}$ by Algorithm~\ref{alg:twostage} with $(\delta_k, \underline{\delta}_k)$, using refined inputs $\{X_i^{(k)}\}_{i=1}^{n}$ as noised points
    \ENDFOR

    \vspace{0.3em}
    \STATE \textbf{Output:} Stage projections $T_{\theta_1},\ldots,T_{\theta_K}$, set $T=T_{\theta_K}\circ \cdots \circ T_{\theta_1}$
    \STATE \textbf{Generation:} Draw $N_{\mathrm{gen}}$ noised points
           $Y_j=X_{\lfloor(j-1)/N_p\rfloor+1}+U_j$, $U_j\!\sim\!\eta_{\delta_1}$, and set
           $\wh{X}_j\leftarrow T(Y_j)$
  \end{algorithmic}
\end{algorithm}

\subsection{Implementation details of experiments in Section~\ref{sec:experiments}}\label{app:numerical-setup}

\subsubsection{Projection learning experiments using MLP network architecture}
All synthetic and low-dimensional image experiments use the same fully connected architecture for the projection model, which maps $\R^D$ to $\R^D$ via
\[
  D \to 100 \to 200 \to 100 \to D,
\]
with SiLU activations and Adam optimization.  The experiment-specific
choices are summarized in Table~\ref{tab:mlp_setup}. All synthetic manifold experiments summarized in the table use Algorithm~\ref{alg:twostage}.

\begin{table}[h]
\centering
\caption{Numerical setup for MLP-based projection experiments.}
\label{tab:mlp_setup}
\small
\setlength{\tabcolsep}{5pt}
\renewcommand{\arraystretch}{1.15}
\begin{tabular}{p{0.27\linewidth}p{0.63\linewidth}}
\toprule
\multicolumn{1}{c}{Experiment} & \multicolumn{1}{c}{Main numerical setup} \\
\midrule
Torus--circle union in $\R^3$ &
$n=700$ ($600$ points from torus, $100$ points from circle), $D=3$, $\delta=0.25$,
$\underline{\delta}=0.125$, $N_p=10$ noisy points per sample, $5,000$ training epochs.  \\
\addlinespace
Sample complexity scaling for $d$-torus &
For $d=1,\ldots,5$, under threshold $\mathrm{MIE}<0.02$, we first search for the smallest
$n^\star(d)$ in $D=2d$ with $\delta=1$, $N_p=100$ noisy points per sample and $2000$ epochs. The second part embeds the same
tori in $D=256$ and compares $\delta=1$ with
$\delta=16/\sqrt d$ with same setting. \\
\addlinespace
$2$-torus embedded in $\R^{256}$ &
$n=480$, $D=256$, $\delta=7$, $\underline{\delta}=0.7$, $N_p=50$ noisy points per sample, $5,000$ training epochs and five independent runs. \\
\addlinespace
$S^1$ in the space of images &
$n=30$, $D=64$, $\delta=3$, $N_p=250$ noisy points per sample. This experiment also
includes the diffusion baseline described below. \\
\bottomrule
\end{tabular}
\end{table}

\subsubsection{$S^1$ in the space of $8\times 8$ images}
\label{app:exp-circle-images}

Let $D=[-1,1]^2$ and discretise $D$ on an $8\times 8$ uniform
Cartesian grid $\{(x_i,y_j)\}_{i,j=1}^{8}$.  For each
$t\in[0,2\pi)$, the image function \eqref{eq:It} defines a vector
$\gamma(t)\in\mathbb{R}^{64}$, and
$\calM=\gamma\bigl([0,2\pi)\bigr)$ is a smooth one-dimensional manifold
embedded in $\mathbb{R}^{64}$.

\paragraph{Basis functions and coefficient analysis.}
To verify that generated samples respect the manifold geometry, we
project images onto three discrete $L^2$-orthogonal basis functions:
\[
  f_0(x,y)=\tfrac{1}{2},\qquad
  f_1(x,y)=\tfrac{\sqrt{3}}{2}\,x,\qquad
  f_2(x,y)=\tfrac{\sqrt{3}}{2}\,y.
\]
The discrete inner product on the $8\times 8$ grid is
$\langle u,v\rangle_h = \sum_{i,j}u_{ij}\,v_{ij}\,h^2$,
where $h$ is the uniform step size.  For any image $x$ (training or
generated) we compute
\[
  c_\ell(x) := \frac{\langle x, f_\ell\rangle_h}{\langle f_\ell, f_\ell\rangle_h},
  \qquad \ell=0,1,2.
\]
Since training images satisfy
$I_t = f_0 + \frac{\cos t}{\sqrt{6}}\,f_1
            + \frac{\sin t}{\sqrt{6}}\,f_2$,
the coefficients $(c_1,c_2)$ of training samples lie on a circle of
radius $1/\!\sqrt{6}$.  A well-trained generative model should
produce samples whose $(c_1,c_2)$ coefficients also lie near 
this circle.

\begin{figure}[h]
  \centering
  \begin{minipage}[b]{0.66\textwidth}
    \centering
    \includegraphics[width=\linewidth]{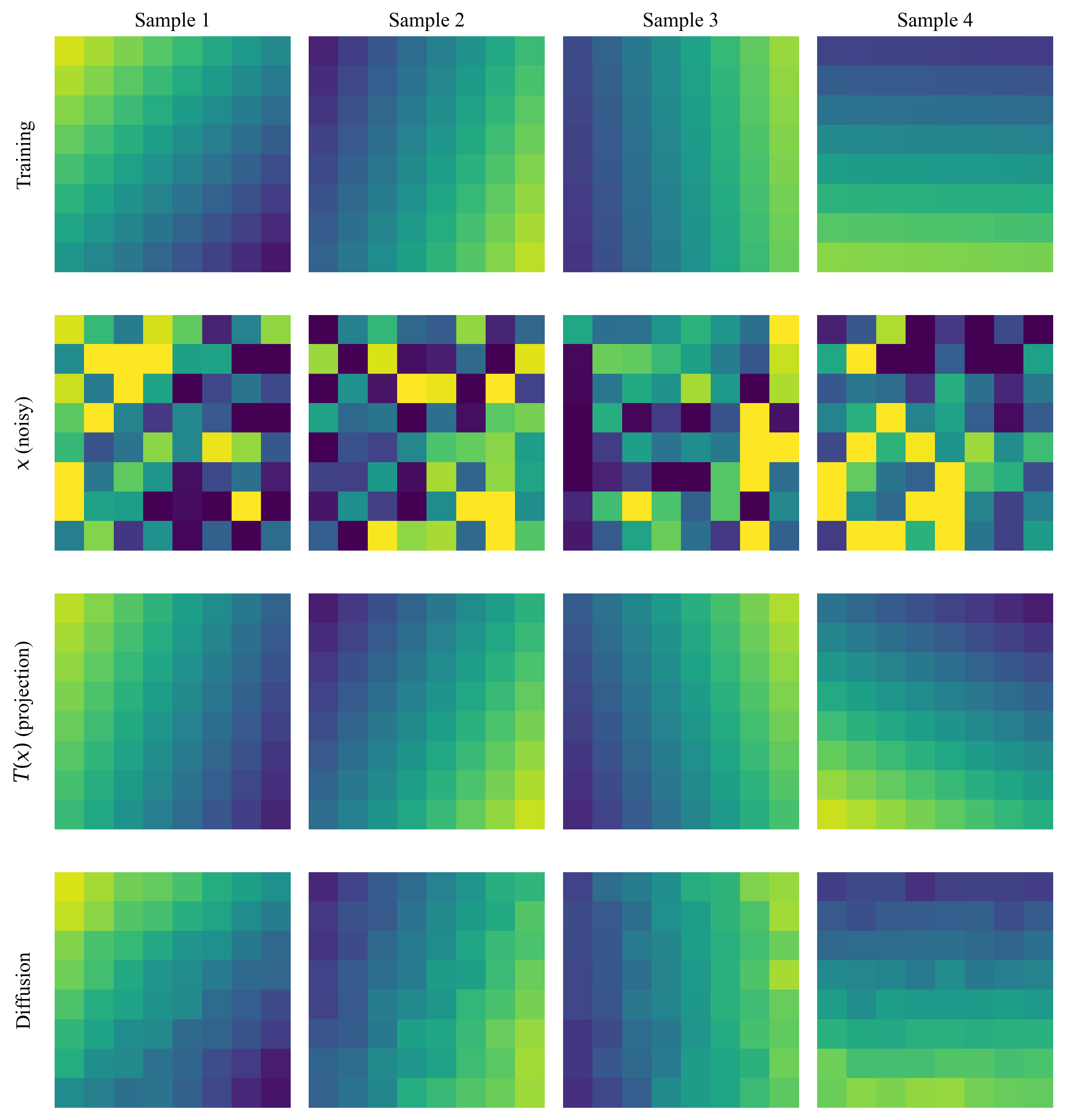}
    \par\smallskip
    \textbf{(a)} Sample images for the image-valued 1-torus.
  \end{minipage}\hspace{-0.01\textwidth}
  \begin{minipage}[b]{0.34\textwidth}
    \centering
    \includegraphics[width=\linewidth]{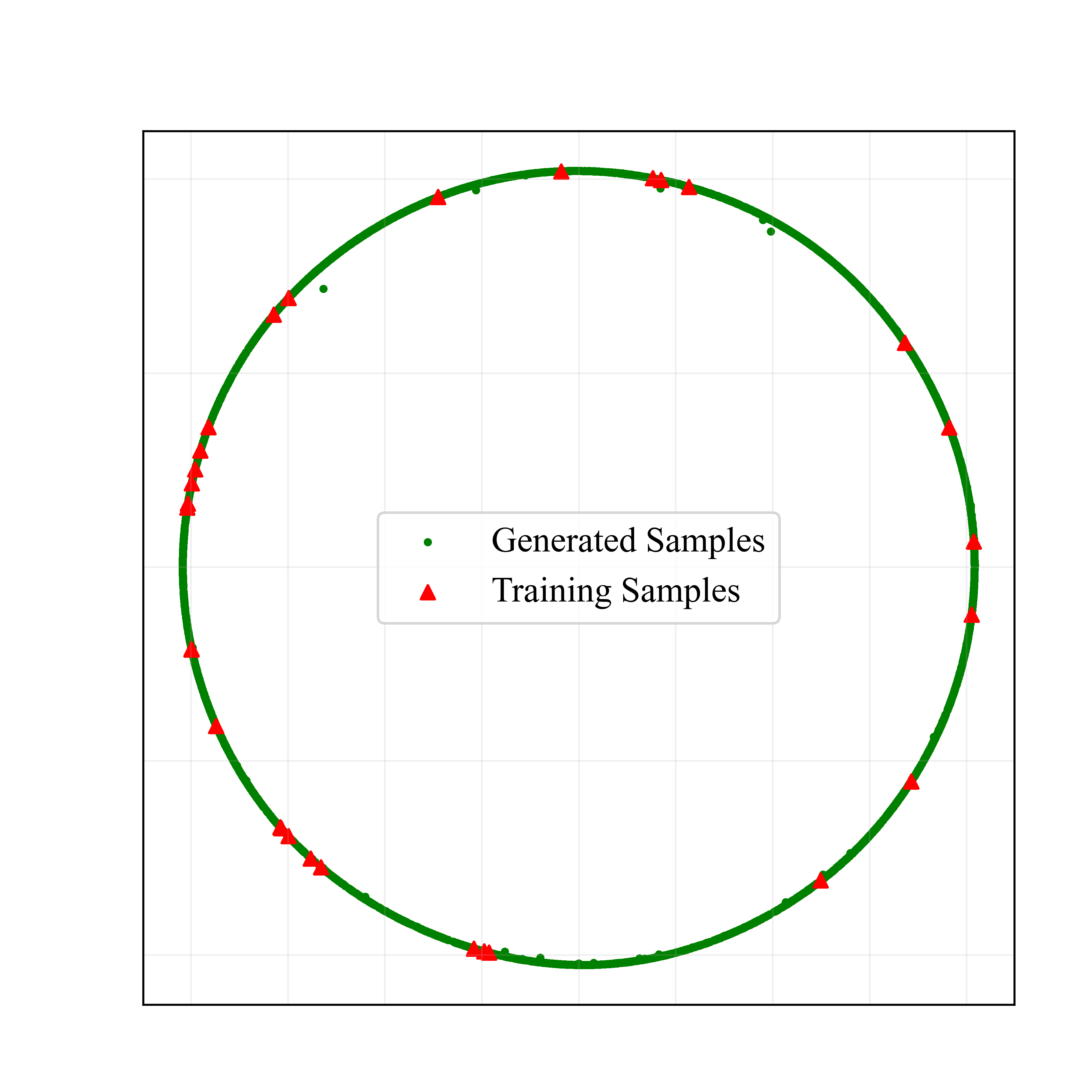}
    \par\smallskip
    \textbf{(b)} Projection model.

    \includegraphics[width=\linewidth]{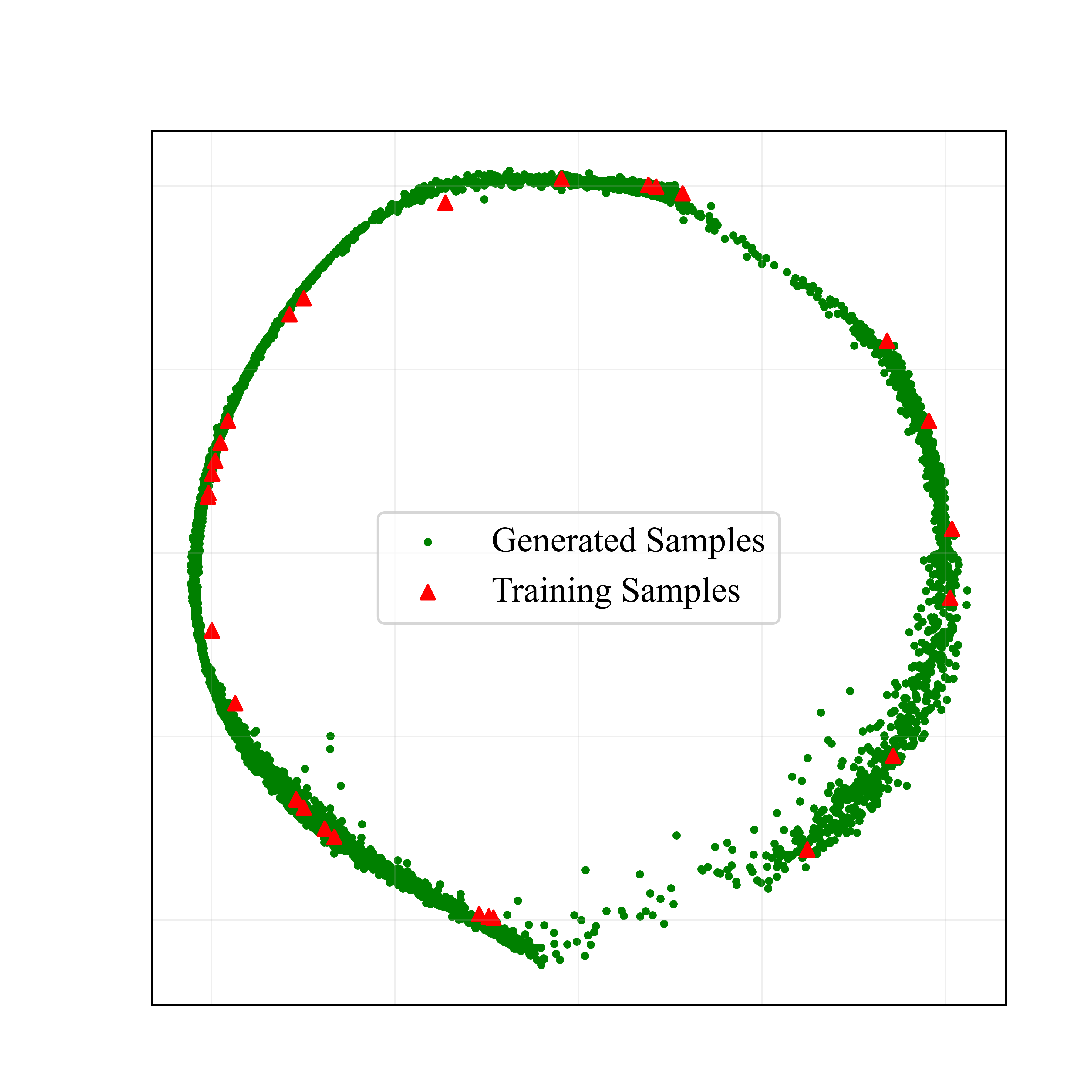}
    \par\smallskip
    \textbf{(c)} Diffusion model.
  \end{minipage}
  \caption{\textbf{Left:} Four representative samples show the training image, a noisy
    tubular point, the corresponding projected output (projection-model
    generated sample), and the diffusion output. \textbf{Right:} Coefficients
    $(c_1,c_2)$ of $5{,}000$ generated samples for the image-valued 1-torus
    experiment. Red triangles denote training samples lying on a circle of
    radius $1/\!\sqrt{6}$, and green dots denote generated samples.
    \textbf{(b)} The projection model produces samples whose coefficients lie
    on the circle. \textbf{(c)} The diffusion model's samples scatter
    noticeably off-circle.}
  \label{fig:circle-illustration-coefficients-combined}
\end{figure}

Figure~\ref{fig:circle-illustration-coefficients-combined} plots the $(c_1,c_2)$
coefficients for $5,000$ generated samples from each model alongside the 30 training samples.
The projection model's coefficients lie tightly on the target circle, confirming that the learned map projects onto $\calM$.
The diffusion model's coefficients exhibit visible dispersion away from the circle, particularly a ``cloud'' of off-manifold points, consistent with its higher MIE and the memorization pattern observed in the $P_r$ curves.

\paragraph{Architecture.}
The \emph{projection model} is a fully connected network
$64 \to 100 \to 200 \to 100 \to 64$ with SiLU activations
(53,264 parameters).
Phase~1 (pretraining) uses 400~epochs with Adam
($\mathrm{lr}=2\times 10^{-3}$).
Phase~2 (projection) uses
4\,600~epochs with Adam ($\mathrm{lr}=10^{-3}$) and a cosine-annealing
learning-rate schedule.
In each epoch, 250 noisy copies per training sample are drawn from the
tubular neighborhood $\calM_\delta$ with $\delta=3$. 

The \emph{diffusion model} uses a three-hidden-layer MLP score network
$64\to 256\to 512\to 256\to 64$ with SiLU activations and a
256-dimensional Gaussian Fourier time embedding
(${\approx}625$k parameters).
It is trained for 50\,000 epochs with denoising
score-matching loss for VE-SDE and 
Adam optimizer ($\mathrm{lr}=10^{-3}$). Since the number of training samples is small, at each epoch we draw $250$ independent noise levels for each training image, together with independent Gaussian perturbations at those noise levels.
Generation uses the probability-flow ODE with $5,000$ Euler steps.

\paragraph{Evaluation metrics.}
We evaluate two quantities every 200 epochs (projection) and every
1\,000 epochs (diffusion), each computed on $5,000$ generated samples:
\begin{itemize}[leftmargin=*]
  \item MIE defined in \eqref{eqn:MIE}: mean
    radial deviation of the $(c_1,c_2)$ coefficients from the circle
    of radius $1/\!\sqrt{6}$.
    \item Memo defined in \eqref{eqn:memoscore}: compute $\mathrm{P}_{r}^{\mathrm{gen}}$ and  $\mathrm{P}_{r}^{\mathrm{true}}$ for $(c_1,c_2)$ coefficients of generated samples and training points, respectively.
\end{itemize}
And the evaluation result is summarized in Figure~\ref{fig:circle-metrics}.

\begin{comment}
\end{comment}

\paragraph{Model capacity sweep.}
We perform a five-seed capacity sweep on the image-valued circle, using the
same 30 training images and fixed validation for projection and VE--SDE models.  We call the four capacity levels tiny, small, medium, and
large, where the small model is the architecture used in
the preceding experiment.  Each projection model is trained for $10{,}000$ projection updates with  $r_{\mathrm{inner}}=1.5$ and 250 noisy copies per training sample and a cosine-annealed Adam learning rate initialized at $10^{-3}$, $8\times10^{-4}$, $4\times10^{-4}$, and $5\times10^{-5}$, respectively.  On the other hand, each VE--SDE score model is
trained by denoising score matching for $50{,}000$ full-data updates with
$M=250$ independent noise levels per training image, $\sigma=25$, and
Adam with learning rate $10^{-3}$; samples are generated with a $5{,}000$-step probability-flow ODE.  For every seed and capacity, we evaluate projection checkpoints every 200 projection updates and VE--SDE checkpoints every 1,000 updates.  We select the checkpoint with the
lowest MIE on the fixed held-out validation set, independently for each run, and compute its metrics on the fixed held-out test set.  Table~\ref{tab:image-capacity-sweep} reports the mean and standard deviation of these test metrics over five seeds. Projection performance is robust across the full capacity range: all four models attain mean MIE below $10^{-3}$ and mean Memo below $9\times10^{-3}$.  VE--SDE is much more capacity-sensitive.  Its tiny and small models substantially underfit, and hence their zero Memo values are not evidence of faithful non-memorizing generation.  Increasing VE--SDE capacity reduces MIE to $0.0139$ for the medium model and $0.0055$ for the large model, but Memo simultaneously rises to $0.136$ and $0.473$, respectively.

\begin{table}[ht]
  \centering \small
  \caption{Five-seed capacity sweep for $S^1$ in the space of images. Values
  are mean $\pm$ standard deviation at the checkpoint minimizing validation
  MIE. PErr is defined only for projection maps.}
  \label{tab:image-capacity-sweep}
  \resizebox{\textwidth}{!}{%
  \begin{tabular}{llrrrr}
    \toprule
    Method & Capacity & Parameters & MIE & PErr & Memo \\
    \midrule
    Projection & Tiny   & 12,480    & $0.0006 \pm 0.0003$ & $(2.8721 \pm 0.7237)\times10^{-5}$ & $0.0033 \pm 0.0024$ \\
    Projection & Small  & 53,264    & $0.0005 \pm 0.0001$ & $(2.6743 \pm 0.7537)\times10^{-5}$ & $0.0066 \pm 0.0028$ \\
    Projection & Medium & 296,256   & $0.0009 \pm 0.0002$ & $(3.0018 \pm 0.4702)\times10^{-5}$ & $0.0084 \pm 0.0038$ \\
    Projection & Large  & 2,167,360 & $0.0009 \pm 0.0004$ & $(3.4688 \pm 1.6441)\times10^{-5}$ & $0.0056 \pm 0.0018$ \\
    \midrule
    VE--SDE & Tiny   & 45,696    & $2.4922 \pm 0.1145$ & -- & $0.0000 \pm 0.0000$ \\
    VE--SDE & Small  & 165,056   & $0.1523 \pm 0.0019$ & -- & $0.0000 \pm 0.0000$ \\
    VE--SDE & Medium & 624,960   & $0.0139 \pm 0.0055$ & -- & $0.1361 \pm 0.0480$ \\
    VE--SDE & Large  & 1,708,352 & $0.0055 \pm 0.0005$ & -- & $0.4737 \pm 0.0511$ \\
    \bottomrule
  \end{tabular}%
  }
\end{table}

\begin{figure}[ht]
  \centering
  \includegraphics[width=\textwidth]{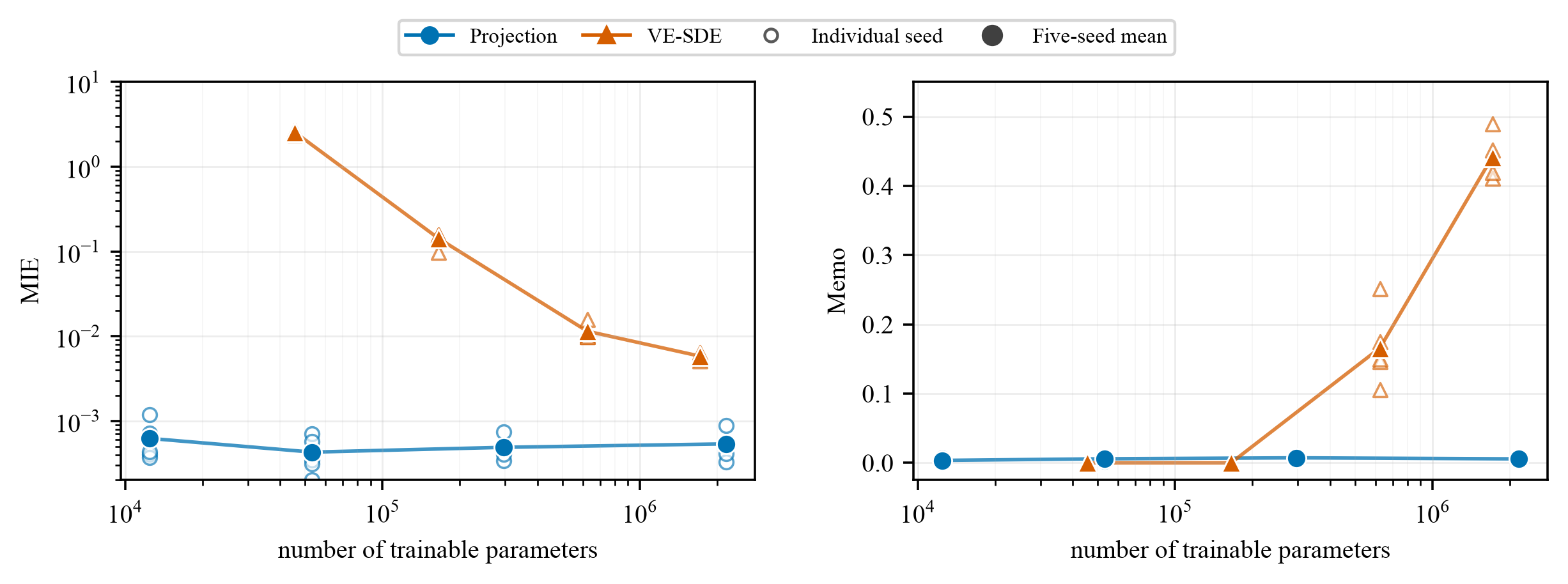}
  \caption{Five-seed capacity sweep results versus the number of trainable
  parameters. Open markers show the five individual seeds, and filled markers
  show their means. Circles denote projection models and triangles denote
  VE--SDE diffusion models. For each method, the four points from left to right
  correspond to the tiny, small, medium, and large capacities in Table~\ref{tab:image-capacity-sweep}.}
  \label{fig:image-capacity-selected-summary}
\end{figure}

\subsubsection{U-Net MNIST setup and diffusion baseline}\label{app:mnist}

\paragraph{Dataset and CapsNet checkpoints.}
For the MNIST evaluation, we use the MNIST dataset~\citep{lecun1998gradient} and a pretrained CapsNet checkpoint from the open-source PyTorch implementation of CapsNet by Cedrick Chee:
\url{https://github.com/cedrickchee/capsule-net-pytorch}.  The CapsNet implementation and checkpoint are released under the MIT License. We use the checkpoint only as a fixed feature extractor for evaluation metrics, not as a trained component of our method. We cite the original CapsNet paper~\citep{sabour2017dynamic} and credit the implementation/checkpoint source. All synthetic datasets are generated by our own code.

\paragraph{Data split and reference sets.}
For each random split, both IPL and VE-SDE diffusion model use the same $10,000$ MNIST training samples. The remaining training pool consists of $50K$ images and is randomly subsampled to $10,000$ samples as the held-out reference. The official MNIST test dataset ($10,000$ images) is also evaluated as a secondary reference. Since these two references give very similar FID/KID values , we only report the summarized result in Table~\ref{tab:main} in the main content. Table~\ref{tab:perseed_ipl} lists every metric for every IPL seed against both the held-out reference and the MNIST test dataset. The two reference choices agree to within $\Delta\mathrm{FID}<0.004$ on every seed, which justifies the single-reference reporting in the MNIST evaluation in Section~\ref{sec:experiments}. Since the MNIST test dataset is fixed for each seed, we use it as the reference to approximate empirical $P_r^{\mathrm{true}}$ in Figure~\ref{fig:agg_pr_test}.

\textbf{IPL architecture and training.}
The MNIST projection model is a time-independent U-Net with channel widths $(16,32,64,128)$, $3\times3$ convolutions, GroupNorm, Swish activation
$x\sigma(x)$, and skip connections.  We train the U-Net using
Algorithm~\ref{alg:ipl} with the three-stage schedule
\[
  (\delta_k,\underline\delta_k)
  \in \{(20,8),(8,4),(4,2)\}.
\]
Each stage uses 100 pretraining epochs and 50 projection epochs, Adam with
learning rates $2\times10^{-3}$ and $5\times10^{-4}$ for the two phases,
50 noisy points per image for the refined-anchor construction, and mini-batches of size 64.  Generation perturbs each training image once at the largest radius $\delta_1=20$ and applies the learned cascade, producing $10{,}000$ generated images per seed.

\textbf{Diffusion baseline.}
The unconditional VE-SDE score model of \citet{song2021scorebased} is
trained on the same five splits used for IPL, with $\sigma=25$, batch size
$64$, Adam with learning rate $10^{-4}$ and $300$ epochs ($\approx\!47$K gradient
steps per seed).
Sampling employs a $500$-step Predictor-Corrector solver with
signal-to-noise ratio $0.16$.
The baseline reaches FID and KID values comparable to the IPL cascade and is therefore included in the MNIST evaluation in Section~\ref{sec:experiments} as a quality benchmark.

\begin{table}[h]
\centering
\caption{IPL per-seed metrics with held-out (ho) reference and test (te) reference.
         The bottom block reports the across-seed mean, the sample standard
         deviation, the $95\%$ Student-$t$ confidence half-width (df=4),
         the one-sample $t$-statistic and the corresponding $p$-value
         against $H_{0}\!:\!x=0$.}
\label{tab:perseed_ipl}
\renewcommand{\arraystretch}{1.10}
\small
\begin{tabular}{lccccccc}
\toprule
Seed & FID$_{\mathrm{ho}}$ & KID$_{\mathrm{ho}}$
     & FID$_{\mathrm{te}}$ & KID$_{\mathrm{te}}$
     & Memo$_{\mathrm{ho}}$ & Memo$_{\mathrm{te}}$ & Memo$_{\mathrm{pix}}$ \\
\midrule
\#1 & 0.0753 & $2.00\!\times\!10^{-4}$ & 0.0718 & $1.90\!\times\!10^{-4}$
  & 0.0000 & 0.0000 & 0.000 \\
\#2 & 0.0470 & $1.30\!\times\!10^{-4}$ & 0.0435 & $1.10\!\times\!10^{-4}$
  & 0.0114 & 0.0232 & 0.000 \\
\#3 & 0.0767 & $1.80\!\times\!10^{-4}$ & 0.0754 & $1.70\!\times\!10^{-4}$
  & 0.0201 & 0.0420 & 0.000 \\
\#4 & 0.0228 & $0.60\!\times\!10^{-4}$ & 0.0221 & $0.50\!\times\!10^{-4}$
  & 0.0257 & 0.0350 & 0.000 \\
\#5 & 0.0607 & $1.60\!\times\!10^{-4}$ & 0.0572 & $1.40\!\times\!10^{-4}$
  & 0.0000 & 0.0000 & 0.000 \\
\midrule
mean        & 0.0565 & $1.40\!\times\!10^{-4}$ & 0.0540 & $1.30\!\times\!10^{-4}$
            & 0.0114 & 0.0200 & 0.000 \\
std (df=4)  & 0.0224 & $0.50\!\times\!10^{-4}$ & 0.0219 & $0.60\!\times\!10^{-4}$
            & 0.0116 & 0.0195 & 0.000 \\
$95\%$ CI $\pm$ & 0.0278 & $0.70\!\times\!10^{-4}$ & 0.0272 & $0.70\!\times\!10^{-4}$
            & 0.0144 & 0.0242 & 0.000 \\
$t$-stat ($H_0\!=\!0$) & --- & $+5.94$ & --- & $+5.17$
            & $+2.20$ & $+2.30$ & --- \\
$p$-value   & --- & 0.004 & --- & 0.007
            & 0.093 & 0.083 & --- \\
\bottomrule
\end{tabular}
\end{table}

\begin{figure}[h]
    \centering
    \includegraphics[width=0.9\linewidth]{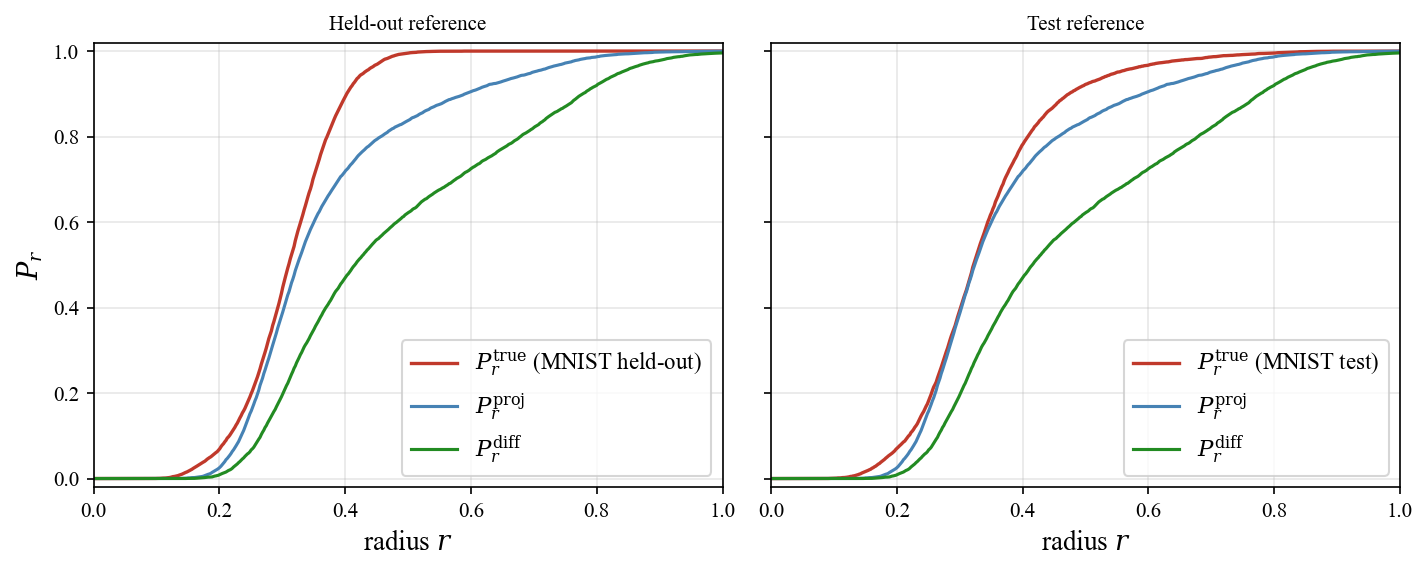}
    \caption{$P_r$ curve in the 160-dimensional CapsNet feature space for one seed. The \emph{left} panel uses the held-out MNIST training pool to approximate empirical $P_r^{\mathrm{true}}$, and the \emph{right} panel uses the MNIST test dataset.}
    \label{fig:prseed0}
\end{figure}

\textbf{Statistical evaluation.}
For every metric we report the seed-wise mean and a Student-$t$ 95\,\%
confidence interval (df $= 4$); the null hypothesis $H_{0}\!:\!x=0$ is
tested by a one-sample $t$-test.
KID is significantly positive at $p\!<\!0.01$, as expected of a
finite-reference biased estimator, and the same conclusion holds for the
diffusion baseline.
The CapsNet-space $P_r$-gap of IPL, defined as in Section~\ref{sec:memorization}, yields a positive point estimate but
does not reject $H_{0}$ at conventional level ($p\!\approx\!0.08$), indicating no statistically detectable excess concentration of generated samples near the training set.
The pixel-space ratio score does not reject $H_{0}$ at any conventional level for either model.

\textbf{Computing resources.}
Each IPL seed was trained on a single NVIDIA A100 (40\,GB) using
$\approx\!12$\,GB of peak GPU memory and $\approx\!70$ minutes of
wall-clock time per seed --- including all three cascade stages and the
final $10{,}000$-sample generation --- amounting to $\approx\!4$ GPU-hours
over the five seeds.
The VE-SDE baseline was trained on an NVIDIA T4 (16\,GB) with
$\approx\!2$\,GB of peak GPU memory; per-seed training took
$\approx\!10$ minutes, and PC sampling of $10{,}000$ images required $\approx\!10$ minutes.
The full evaluation pipeline reproduces every metric of
Table~\ref{tab:main} on a single CPU/MPS machine in $\approx\!15$ minutes.

\textbf{Memorization and distribution faithfulness.} The pixel-space ratio memorization score $\Memo_{\mathrm{pix}}$ and $P_r$ memorization score in CapsNet space measure different phenomena.
The ratio score asks whether a generated image is markedly
closer to a single training image than to its second-nearest neighbor, i.e.\ it is a hard near-copy detector.
While $P_r$-gap defined in Section~\ref{sec:memorization}, by contrast, asks whether the \emph{distribution} of feature
distances to training data is close.
In the saved evaluation, both models have zero pixel-space ratio score, while the IPL cascade has a small positive CapsNet-space memorization whose confidence interval includes zero. Figure~\ref{fig:prseed0} shows that the IPL curve tracks the $P_r^{\mathrm{true}}$ curve more closely than the diffusion curve in both reference approximations throughout the radius range. This provides direct evidence that IPL produces a distributionally faithful approximation of the data distribution without collapsing to the training dataset.

\end{document}